\documentclass[runningheads]{llncs}

\usepackage{eccv}

\usepackage{eccvabbrv}

\usepackage{graphicx}
\usepackage{booktabs}

\usepackage[accsupp]{axessibility}  

\usepackage{booktabs,tabularx}
\usepackage{pifont} 
\usepackage{multirow}
\usepackage{makecell}
\usepackage{caption}
\usepackage{wrapfig}
\usepackage{tcolorbox}
\usepackage{enumitem}
\usepackage{xcolor}
\definecolor{tickbg}{RGB}{230,255,230}
\definecolor{crossbg}{RGB}{255,230,230}
\definecolor{darkgreen}{RGB}{0,100,0}
\definecolor{warningbg}{RGB}{255,248,220}
\definecolor{tickbg}{HTML}{E6F4EA}   
\definecolor{crossbg}{HTML}{FDECEA}  
\newlength{\ynboxw}
\newcommand{\YNbadge}[2]{%
  \begingroup
    \setlength{\fboxsep}{0.5pt}
    \colorbox{#1}{\makebox[\ynboxw]{%
      \raisebox{0pt}[1.8ex][0.7ex]{\footnotesize #2}%
    }}%
  \endgroup
}

\newcommand{\Yes}{\YNbadge{tickbg}{\ding{51}}} 
\newcommand{\No}{\YNbadge{crossbg}{\ding{55}}} 
\usepackage{algorithm}
\usepackage[noend]{algpseudocode} 

\usepackage{hyperref}
\hypersetup{
    colorlinks=true,
    urlcolor=[RGB]{0, 90, 180},   
    linkcolor=[RGB]{0, 90, 180},
    citecolor=[RGB]{0, 90, 180}
}

\usepackage{orcidlink}
\usepackage{graphicx}
\usepackage{subcaption}
\usepackage[dvipsnames]{xcolor} 
\usepackage{ifthen}             

\newif\ifshowcomments
\showcommentstrue          

\newcommand{\dataset}{DermCase}
\newcommand{\llm}{DeepSeek-R1}

\begin{document}

\title{Mind the Rarities: Can Rare Skin Diseases Be Reliably Diagnosed via Diagnostic Reasoning?} 





\author{Yang Liu\inst{1}$^\dagger$ \and
Jiyao Yang\inst{2}$^\dagger$ \and
Hongjin Zhao\inst{3} \and
Xiaoyong Li\inst{2} \and
Yanzhe Ji\inst{3} \and
Xingjian Li\inst{1} \and
Runmin Jiang\inst{1} \and
Tianyang Wang\inst{2} \and
Saeed Anwar\inst{4} \and
Dongwoo Kim\inst{5} \and
Yue Yao\inst{3} \and
Zhenyue Qin\inst{6} \and
Min Xu\inst{1}}

\authorrunning{Y.~Liu et al.}

\institute{
Carnegie Mellon University \and
University of Alabama at Birmingham \and
Australian National University \and
University of Western Australia \and
POSTECH \and
Yale University \\
\smallskip
\small{$^\dagger$Equal contribution.}
}

\maketitle

\begin{abstract}
  Large vision-language models (LVLMs) demonstrate strong performance in dermatology; however, evaluating diagnostic reasoning for rare conditions remains largely unexplored. Existing benchmarks focus on common diseases and assess only final accuracy, overlooking the clinical reasoning process, which is critical for complex cases. We address this gap by constructing DermCase, a long-context benchmark derived from peer-reviewed case reports. Our dataset contains 26,030 multi-modal image-text pairs and 6,354 clinically challenging cases, each annotated with comprehensive clinical information and step-by-step reasoning chains. To enable reliable evaluation, we establish DermLIP-based similarity metrics that achieve stronger alignment with dermatologists for assessing differential diagnosis quality. Benchmarking 22 leading LVLMs exposes significant deficiencies across diagnosis accuracy, differential diagnosis, and clinical reasoning. Fine-tuning experiments demonstrate that instruction tuning substantially improves performance while Direct Preference Optimization (DPO) yields minimal gains. Systematic error analysis further reveals critical limitations in current models' reasoning capabilities. Our project website, dataset and code are available at 
  \href{https://yliu1082.github.io/DermCase/}{https://yliu1082.github.io/DermCase/}.
  
  \keywords{Dermatology \and Large vision-language models \and Diagnostic reasoning \and Rare skin diseases}
\end{abstract}

\section{Introduction}
\label{sec:intro}
\begin{figure}[t]
    \centering
    \includegraphics[width=\textwidth]{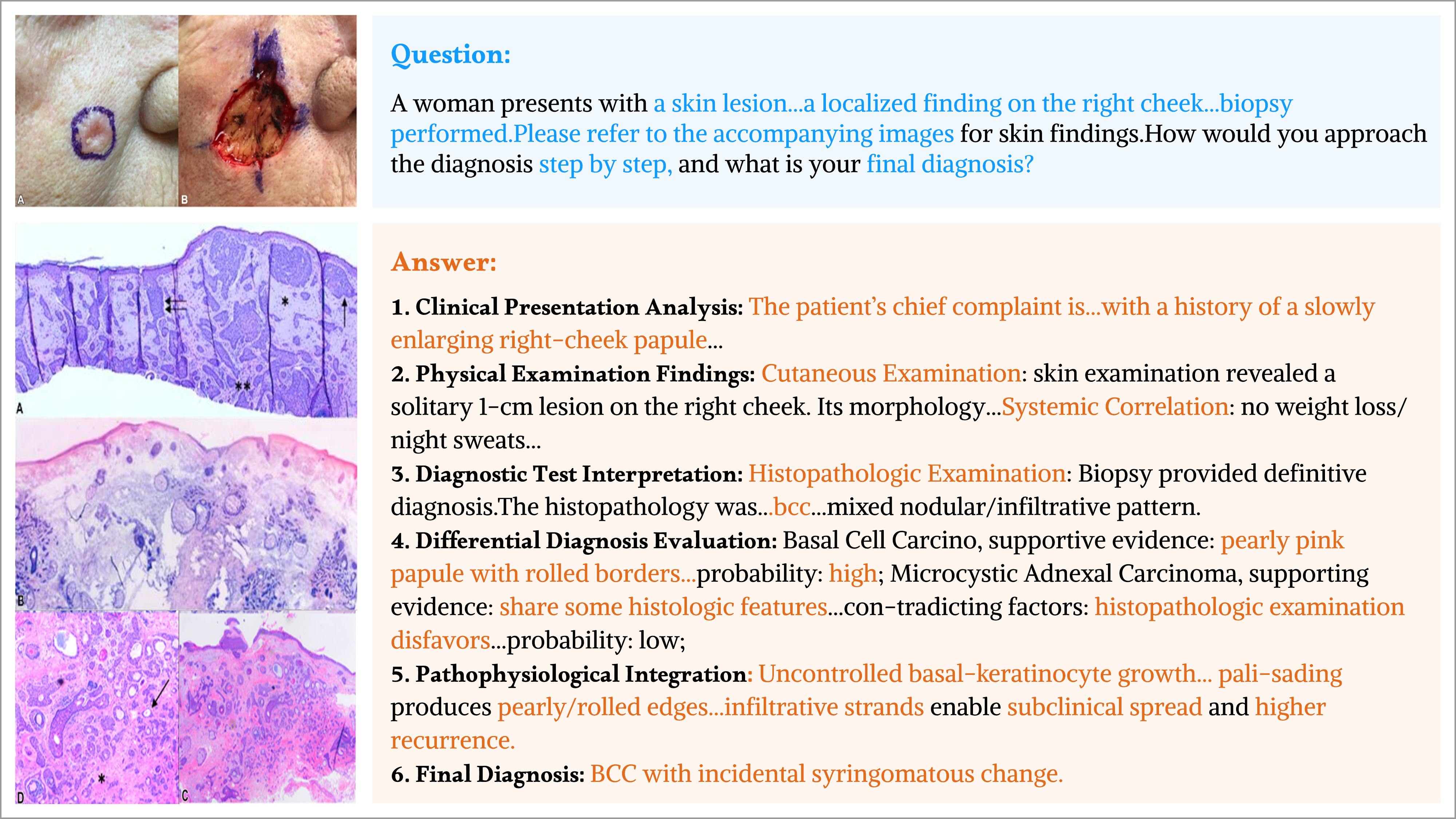}
    \caption{An example from DermCase. Each case comprises a clinical question with detailed context (patient history, laboratory results), multi-modal images, and step-wise diagnostic reasoning.}
    \label{fig:dataset_samole}
\end{figure}
\begin{table*}[t]

\caption{Comparison of existing dermatology benchmarks. Avg. Tokens indicates the average text length per data instance (\emph{e.g.}, case narrative, image caption, or question-answer pair). \dataset{} is the first benchmark providing full-length case narratives with multi-modal grounded images, enabling evaluation of diagnostic accuracy, differential diagnosis and multi-step clinical reasoning.}
\vspace{-8pt}
\begin{center}
\footnotesize
\resizebox{\textwidth}{!}{%
\begin{tabular}{@{}l rr c ccc c@{}}
\toprule
\hline
\multirow{3}{*}{\textbf{Dataset}} &
\multirow{3}{*}{\makecell[r]{\textbf{Num.\ of}\\\textbf{Samples}}} &
\multirow{3}{*}{\makecell[r]{\textbf{Avg. Tokens}\\\textbf{Per Data}}} &
\multirow{3}{*}{\makecell{\textbf{Data Type}}} &
\multicolumn{3}{c}{\textbf{Evaluation Perspectives}} &
\multirow{3}{*}{\textbf{Open Source}} \\
\cmidrule(lr){5-7}
& & & &
\makecell{\textbf{Diagnosis}\\\textbf{Accuracy}} &
\makecell{\textbf{Differential}\\\textbf{Diagnosis}} &
\makecell{\textbf{Rubrics\mbox{-}based}\\\textbf{Reasoning}} & \\
\midrule
\textbf{DermNet} \cite{goel2020dermnet} & 19{,}500 & -- & Image only & \Yes & \No & \No & \Yes \\ 
\textbf{SCIN} \cite{scin_dataset} & 10{,}408 & 10 & Image-caption pair & \Yes & \No & \No & \Yes \\
\textbf{EVAL\mbox{-}GPT\mbox{-}Derm} \cite{ferreira2023evaluation} & 93 & 10 & Text only & \Yes & \No & \No & \No \\
\textbf{Pillai et al.} \cite{pillai2024evaluating} & 102 & 50 & Image-caption pair & \Yes & \No & \No & \No \\
\textbf{SkinGPT4\mbox{-}Clinical\mbox{-}Eval}~\cite{Zhou2024SkinGPT4} & 150 & 20 & Image-caption pair & \Yes & \No & \No & \No \\
\textbf{DermaVQA}~\cite{Yim_DermaVQA_MICCAI2024} & 1{,}488 & 40 & Image-caption pair & \No & \No & \No & \Yes \\
\textbf{SkinCAP}~\cite{zhou2024skincap} & 4{,}000 & 60 & Image-caption pair & \No & \No & \No & \Yes \\
\textbf{Derm1M} \cite{yanderm1m} & 1{,}029{,}761 & 41 & Image-caption pair & \Yes & \No & \No & \Yes \\
\midrule
\textbf{\dataset{} (Ours)} & 6{,}354 & 1{,}100 & \makecell{Image-caption pair\\long-context reasoning} & \Yes & \Yes & \Yes & \Yes \\
\bottomrule
\end{tabular}
}
\end{center}
\label{tab:compare_with_other_benchmarks}
\end{table*}

Timely and accurate diagnosis of skin diseases is crucial for global public health, affecting an estimated 1.8 billion people worldwide and ranking as the 4th leading cause of nonfatal disease burden~\cite{hay2014global}. The challenge is particularly acute in resource-limited settings with only 0-3 dermatologists per million population~\cite{freeman2023global,tiwari2022counting}, where delayed diagnosis leads to serious clinical consequences~\cite{Seth2017GlobalBO,WHO_2023_skin_NTDs_meeting}. 
Recent large vision-language models (LVLMs) have shown promising capabilities on common dermatological conditions: SkinGPT-4 integrated vision transformers with Llama-2 for interactive diagnosis~\cite{Zhou2024SkinGPT4}, GPT-4V reached 89\% accuracy on 9 common conditions~\cite{pillai2024evaluating} and MedGemma achieved 71.8\% accuracy on 79-class classification~\cite{sellergren2025medgemma}. 
These systems demonstrate dermatologist-level performance on specific tasks, suggesting potential as clinical decision support tools.



However, two critical limitations constrain clinical translation. First, existing benchmarks predominantly evaluate common conditions, whereas rare diseases that demand sophisticated differential diagnosis and clinical reasoning remain underexplored. Second, evaluations assess only final accuracy, ignoring the reasoning steps. This gap is concerning: studies reveal that 65\% of incorrect diagnoses contain logic errors versus 18\% of correct ones~\cite{Savage_2024_DxReasoningPrompts_npjDM}, and structured analytical reasoning significantly improves outcomes~\cite{Moell_2025_DeepSeekR1_FrontAI}. While Chain-of-Thought prompting can decompose complex tasks, the clinical validity of such reasoning chains remains unvalidated in dermatology. 
The lack of datasets evaluating diagnostic reasoning has fundamentally limited our understanding of whether enhanced reasoning can improve performance in rare dermatological cases.

To address this critical gap, we present \dataset{}, a long-context dermatology dataset for evaluating diagnostic reasoning on rare cases, with an example shown in Figure~\ref{fig:dataset_samole}.
We derive \dataset{} from publicly licensed case reports extracted from peer-reviewed dermatology journals using a five-stage pipeline with systematic quality control. The dataset comprises 6,354 unique cases with complete diagnostic reasoning chains, encompassing 26,030 high-quality multi-modal dermatological image-caption pairs across diverse diseases. Each case includes comprehensive clinical information: patient history and presentation, multi-modal imaging (clinical photographs and histopathology), physical examination findings, diagnostic test results, differential diagnoses, and step-by-step clinical reasoning.
We systematically evaluate LVLMs across three dimensions: final diagnosis accuracy; differential diagnosis quality;
and rubric-based evaluation of stepwise clinical reasoning. 


Our contributions are threefold:

\begin{itemize}
\item \textbf{\dataset{} Dataset:} We introduce the first long-context dermatology dataset for diagnostic reasoning evaluation, comprising 26,030 multi-modal image-caption pairs and 6,354 rare cases with complete reasoning chains.

\item \textbf{DermLIP-based Evaluation Metrics:} We propose novel similarity metrics leveraging the pretrained DermLIP for dermatological differential diagnosis evaluation. Dermatologist validation confirms significantly stronger alignment with the expert compared to traditional metrics.

\item \textbf{Benchmark and Fine-tuning Analysis:} We evaluate 22 LVLMs across diagnosis accuracy, differential diagnosis, and reasoning quality. Supervised fine-tuning (SFT) on \dataset{} yields substantial improvements, while Direct Preference Optimization (DPO) shows limited gains. Systematic failure analysis identifies key reasoning limitations across model families.

\end{itemize}
\begin{figure*}[thbp]
    \centering
    \includegraphics[width=\textwidth]{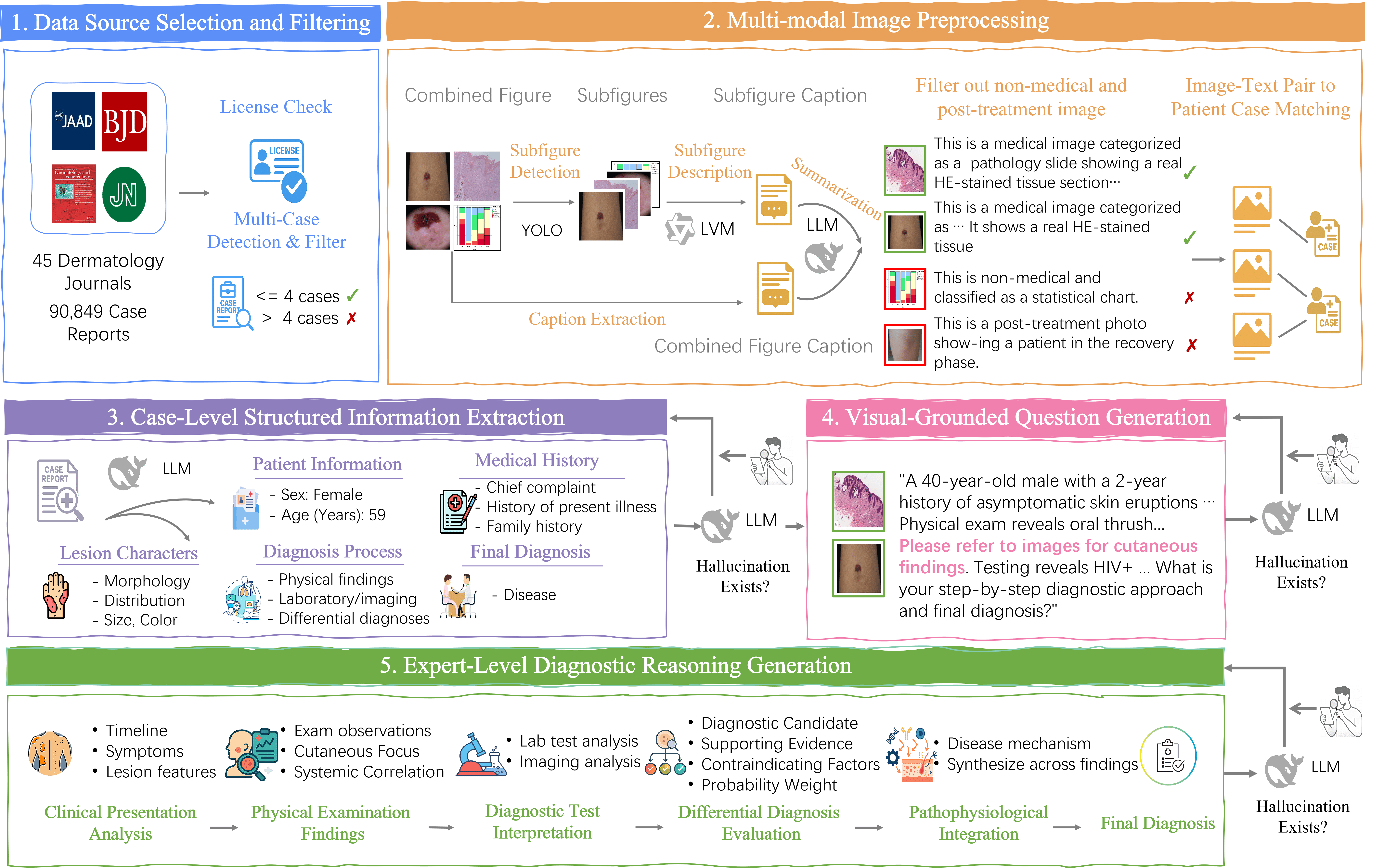}
    \caption{The dataset curation pipeline includes: data source selection and filtering, multi-modal image preprocessing, case-level information extraction, visual-grounded question generation, and expert-level diagnostic reasoning generation. Quality control (steps 3-5) combines LLM-based cross-verification and human sampling validation against source reports, triggering regeneration for inconsistencies.}
    \label{fig:dataset_curation}
    \vspace{-4mm}
\end{figure*}

\section{Related Work}
\label{sec:related_work}

We review vision-language models applied to dermatology, advances in multimodal reasoning capabilities, and existing dermatological datasets, highlighting key gaps that motivate our work.

\noindent \textbf{Vision-Language Models for Dermatology.}
Recent dermatological AI has advanced across multiple directions. Domain-specific foundation models established strong baselines: PanDerm achieved state-of-the-art results on 28 dermatology benchmarks spanning classification, segmentation, and retrieval through self-supervised pretraining~\cite{Yan_2025_PanDerm_NatMed}, while DermLIP excelled at image-text retrieval and zero-shot classification via contrastive learning on the million-scale Derm1M dataset~\cite{yanderm1m}. 
General-purpose LVLMs showed promising performance on common conditions, with GPT-4V reaching 89\% accuracy on nine frequent dermatological diseases~\cite{pillai2024evaluating}. Medical-specialized models including Med-Flamingo~\cite{moor2023medflamingo}, LLaVA-Med~\cite{li2023llavamed}, and Med-PaLM~\cite{tu2024towards}, fine-tuned on clinical data, demonstrated broad medical diagnostic capabilities, though dermatology-specific evaluation remains limited. SkinGPT-4 enabled interactive dermatological diagnosis through targeted fine-tuning~\cite{Zhou2024SkinGPT4}, yet focused primarily on common conditions. Critically, no existing work has systematically evaluated LVLMs on rare dermatological cases, which present greater diagnostic challenges due to underrepresentation in the training data and clinical complexity.

\noindent \textbf{Reasoning Capabilities in Vision-Language Models}
Beyond accuracy, step-wise clinical reasoning is crucial for trustworthy medical AI. Chain-of-thought (CoT) prompting has shown substantial gains by eliciting intermediate reasoning~\cite{wei2022chain,nachane2024fewshot}, motivating recent LVLMs to enhance reasoning through architectural and training innovations. InternVL2.5~\cite{wang2025enhancingreasoningabilitymultimodal} and InternVL3~\cite{zhu2025internvl3exploringadvancedtraining} employ Mixed Preference Optimization (MPO) combining DPO~\cite{rafailov2023direct} and Binary Classifier Optimization (BCO)~\cite{jung2025binaryclassifieroptimizationlarge} for multimodal CoT; InternVL3.5 uses Cascade RL for refined alignment~\cite{internvl3_5}; Qwen3 integrates adaptive thinking budgets for multi-step reasoning~\cite{qwen3technicalreport}; and MedGemma demonstrates medical-specific reasoning via continued medical-domain pretraining and downstream fine-tuning~\cite{sellergren2025medgemma}. 
However, systematic evaluation of visually grounded diagnostic reasoning on rare dermatological cases remains absent, leaving unclear whether current LVLMs can reliably reason on such challenging scenarios.

\noindent \textbf{Dermatology Vision-Language Datasets.}
Classical dermatology datasets such as ISIC~\cite{gutman2016skin}, HAM10000~\cite{tschandl2018ham10000}, and PAD-UFES-20~\cite{pacheco2020pad} focus on image classification tasks. Recent work has introduced image-text pairs: SkinCAP~\cite{zhou2024skincap} curates 4k clinical images with expert captions; MM-Skin contributes 10k textbook pairs with 27k VQA samples~\cite{zeng2025mm}; Derm1M spans over 390 conditions with 1 million ontology-aligned pairs~\cite{yanderm1m}; and DermaVQA~\cite{Yim_DermaVQA_MICCAI2024} offers multilingual question-answer pairs from online consultations. However, acquiring complete diagnostic processes with detailed reasoning is challenging, and no existing dataset (Table~\ref{tab:compare_with_other_benchmarks}) captures reasoning chains linking visual observations, clinical examinations, and laboratory results to diagnostic conclusions. 
Moreover, existing datasets predominantly cover common skin conditions, which risk overlapping with LVLM training data and fail to challenge true reasoning capabilities. Rare diseases, with their complex presentations and limited clinical exposure, provide stronger evaluation of diagnostic reasoning.
Our proposed \dataset{} dataset addresses these gaps with 6,354 long-context rare cases featuring comprehensive clinical context,  differential diagnoses and step-wise reasoning.
\section{Method}
We present a framework with a data curation pipeline for high-quality clinical reasoning QA pairs and metrics evaluating diagnostic reasoning beyond conventional accuracy. The LLM we adopt in this pipeline is \llm{}.


\subsection{Dataset Curation Pipeline}
To obtain high-quality reasoning question-answering pairs for rare dermatological conditions, we design a multi-stage dataset curation pipeline. As illustrated in Figure~\ref{fig:dataset_curation}, our pipeline comprises five primary steps.

\noindent \textbf{Step 1: Data Source Selection and Filtering.}
Rare dermatological conditions demand a comprehensive clinical context beyond visual inspection. Accurate diagnosis requires integrating multi-modal information, including patient history, laboratory results, clinical photographs, and histopathological images, which clinicians synthesize through systematic reasoning to formulate differential diagnoses.
Given the scarcity of publicly available data with detailed clinical narratives and diagnostic reasoning, we curate a high-quality dataset from authoritative sources. We collected 90,849 case reports from 45 peer-reviewed dermatology journals. We then filter for permissive licenses and employ an LLM to exclude articles containing more than four cases, ensuring sufficient clinical context per case. Each case is extracted independently with its complete clinical narrative. After filtering, our dataset comprises 6,354 unique clinical cases with rich multi-modal information. Details can be found in \autoref{supp:dataset_details}.

\noindent \textbf{Step 2: Multi-modal Image Preprocessing.} We employ the YOLO~\cite{khanam2024yolov11overviewkeyarchitectural} detector to segment multi-panel figures. To ensure quality, Qwen3-VL~\cite{qwen3technicalreport} generates image descriptions, which an LLM then processes into structured annotations covering clinical relevance, treatment stage, and content. Non-medical (e.g., charts) and post-treatment images are filtered, retaining only clinical, histopathological, and diagnostic (dermoscopy, radiology) photographs. For multi-case articles, subfigures are matched to cases using contextual descriptions and narratives to ensure data integrity.

\noindent \textbf{Step 3: Case-Level Structured Information Extraction.} 
We extract clinical data from case reports across five aspects: patient age/gender, medical history, lesion characteristics, diagnostic process, and final diagnosis. 
To mitigate hallucination, we implement a two-stage verification process where the LLM extracts information from direct quotations and self-verifies against the source text, supplemented by human validation, with re-extraction for inconsistencies.

\noindent \textbf{Step 4: Visual-Grounded Question Generation.} 
To ensure genuine visual reasoning, we adopt a visual-grounding strategy that omits lesion descriptions from questions, directing models to extract visual features directly from images (\emph{e.g.,} ``Please refer to images for cutaneous findings"). This design prevents data leakage and evaluates true vision-language capabilities. We generate questions with patient demographics, medical history, non-lesion findings, and diagnostic results, requesting step-by-step reasoning toward final diagnosis. 
LLM performs verification to detect hallucinations and trigger automatic regeneration when needed, coupled with human validation on sampled cases.

\noindent \textbf{Step 5: Expert-Level Diagnostic Reasoning Generation.}
Using Step 3 data, LLM generates diagnostic reasoning across five steps: clinical presentation analysis, physical examination synthesis, diagnostic test interpretation, differential diagnosis, and pathophysiological integration, emphasizing narrative coherence via transitional phrases and mechanism-based explanations. To ensure clinical fidelity, the LLM cross-verifies against source reports---supplemented by human spot-checking---triggering regeneration for inconsistencies.

\begin{figure}[t]
    \centering
    \begin{subfigure}[t]{0.28\textwidth}
        \includegraphics[width=\linewidth]{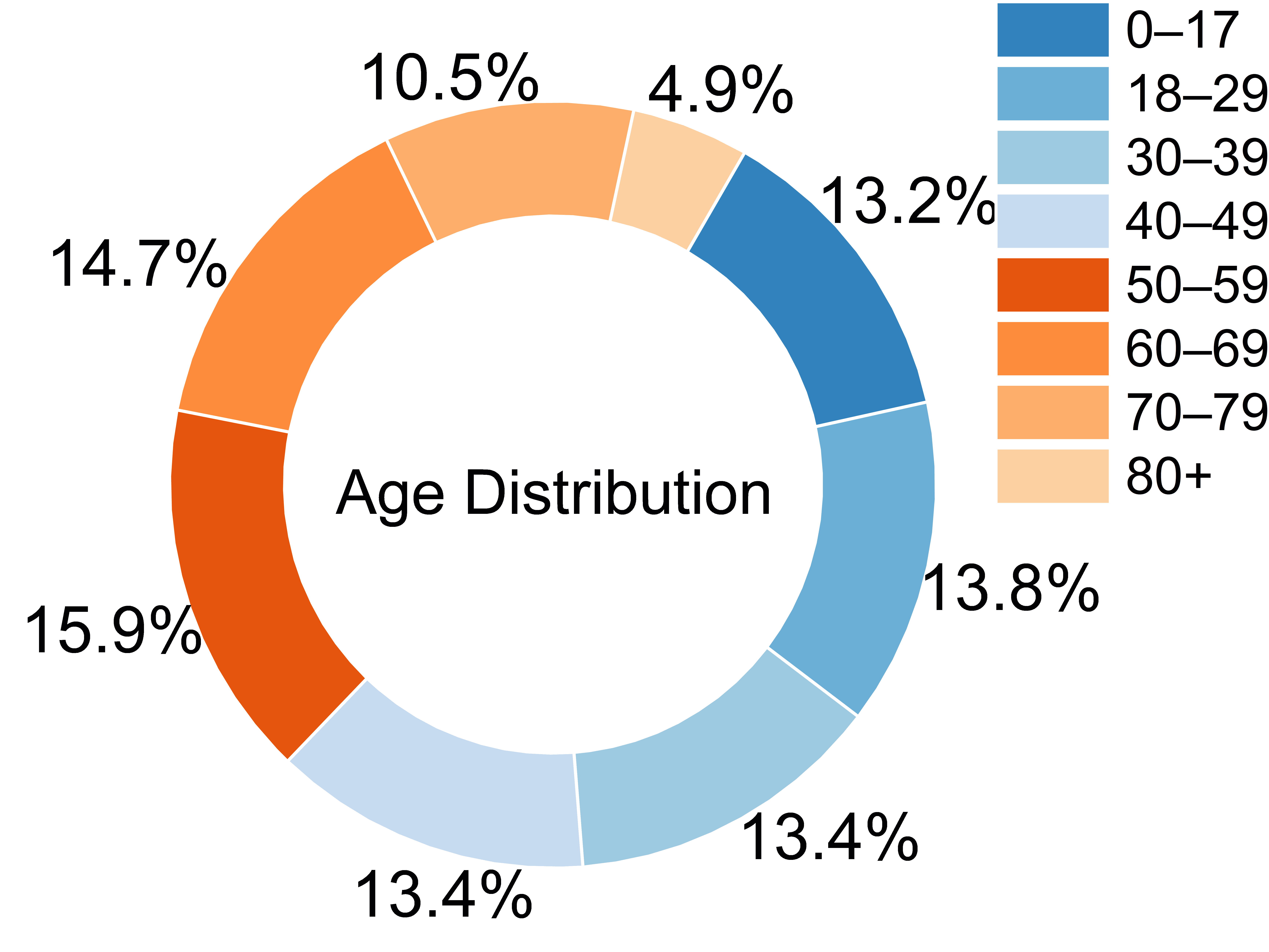}
        \caption{Age distribution}
        \label{fig:age_distribution}
    \end{subfigure}
    \hfill
    \begin{subfigure}[t]{0.28\textwidth}
        \includegraphics[width=\linewidth]{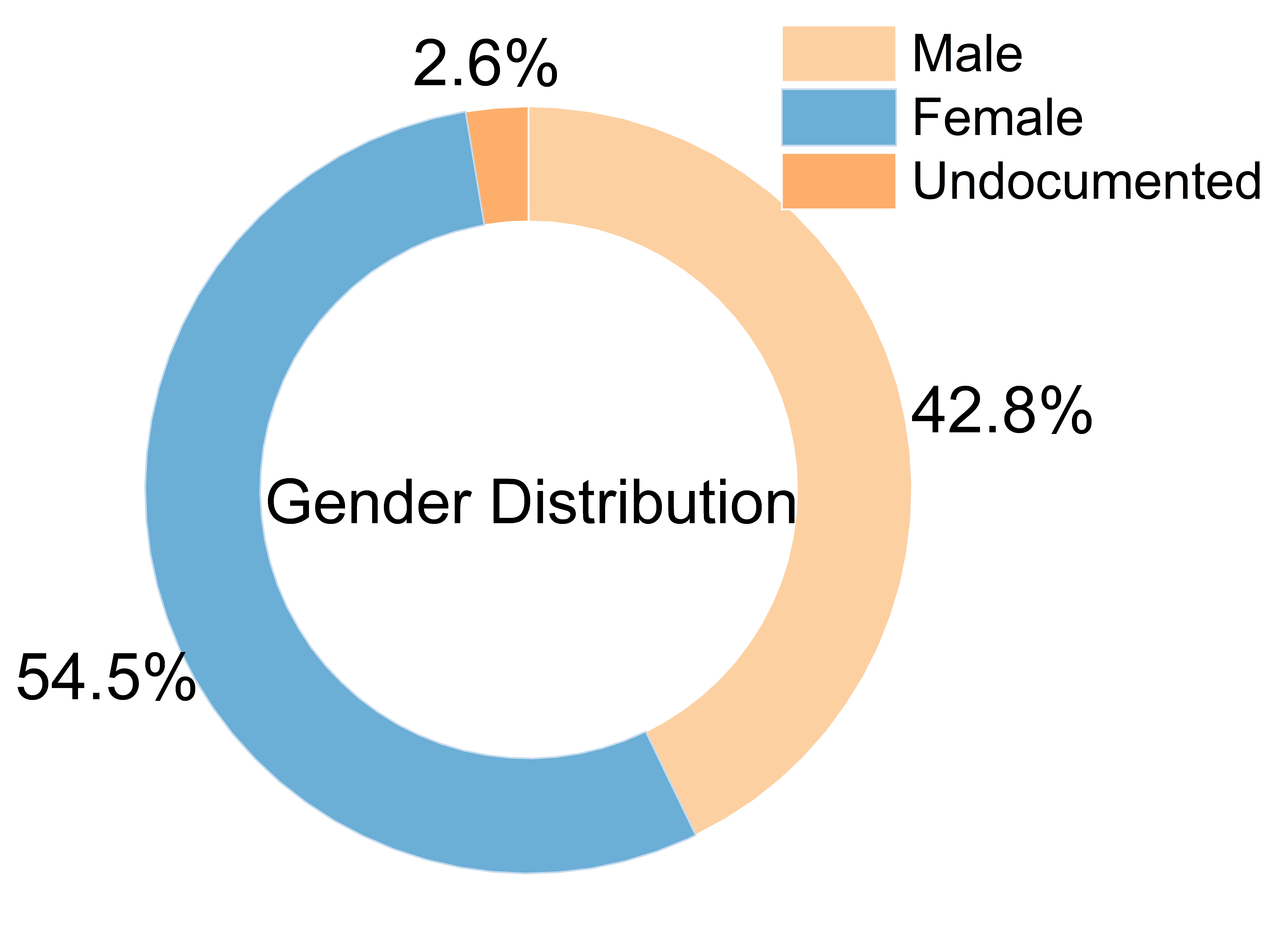}
        \caption{Gender distribution}
        \label{fig:gender_distribution}
    \end{subfigure}
    \hfill
    \begin{subfigure}[t]{0.4\textwidth}
        \includegraphics[width=\linewidth]{images/distribution.png}
        \caption{Top 10 disease by frequency}        \label{fig:disease_distribution}
    \end{subfigure}
    \caption{Statistical overview of the DermCase dataset. (a) Age distribution peaking at 50–59 years (15.9\%). (b) Female-dominant gender distribution (54.5\%). (c) Top 10 diseases, led by Basal Cell Carcinoma and Squamous Cell Carcinoma.}
    \label{fig:dataset_distribution_simple}
\end{figure}

\noindent \textbf{Discussion on Data Quality and Expert Validation.}
A key concern in LLM-assisted dataset construction is hallucination and clinical fidelity. We emphasize that the diagnostic reasoning chains in \dataset{} are grounded in peer-reviewed case reports rather than LLM-generated content: LLMs serve solely to extract and structure information, with cross-verification against source texts at each stage. To validate pipeline quality, a board-certified dermatologist reviewed 50 sampled cases across Steps 3--5, yielding pass rates of 94\% (extraction accuracy), 86\% (medically grounded questions), and 90\% (clinically sound reasoning chains), confirming that the pipeline reliably preserves clinical logic from the source literature while maintaining high standards of medical integrity.

One may question whether a 50-case expert review provides sufficient validation coverage. We clarify that \textit{this evaluation is designed to assess pipeline fidelity rather than per-sample clinical correctness}, as all diagnostic conclusions and supporting evidence originate from published peer-reviewed case reports. The validation objective is therefore to confirm that the extraction process faithfully preserves clinical intent, rather than to re-adjudicate each diagnosis independently. Unlike datasets consisting of simple image descriptions or short-form answers, evaluating long-context diagnostic reasoning demands careful scrutiny of multi-step clinical logic, where review depth is considerably more informative than review breadth. Moreover, the depth of each case makes exhaustive review impractical in practice: each reasoning chain spans approximately 1,100 tokens and involves intricate multi-step diagnostic logic, requiring 30 to 45 minutes of expert review time per case. Our validation strategy thus follows the established practice of medical benchmarks such as MedQA~\cite{jin2021disease} and PubMedQA~\cite{jin2019pubmedqa}, which validate annotation protocols on carefully selected representative subsets rather than conducting exhaustive case-by-case review.

\noindent \textbf{Dataset Statistics.}
\dataset{} contains 6,354 unique cases with diverse demographics: balanced gender, age spanning 0-80+ years (Figure~\ref{fig:dataset_distribution_simple}), and broad ethnic/racial representation across geographic regions. The dataset encompasses 2,101 dermatological conditions, with the 10 most frequent shown in Figure~\ref{fig:disease_distribution} (detailed statistics in Appendix~\ref{supp:data_distribution}).

\subsection{Evaluation Metrics for Diagnosis Accuracy}
We employ LLM to extract predicted diagnoses from model responses (see Appendix~\ref{supp_fig:final_extraction_prompt}) and evaluate diagnostic accuracy. Traditional metrics (e.g. BERTScore) inadequately assess diagnosis accuracy due to lack of medical domain knowledge: semantically similar terms may represent distinct conditions (\emph{e.g.}, ``psoriasis" vs. ``eczema"), while different expressions may denote the same disease (\emph{e.g.}, ``atopic dermatitis" and ``eczema"). We therefore employ LLM as a judge (see Appendix~\ref{supp_fig:final_evaluation}) to determine semantic equivalence between predictions and ground truth, computing accuracy from these clinically meaningful judgments.

\begin{table*}[htbp]
\centering
\caption{Benchmark on final diagnosis and differential diagnosis. Differential diagnosis is assessed by coverage and our proposed DermLIP-based metrics (D-Precision, Recall, F1, and Jaccard). \textbf{\textcolor{red}{Best}} and \textbf{\textcolor{blue}{second-best}} are highlighted in red and blue, respectively.}
\label{tab:differential_diagnosis_83}
\resizebox{\textwidth}{!}{%
\begin{tabular}{l cc cccc cccc}
\toprule
\multirow[c]{3}{*}{\raisebox{-2ex}{\textbf{Methods}}} & \multicolumn{1}{c}{\textbf{Final Diagnosis}} & \multicolumn{9}{c}{\textbf{Differential Diagnosis}} \\
\cmidrule(lr){2-2} \cmidrule(lr){3-11}
& \multirow[c]{2}{*}{\raisebox{-1ex}{\textbf{Accuracy}}} & \multirow[c]{2}{*}{\raisebox{-1ex}{\textbf{Coverage}}} & \multicolumn{4}{c}{\textbf{Macro}} & \multicolumn{4}{c}{\textbf{Micro}} \\
\cmidrule(lr){4-7} \cmidrule(lr){8-11}
& & & \textbf{D-Precision} & \textbf{D-Recall} & \textbf{D-F1} & \textbf{D-Jaccard} & \textbf{D-Precision} & \textbf{D-Recall} & \textbf{D-F1} & \textbf{D-Jaccard} \\
\midrule
\multicolumn{11}{c}{\textbf{Flamingo Series}} \\
\hline
Med-Flamingo~\cite{moor2023medflamingo} & 0.1098 & 1.0000 & 0.6544 & 0.1875 & 0.2676 & 0.1680 & 0.4742 & 0.1814 & 0.2624 & 0.1510 \\
\hline
\multicolumn{11}{c}{\textbf{InternVL Series}} \\
\hline
InternVL2.5-2B~\cite{internvl} & 0.0616 & 0.8952 & 0.3597 & 0.2734 & 0.2886 & 0.1885 & 0.3269 & 0.2692 & 0.2953 & 0.1732 \\
InternVL2.5-2B-MPO~\cite{wang2025enhancingreasoningabilitymultimodal} & 0.0428 & 0.9439 & 0.3463 & 0.2904 & 0.2991 & 0.1986 & 0.3232 & 0.2856 & 0.3032 & 0.1787 \\
InternVL2.5-4B~\cite{internvl} & 0.1526 & 0.9140 & 0.4272 & 0.3528 & 0.3691 & 0.2492 & 0.3937 & 0.3448 & 0.3676 & 0.2252 \\
InternVL2.5-4B-MPO~\cite{wang2025enhancingreasoningabilitymultimodal} & 0.1017 & 0.9283 & 0.4078 & 0.3597 & 0.3583 & 0.2412 & 0.3701 & 0.3520 & 0.3608 & 0.2201 \\
InternVL2.5-8B~\cite{internvl} & 0.1218 & 0.9661 & 0.4329 & 0.3973 & 0.3918 & 0.2674 & 0.3910 & 0.3894 & 0.3902 & 0.2424 \\
InternVL2.5-8B-MPO~\cite{wang2025enhancingreasoningabilitymultimodal} & 0.1178 & 0.9739 & 0.3922 & 0.4129 & 0.3771 & 0.2539 & 0.3492 & 0.4051 & 0.3751 & 0.2308 \\
InternVL3-8B-Instruct~\cite{internvl} & 0.1432 & 0.9478 & 0.4410 & 0.4184 & 0.4066 & 0.2831 & 0.3915 & 0.4095 & 0.4003 & 0.2502 \\
InternVL3.5-2B-Instruct~\cite{internvl3_5} & 0.0872 & 1.0000 & 0.5777 & 0.2797 & 0.3240 & 0.2124 & 0.4185 & 0.2735 & 0.3308 & 0.1982 \\
InternVL3.5-4B-Instruct~\cite{internvl3_5} & 0.1195 & 1.0000 & 0.5622 & 0.3025 & 0.3558 & 0.2399 & 0.4856 & 0.2930 & 0.3655 & 0.2236 \\
InternVL3.5-8B-Instruct~\cite{internvl3_5} & 0.1450 & 1.0000 & 0.5964 & 0.2947 & 0.3545 & 0.2382 & \textcolor{blue}{0.5109} & 0.2867 & 0.3673 & 0.2250 \\
\hline
\multicolumn{11}{c}{\textbf{LLaVA Series}} \\
\hline
LLaVA-1.5-7B~\cite{liu2023improvedllava} & 0.1138 & 0.6871 & 0.5667 & 0.2862 & 0.3501 & 0.2324 & 0.4855 & 0.2795 & 0.3547 & 0.2156 \\
LLaVA-1.6-Vicuna-7B~\cite{liu2023llava} & \textbf{\textcolor{blue}{0.2062}} & 1.0000 & \textcolor{blue}{0.6705} & 0.2279 & 0.2845 & 0.1778 & 0.3304 & 0.2229 & 0.2662 & 0.1535 \\
LLaVA-Med-1.5-Mistral-7B~\cite{li2023llavamed} & 0.1914 & 1.0000 & \textbf{\textcolor{red}{0.7503}} & 0.1790 & 0.2808 & 0.1747 & \textbf{\textcolor{red}{0.7127}} & 0.1718 & 0.2768 & 0.1606 \\
\hline
\multicolumn{11}{c}{\textbf{MiniCPM Series}} \\
\hline
MiniCPM-V4.5~\cite{yu2025minicpmv45cookingefficient} & 0.1383 & 1.0000 & 0.4186 & 0.4189 & 0.3798 & 0.2556 & 0.3469 & 0.4096 & 0.3757 & 0.2313 \\
\hline
\multicolumn{11}{c}{\textbf{Qwen Series}} \\
\hline
Qwen2-VL-2B-Instruct~\cite{Qwen2VL} & 0.1151 & 0.6767 & 0.4690 & 0.3011 & 0.3360 & 0.2268 & 0.3997 & 0.2933 & 0.3383 & 0.2036 \\
Qwen2.5-VL-3B-Instruct~\cite{Qwen2.5-VL} & 0.1151 & 0.6910 & 0.4492 & 0.3344 & 0.3531 & 0.2382 & 0.4012 & 0.3288 & 0.3614 & 0.2206 \\
Qwen2.5-VL-7B-Instruct~\cite{Qwen2.5-VL} & 0.1553 & 0.8827 & 0.4465 & 0.4049 & 0.3972 & 0.2722 & 0.4022 & 0.3954 & 0.3988 & 0.2490 \\
Qwen3-VL-4B-Instruct~\cite{qwen3technicalreport} & 0.1584 & 1.0000 & 0.4321 & 0.3825 & 0.3588 & 0.2430 & 0.3469 & 0.3714 & 0.3587 & 0.2186 \\
Qwen3-VL-8B-Instruct~\cite{qwen3technicalreport} & 0.1544 & 1.0000 & 0.4654 & 0.3683 & 0.3547 & 0.2399 & 0.3510 & 0.3569 & 0.3539 & 0.2150 \\
\hline
\multicolumn{11}{c}{\textbf{MedGemma Series}} \\
\hline
MedGemma-4B-it~\cite{sellergren2025medgemma} & 0.1299 & 0.9896 & 0.4009 & \textcolor{blue}{0.4877} & \textcolor{blue}{0.4204} & \textcolor{blue}{0.2896} & 0.3628 & \textcolor{blue}{0.4786} & \textcolor{blue}{0.4127} & \textcolor{blue}{0.2600} \\
MedGemma-27B-it~\cite{sellergren2025medgemma} & \textbf{\textcolor{red}{0.2610}} & 0.9909 & 0.3745 & \textbf{\textcolor{red}{0.6066}} & \textbf{\textcolor{red}{0.4400}} & \textbf{\textcolor{red}{0.3053}} & 0.3236 & \textbf{\textcolor{red}{0.5971}} & \textbf{\textcolor{red}{0.4198}} & \textbf{\textcolor{red}{0.2656}} \\
\bottomrule
\end{tabular}%
}
\end{table*}

\subsection{Evaluation Metrics for Differential Diagnosis}
Beyond final diagnosis, differential diagnosis plays a crucial role in clinical practice, where physicians consider multiple plausible conditions ranked by likelihood. This is particularly critical for rare dermatological diseases, where diagnostic uncertainty is high and the differential list helps narrow down possibilities, guide additional testing, and prevent missed diagnoses. 
To evaluate LVLM's differential diagnosis capability, we employ LLM to extract differential diagnoses from both model responses and ground truth annotations (see Appendix~\ref{supp:dd_extraction}).

Given $N$ total cases, for each of the case $i$, the predicted differential diagnosis set is $P_i = \{x_1, ..., x_{n_i}\}$ and ground truth non-empty set is $G_i = \{y_1, ..., y_{m_i}\}$. 
We define the valid set $\mathcal{V} = \{i : n_i > 0\}$ containing samples with non-empty predictions. We define the following metrics.

\noindent \textbf{Coverage Rate.} We define coverage rate as $|\mathcal{V}|/N$, the proportion of samples with valid differential diagnosis.

\noindent \textbf{Similarity based on DermLIP Score.} 
To quantify semantic similarity between dermatological terms, we leverage DermLIP~\cite{yanderm1m}, a vision-language foundation model fine-tuned on over 1 million dermatological image-text pairs, to extract embeddings. Unlike general-purpose models (\emph{e.g.}, BERT), DermLIP embeddings capture clinically meaningful semantic relationships, recognizing synonymous conditions (\emph{e.g.}, tinea corporis" v.s. ringworm") while distinguishing visually similar but clinically distinct diseases (\emph{e.g.}, psoriasis" v.s. eczema").

We encode each diagnosis term using DermLIP's text encoder with the clinical prefix ``a clinical dermatology photo of [term]'', which grounds the text embeddings in the visual-semantic space learned during pretraining. 
Each term is represented as an $\ell_2$-normalized embedding $\mathbf{e} \in \mathbb{R}^{512}$, and the raw cosine similarity is:
\begin{equation}
\text{DermLIP}(x_k, y_j) = \mathbf{e}_{x_k}^\top \mathbf{e}_{y_j} 
\end{equation}

Since DermLIP produces scores in the empirical range $[s_{\min}, s_{\max}] = [0.6, 1.0]$ for dermatological terms (see distribution in Appendix~\ref{supp:dermlip_distribution}), we apply min-max normalization to $[0, 1]$ to ensure proper similarity interpretation.

\noindent \textbf{D-Precision, Recall, F1, Jaccard.}
We construct the similarity matrix $\mathbf{S}_i \in \mathbb{R}^{n_i \times m_i}$ for sample $i$, where $S_{kj} = s(x_k, y_j) \in \mathbf{S}_i$.
To ensure clinically meaningful matches, we require a semantic similarity threshold $\tau$ to distinguish clinical equivalence or near-synonymy from weak semantic relatedness. The candidate match set $C_i$ is defined as:
\begin{equation}
C_i = \{(k, j) \mid S_{kj} \geq \tau, \, k \in [1, n_i], \, j \in [1, m_i]\}
\end{equation}
This ensures only highly similar terms representing clinical synonyms (\emph{e.g.}, ``tinea corporis" and ``ringworm") are considered matches, while excluding moderately similar but diagnostically distinct conditions (\emph{e.g.}, ``psoriasis" and ``eczema").
We then apply greedy bipartite matching to obtain the matched set $\mathcal{M}_i$ with one-to-one correspondence, aligning with clinical practice where each differential consideration represents a distinct condition.
Details in Appendix~\ref{supp:algo}.
We report both macro-averaged and micro-averaged D-Precision, Recall, F1, Jaccard scores (see Appendix~\ref{supp:metrics}) across all valid samples.

\begin{table*}[t]
\centering
\caption{Rubrics-based reasoning evaluation. 22 LVMs across five dimensions: Medical Knowledge (0-5), Image Interpretation (0-25), Laboratory Interpretation (0-15), Differential Diagnosis (0-25), and Clinical Reasoning (0-30). Mean ± Std are reported.}
\label{tab:weighted_scores}
\resizebox{\textwidth}{!}{%
\begin{tabular}{lcccccc}
\hline
\rule{0pt}{3ex}\large Model & Medical Knowledge & Image Interpretation & Laboratory Interpretation & Differential Diagnosis & Clinical Reasoning & Total \\
 & (0-5) & (0-25) & (0-15) & (0-25) & (0-30) & (0-100) \\[0.5ex]
\hline
\multicolumn{7}{c}{\textbf{LLaVA Series}} \\
\hline
Llava-1.5-7B~\cite{liu2023improvedllava} & 1.96 $\pm$ 0.85 & 5.68 $\pm$ 4.36 & 5.41 $\pm$ 3.43 & 6.21 $\pm$ 1.76 & 7.91 $\pm$ 2.20 & 27.02 $\pm$ 8.20 \\
Llava-1.6-Vicuna-7B~\cite{liu2023llava} & 1.45 $\pm$ 1.65 & 3.24 $\pm$ 6.10 & 4.03 $\pm$ 5.14 & 4.41 $\pm$ 5.60 & 5.47 $\pm$ 7.09 & 18.53 $\pm$ 23.02 \\
Llava-Med-1.5-Mistral-7B~\cite{li2023llavamed} & 1.99 $\pm$ 1.34 & 4.05 $\pm$ 6.78 & 5.45 $\pm$ 4.66 & 6.51 $\pm$ 5.37 & 8.63 $\pm$ 6.65 & 26.53 $\pm$ 22.23 \\
\hline
\multicolumn{7}{c}{\textbf{Med-Flamingo}} \\
\hline
Med-Flamingo~\cite{moor2023medflamingo} & 1.47 $\pm$ 1.33 & 3.19 $\pm$ 6.18 & 3.18 $\pm$ 4.48 & 4.86 $\pm$ 5.51 & 6.59 $\pm$ 6.44 & 19.26 $\pm$ 21.47 \\
\hline
\multicolumn{7}{c}{\textbf{InternVL Series}} \\
\hline
InternVL2.5-2B~\cite{internvl} & 1.68 $\pm$ 0.71 & 5.58 $\pm$ 3.99 & 5.25 $\pm$ 2.65 & 6.49 $\pm$ 1.27 & 7.95 $\pm$ 1.82 & 26.85 $\pm$ 7.00 \\
InternVL2.5-2B-MPO~\cite{wang2025enhancingreasoningabilitymultimodal} & 1.57 $\pm$ 0.63 & 5.48 $\pm$ 3.77 & 4.73 $\pm$ 2.26 & 6.45 $\pm$ 1.13 & 7.79 $\pm$ 1.50 & 25.93 $\pm$ 6.07 \\
InternVL2.5-4B~\cite{internvl} & 2.24 $\pm$ 0.94 & 6.76 $\pm$ 4.27 & 6.68 $\pm$ 2.99 & 7.36 $\pm$ 2.43 & 9.26 $\pm$ 3.27 & 32.14 $\pm$ 10.11 \\
InternVL2.5-4B-MPO~\cite{wang2025enhancingreasoningabilitymultimodal} & 2.25 $\pm$ 0.94 & 7.08 $\pm$ 4.19 & 6.73 $\pm$ 2.98 & 7.34 $\pm$ 2.42 & 9.22 $\pm$ 3.27 & 32.45 $\pm$ 10.31 \\
InternVL2.5-8B~\cite{internvl} & 2.40 $\pm$ 0.93 & 7.47 $\pm$ 4.26 & 7.06 $\pm$ 3.06 & 8.14 $\pm$ 2.99 & 9.81 $\pm$ 3.75 & 34.70 $\pm$ 11.46 \\
InternVL2.5-8B-MPO~\cite{wang2025enhancingreasoningabilitymultimodal} & 2.37 $\pm$ 0.96 & 7.94 $\pm$ 4.03 & 7.00 $\pm$ 3.05 & 7.98 $\pm$ 2.85 & 9.80 $\pm$ 3.70 & 34.89 $\pm$ 11.44 \\
InternVL3-8B-Instruct~\cite{internvl} & 2.43 $\pm$ 0.99 & 8.32 $\pm$ 4.62 & 7.44 $\pm$ 3.30 & 8.36 $\pm$ 3.25 & 10.25 $\pm$ 4.11 & 36.63 $\pm$ 12.73 \\
InternVL3.5-2B-Instruct~\cite{internvl3_5} & 2.05 $\pm$ 0.96 & 7.29 $\pm$ 4.24 & 6.13 $\pm$ 3.26 & 6.75 $\pm$ 2.60 & 8.80 $\pm$ 3.38 & 30.87 $\pm$ 10.55 \\
InternVL3.5-4B-Instruct~\cite{internvl3_5} & 2.43 $\pm$ 1.04 & 9.59 $\pm$ 5.46 & 7.32 $\pm$ 3.64 & 7.89 $\pm$ 3.53 & 10.47 $\pm$ 4.81 & 37.52 $\pm$ 14.86 \\
InternVL3.5-8B-Instruct~\cite{internvl3_5} & 2.51 $\pm$ 1.06 & 8.32 $\pm$ 5.42 & 7.60 $\pm$ 3.71 & 7.50 $\pm$ 2.88 & 10.66 $\pm$ 4.51 & 36.43 $\pm$ 13.55 \\
\hline
\multicolumn{7}{c}{\textbf{Qwen Series}} \\
\hline
Qwen2-VL-2B-Instruct~\cite{Qwen2VL} & 1.87 $\pm$ 0.80 & 4.23 $\pm$ 4.37 & 5.53 $\pm$ 2.58 & 6.24 $\pm$ 1.51 & 7.96 $\pm$ 1.99 & 25.69 $\pm$ 7.53 \\
Qwen2.5-VL-3B-Instruct~\cite{Qwen2.5-VL} & 2.19 $\pm$ 0.94 & 6.86 $\pm$ 4.48 & 6.84 $\pm$ 3.11 & 6.74 $\pm$ 1.80 & 9.12 $\pm$ 3.16 & 31.58 $\pm$ 9.52 \\
Qwen2.5-VL-7B-Instruct~\cite{Qwen2.5-VL} & 2.54 $\pm$ 0.99 & 9.18 $\pm$ 4.54 & 7.47 $\pm$ 3.39 & 8.17 $\pm$ 3.05 & 10.32 $\pm$ 4.16 & 37.49 $\pm$ 12.41 \\
Qwen3-VL-4B-Instruct~\cite{qwen3technicalreport} & 2.77 $\pm$ 1.08 & 10.84 $\pm$ 5.17 & 8.15 $\pm$ 3.91 & 8.65 $\pm$ 3.52 & 11.56 $\pm$ 5.00 & 41.78 $\pm$ 15.58 \\
Qwen3-VL-8B-Instruct~\cite{qwen3technicalreport} & \textcolor{blue}{2.88 $\pm$ 1.18} & \textcolor{blue}{11.66 $\pm$ 5.34} & \textcolor{blue}{8.68 $\pm$ 4.11} & \textcolor{blue}{9.04 $\pm$ 3.81} & \textcolor{blue}{12.30 $\pm$ 5.73} & \textcolor{blue}{44.38 $\pm$ 17.42} \\
\hline
\multicolumn{7}{c}{\textbf{MiniCPM Series}} \\
\hline
MiniCPM-V4.5~\cite{yu2025minicpmv45cookingefficient} & 2.63 $\pm$ 1.07 & 10.64 $\pm$ 4.87 & 8.00 $\pm$ 3.76 & 8.91 $\pm$ 3.80 & 11.36 $\pm$ 5.58 & 41.32 $\pm$ 16.36 \\
\hline
\multicolumn{7}{c}{\textbf{Medgemma Series}} \\
\hline
MedGemma-4B-it~\cite{sellergren2025medgemma} & 2.49 $\pm$ 0.96 & 9.04 $\pm$ 4.29 & 6.95 $\pm$ 3.55 & 8.96 $\pm$ 3.60 & 10.33 $\pm$ 4.30 & 37.57 $\pm$ 13.52 \\
MedGemma-27B-it~\cite{sellergren2025medgemma} & \textbf{\textcolor{red}{3.28 $\pm$ 1.13}} & \textbf{\textcolor{red}{13.73 $\pm$ 5.88}} & \textbf{\textcolor{red}{10.11 $\pm$ 4.05}} & \textbf{\textcolor{red}{13.17 $\pm$ 6.16}} & \textbf{\textcolor{red}{16.03 $\pm$ 8.32}} & \textbf{\textcolor{red}{56.13 $\pm$ 23.42}} \\
\hline
\end{tabular}
}
\end{table*}

\subsection{Rubrics Based Evaluation for Reasoning Steps}

To assess diagnostic reasoning quality, we design a structured evaluation rubric spanning five weighted dimensions: (1) Medical Knowledge: accuracy of pathophysiology, disease classification, and terminology; (2) Image Interpretation: quality of lesion feature descriptions including morphology, color, texture, and distribution; (3) Laboratory Interpretation: accurate interpretation of test results and their clinical significance; (4) Differential Diagnosis Reasoning: coverage of candidate diagnoses with supporting and opposing evidence; (5) Clinical Reasoning Chain: logical integration of multi-modal findings into coherent diagnostic pathways. We employ an LLM-based automated judge to evaluate model responses against expert ground truth, enabling fine-grained analysis of clinical reasoning capabilities. Details of the rubric are provided in Appendix~\ref{supp:rubric}.

\vspace{-2mm}
\section{Experiment}
We benchmark 18 general-purpose models and 4 medical-specialized variants.
We also apply SFT and DPO~\cite{rafailov2023direct} to investigate performance improvements, validate alignment between the DermLIP Score and dermatologist judgments, and analyze failure patterns across model families.

\subsection{Benchmark on Final Diagnosis}
Final diagnosis accuracy in Table~\ref{tab:differential_diagnosis_83} reveals the complexity of rare dermatological cases. MedGemma's performance drops dramatically from 71.8\% accuracy on common cases~\cite{sellergren2025medgemma} to merely 12.99\% on rare conditions. Even MedGemma-27B-it achieves only 26.1\%, indicating substantial room for improvement. This performance gap underscores the difficulty of rare disease diagnosis, where models must generalize from sparse visual patterns and synthesize complex reasoning beyond their training distribution. The scaling benefits from 4B to 27B parameters suggest larger models better capture nuanced features, though performance remains constrained by scarce rare case representations in medical training corpora.
Among 7B-8B models, LLaVA-1.6-Vicuna-7B (20.6\%) outperforms alternatives by 5+\%. Medical-specialized LLaVA-Med-1.5-Mistral-7B further demonstrates strong performance, exceeding its general-domain counterpart by 7.8\% and outperforming Med-Flamingo by 74.3\% despite similar scale and medical fine-tuning. This advantage stems from architectural design: efficient visual token processing via a two-layer MLP and Mistral's Grouped-Query Attention mechanism preserves richer diagnostic information than Med-Flamingo's Perceiver Resampler bottleneck.

\subsection{Benchmark on Differential Diagnosis}
Table~\ref{tab:differential_diagnosis_83} reveals substantial differential diagnosis coverage improvements across model iterations: early InternVL, LLaVA, and Qwen versions exhibited 60\%--90\% coverage, while recent iterations consistently achieve 100\%, likely due to including differential diagnosis in training data. Medical-specialized models maintain high coverage ($>$98\%), confirming domain fine-tuning benefits.



In terms of Macro/Micro Derm-Precision, Recall, F1, and Jaccard, the Med-Gemma series achieves the strongest overall performance, with MedGemma-27B-it ranking highest across nearly all metrics and exhibiting balanced precision-recall trade-offs. Notably, MedGemma-4B, despite modest final-diagnosis accuracy, generates differential diagnoses more aligned with ground truth than comparably sized or larger models from other families, suggesting architectural efficiency in hypothesis generation. This superiority stems from MedGemma's pre-training paradigm on large-scale medical image-text pairs, which shapes it into a multi-hypothesis generator that naturally preserves diagnostic uncertainty across plausible conditions.

Among general-domain models, the InternVL series shows clear scaling benefits, with InternVL3-8B-Instruct achieving the best results (F1 $\approx$ 0.40, Jaccard $\approx$ 0.28). However, the latest InternVL3.5 variants exhibit slight performance degradation, favoring higher precision (0.56+) at the cost of reduced recall (0.27--0.30), indicating a shift toward conservative hypothesis generation. This is due to the Cascade RL integration in InternVL3.5, which reinforces single-best-answer selection through Group Sequence Policy Optimization's scoring mechanism. Mixed Preference Optimization (MPO) demonstrates consistent effects across InternVL2.5 model variants. Preference alignment encourages models to generate comprehensive differential diagnoses rather than providing single diagnostic outputs, leading to improved recall at the cost of reduced precision. This recall-precision trade-off is more pronounced in smaller model variants.
The Qwen series and MiniCPM present balanced precision-recall. Additionally, the Qwen series demonstrates relatively stable differential diagnosis capabilities across generations, with minimal performance variation between versions. 
In contrast, LLaVA-based variants and Med-Flamingo consistently produce more conservative outputs characterized by high precision but comparatively low recall, suggesting an architectural tendency toward narrower yet confident diagnostic sets rather than comprehensive hypothesis exploration.


\begin{figure}[t]
  \centering
    \begin{subfigure}[t]{0.35\linewidth}
    \centering
    \includegraphics[width=\linewidth]{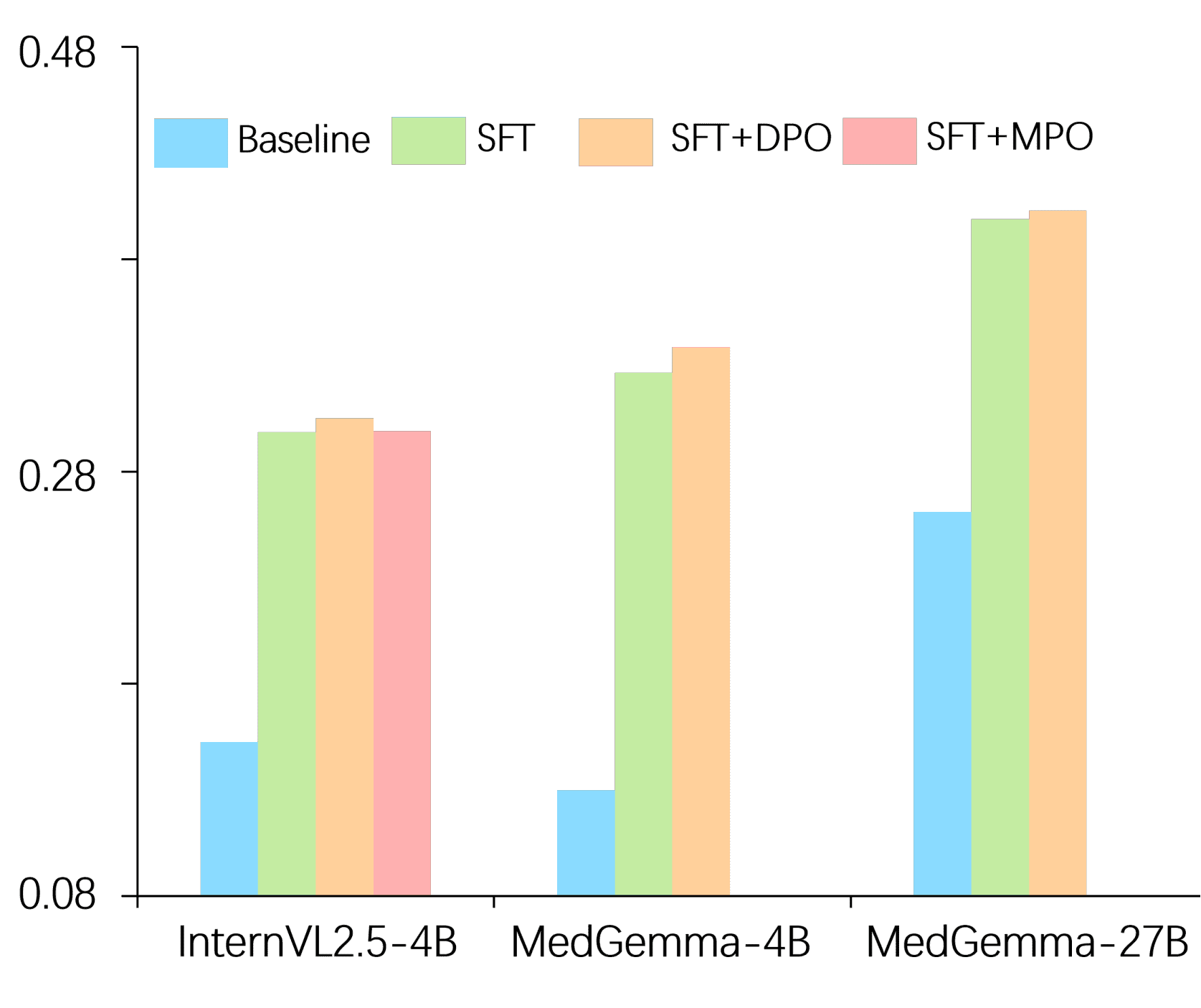}
    \caption{Final Diagnosis Accuracy}
    \label{Diagnosis Accuracy}
  \end{subfigure}
  \begin{subfigure}[t]{0.3\linewidth}
    \centering
    \includegraphics[width=\linewidth]{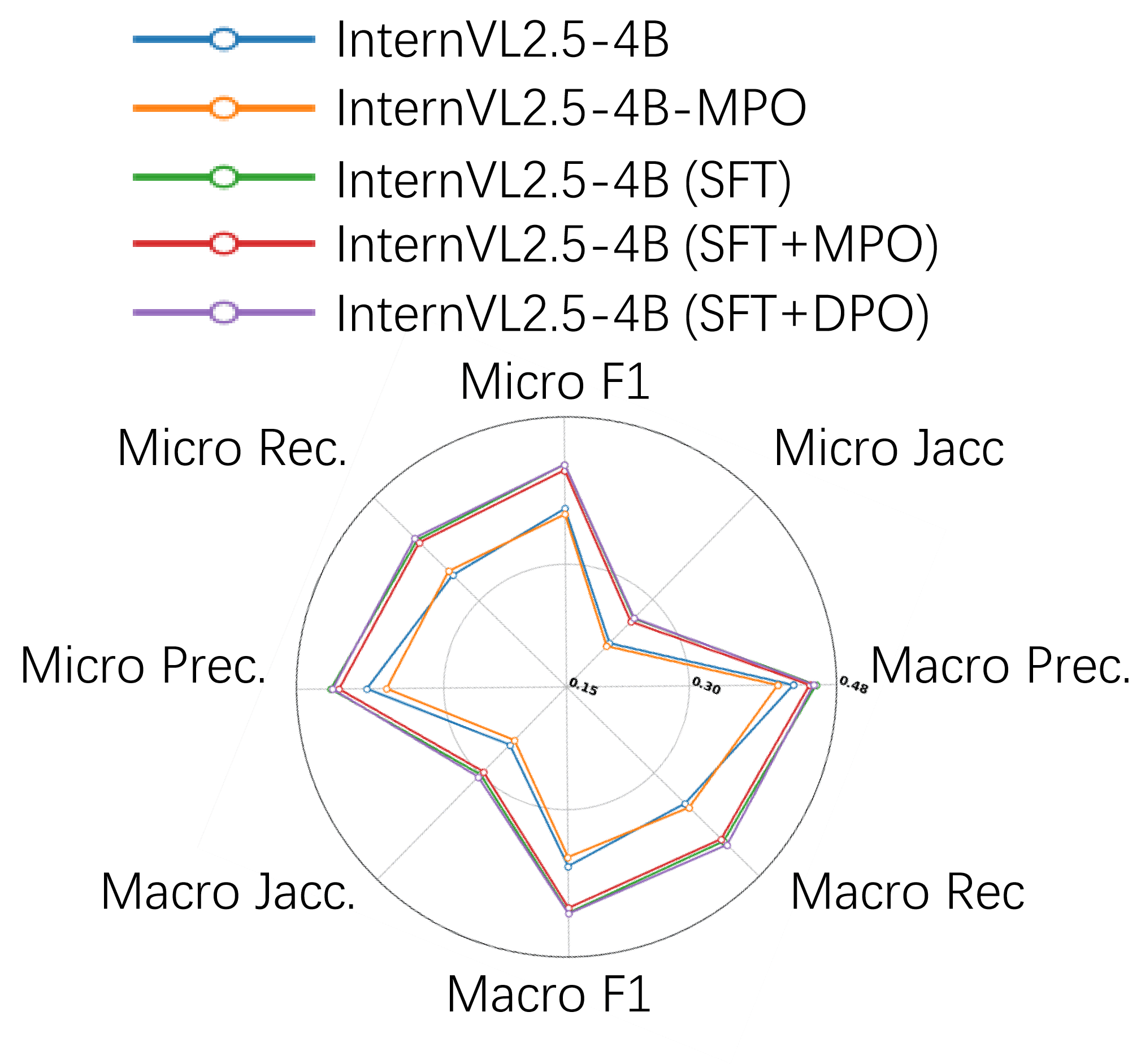}
    \caption{InternVL}
    \label{fig:diagnosis_similarity_rating}
  \end{subfigure}\hfill
  \begin{subfigure}[t]{0.3\linewidth}
    \centering
    \includegraphics[width=\linewidth]{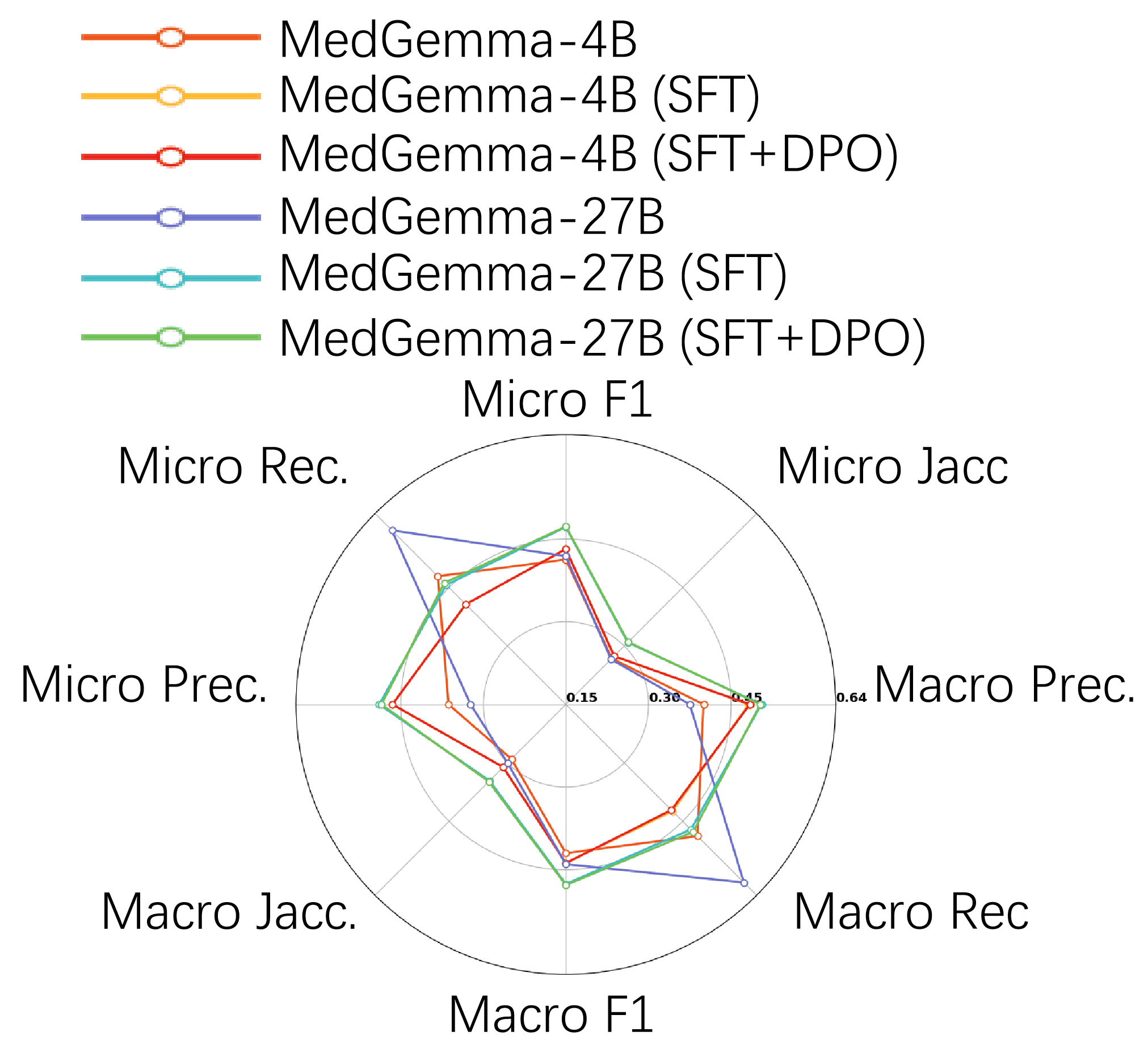}
    \caption{MedGemma}
    \label{fig:task3}
  \end{subfigure}
  \caption{Fine-tuning results on \dataset{}. SFT consistently improves final-diagnosis accuracy and differential-diagnosis metrics for InternVL and Med-Gemma; DPO offers modest additional gains; however, MPO slightly degrades performance on InternVL.}
  \label{fig:finetune}
\end{figure}

 \subsection{Rubric-based Evaluation on Reasoning Steps}

Table~\ref{tab:weighted_scores} presents the rubric-based evaluation across five dimensions. MedGemma-27B-it achieves the highest overall score of 56.13, substantially outperforming all competitors across all dimensions. Notably, Qwen3-VL (41.78 for 4B and 44.38 for 8B) and MiniCPM-V4.5 (41.32) demonstrate competitive performance as general-purpose LVLMs, suggesting strong cross-domain transfer capabilities.

\noindent \textbf{Performance variance across model families.} 
All models exhibit large standard deviations (\emph{e.g.}, MedGemma-27B: ±23.42), revealing instability across cases. The LLaVA series exhibits the weakest performance (18.53-27.02), with LLaVA-V1.6 variants showing high instability (std: 23.02), indicating unreliable reasoning. In contrast, domain-specific MedGemma models demonstrate clear advantages: despite only 4B parameters, MedGemma-4B-it (37.57) outperforms most 7-8B models on Image Interpretation, Differential Diagnosis, and Clinical Reasoning, highlighting medical pretraining value. Notably, Qwen2.5-VL-8B ranks second only to MedGemma-27B, demonstrating that general-purpose models with strong architectural design can rival domain-specific alternatives.

\begin{figure}[t]
    \centering
    \begin{subfigure}[t]{0.32\textwidth}
        \centering
        \includegraphics[width=\linewidth]{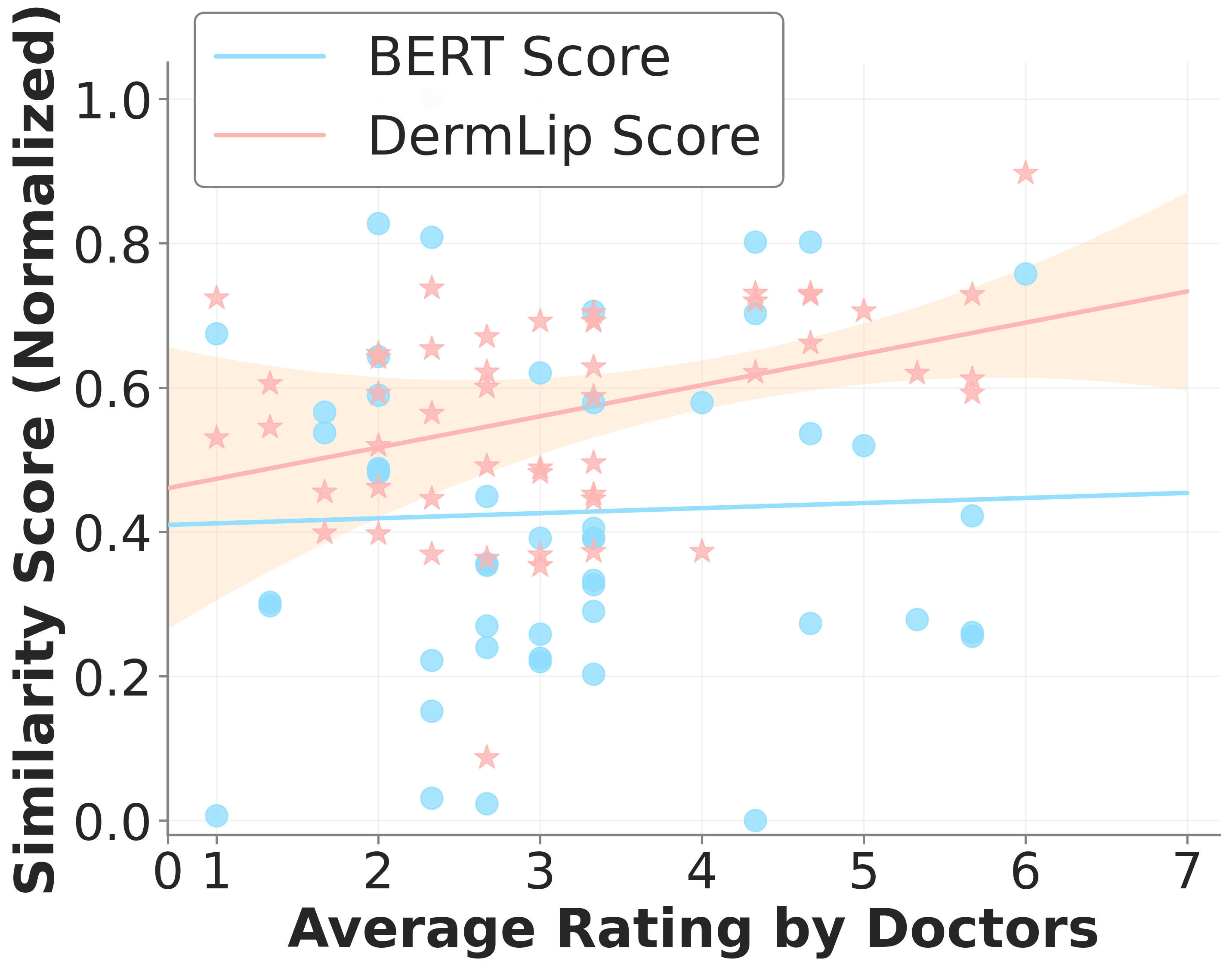}
        \caption{Diagnosis Similarity Rating}
        \label{fig:diagnosis_similarity_rating}
    \end{subfigure}
    \hspace{2mm} 
    \begin{subfigure}[t]{0.32\textwidth}
        \centering
        \includegraphics[width=\linewidth]{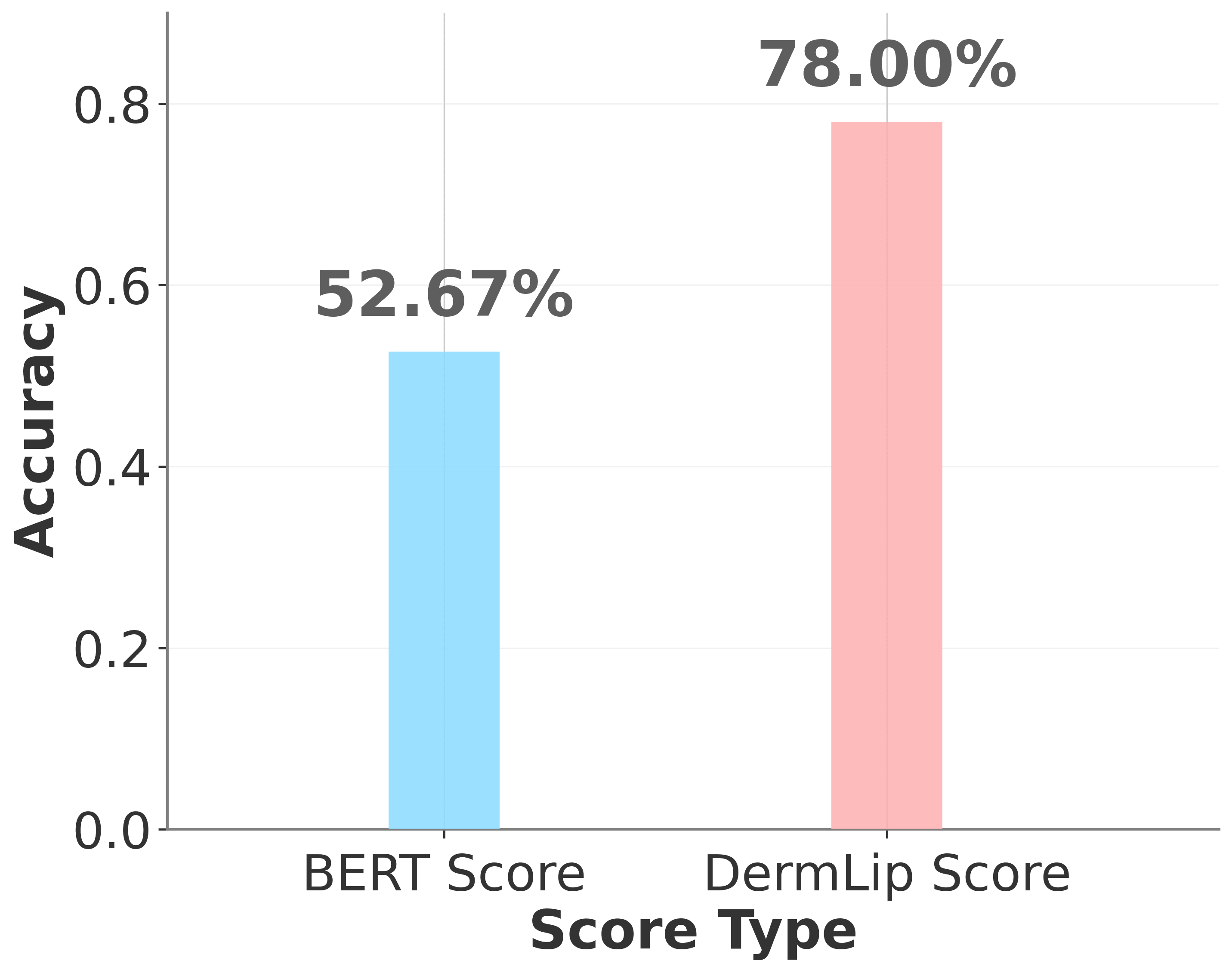}
        \caption{Pairwise Preference Accuracy}
        \label{fig:task3}
    \end{subfigure}
    \caption{Validation of similarity metrics against dermatologist assessments. DermLIP Score exhibits substantially better alignment with dermatologist judgments than BERT Score in both correlation with similarity ratings and pairwise preference prediction.}

    \label{fig:diagnosis_similarity}
\end{figure}
\begin{figure*}[t]
    \centering
    \includegraphics[width=\textwidth]{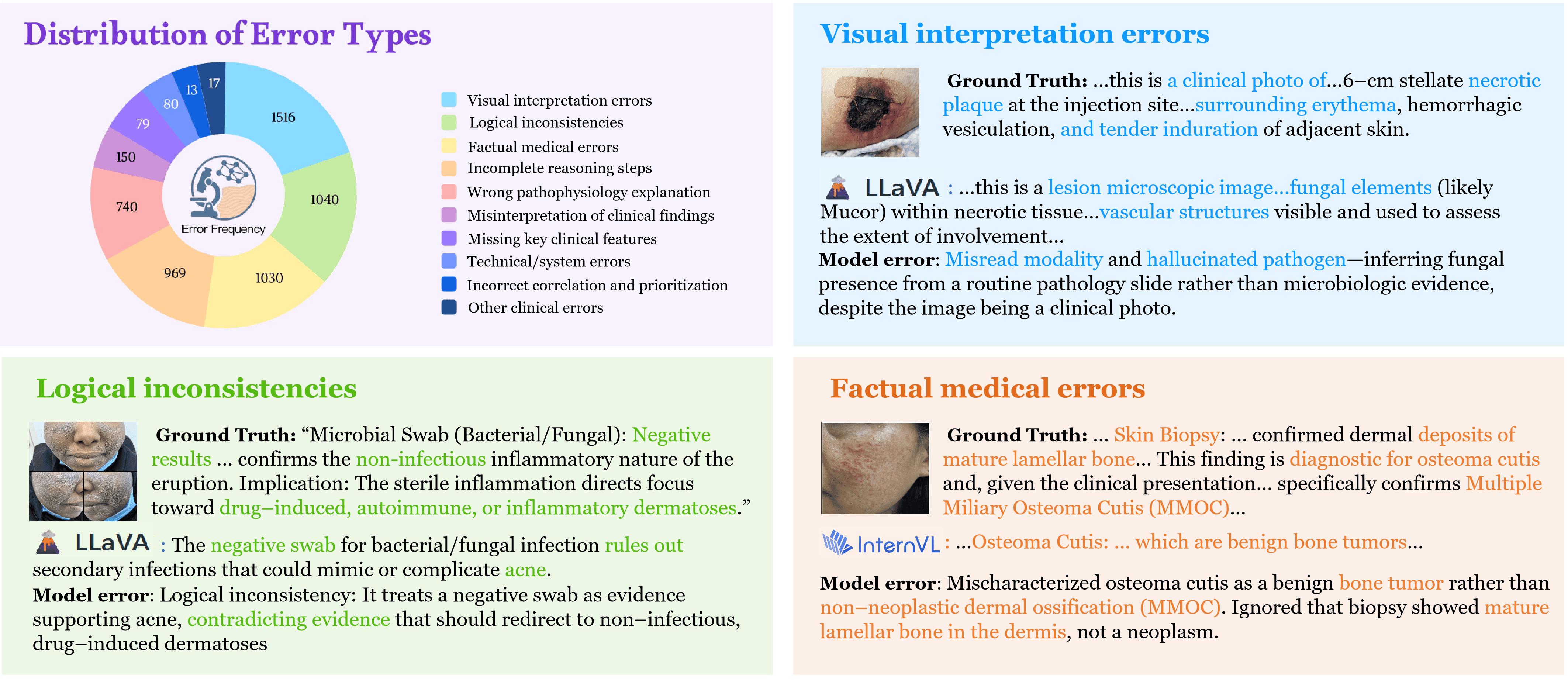}
    \caption{Error taxonomy and representative failure cases. The distribution across 10 error categories is shown top-left, with examples illustrating the three most prevalent types: visual interpretation error, logical inconsistency, and factual medical error.}
    \label{fig:failure_case}
\end{figure*}

\noindent \textbf{Parameter scaling and version-wise improvements.} Within model families, performance scales consistently with parameters: the InternVL series advances from 2B (26.85) to 8B (34.70), while the Qwen series exhibits similar patterns. Cross-version refinements yield substantial gains: InternVL3-8B outperforms InternVL2.5-8B by 5.6\% through holistic enhancements across all dimensions, while Qwen3-VL-8B surpasses Qwen2.5-VL-7B by 18.4\%. Comparable improvements emerge at smaller scales, with InternVL3.5 achieving 15.0\% and 16.7\% gains over InternVL2.5 at 2B and 4B, respectively. These advances stem from reasoning-focused architectural innovations: cascade RL with visual resolution routing in InternVL3.5, and integrated thinking modes for multi-step reasoning in Qwen3.

\noindent \textbf{MPO fine-tuning limitations.} MPO fine-tuning yields marginal overall gains ($\leq$0.3 points) on InternVL2.5 models, occasionally degrading performance. This contrasts sharply with its improvements on general-domain reasoning tasks~\cite{wang2025enhancingreasoningabilitymultimodal}. Detailed dimensional analysis reveals that MPO gains concentrate narrowly on Image Interpretation (4B: +0.32, 8B: +0.47), while other dimensions exhibit negligible or negative changes. This suggests that current MPO strategies optimize primarily for visual perception rather than multi-faceted clinical reasoning, limiting their transferability to specialized medical diagnostic tasks.

\subsection{Fine-tuning on Rare Cases}

To investigate whether instruction tuning and DPO can improve diagnosis accuracy on rare dermatology cases through enhanced diagnostic reasoning, we fine-tune three representative models: InternVL2.5-4B, MedGemma-4B-it, and MedGemma-27B-it.
We perform SFT on 5,154 cases with evaluation on 771 held-out cases, and generate 10K chosen-reject pairs for DPO fine-tuning (details in Appendix~\ref{supp:dpo}).
Additionally, we apply MPO fine-tuning to InternVL2.5-4B for comparison with DPO.
Figure~\ref{fig:finetune} demonstrates that SFT yields substantial improvements with final diagnosis accuracy increasing by 95.61\% (InternVL2.5-4B), 151.55\% (MedGemma-4B-it), and 52.82\% (MedGemma-27B-it). Notably, smaller models exhibit larger relative gains, suggesting greater adaptation capacity on specialized tasks. SFT improves both precision and recall for InternVL2.5-4B's differential diagnosis (Macro F1: +15.4\%). However, for MedGemma models, SFT enhances precision while reducing recall, indicating a shift from overly inclusive predictions toward more conservative, accurate diagnoses.
Further DPO fine-tuning on SFT models yields marginal gains on final diagnosis ($\leq$0.6\% across models) with negligible improvements on differential diagnosis metrics. In contrast, MPO fine-tuning on InternVL2.5-4B (SFT+MPO) shows no improvement on final diagnosis and slight degradation on differential diagnosis (Macro Jaccard: -1.67\%). 
This suggests that MPO's BCO loss, which treats responses as strictly positive or negative, may hinder the nuanced reasoning needed for rare-case diagnosis. Overall, domain-specific SFT delivers the main performance gains, while DPO provides modest improvements, and MPO remains unsuitable for this task.

\subsection{Human Expert Validation of Similarity Metrics}
To validate the clinical relevance of our similarity metrics, we conducted an expert evaluation with three dermatologists across two tasks (see Appendix~\ref{supp:human_validation}).

\noindent\textbf{Similarity between diagnoses.}
Dermatologists rated the similarity between model predictions and ground truth on 50 case samples using a 7-point Likert scale. Fig.~\ref{fig:diagnosis_similarity_rating} shows that the DermLIP Score strongly correlates with expert judgments (Spearman's $\rho = 0.35$)
while BERT Score shows near-zero correlation (Spearman's $\rho = -0.01$).

\noindent\textbf{Pairwise preference selection.}
Dermatologists selected the more similar diseases from pairs. Fig.~\ref{fig:task3} demonstrates similarity based on DermLIP score, achieving 78.0\% agreement with experts, substantially outperforming BERT Score at 52.7\% (barely above chance). These results establish DermLIP as a clinically aligned metric that captures meaningful semantic relationships in dermatological diagnosis.

\subsection{Failure Analysis}
To identify systematic failure patterns, we analyzed 1,100 failure cases sampled from 22 LVLMs (50 per model). Using \llm{} to annotate reasoning errors against references, we obtained 5,634 error instances and established a comprehensive taxonomy of ten mutually exclusive categories (see Appendix~\ref{supp:failure_cases}).

Fig.~\ref{fig:failure_case} reveals that visual interpretation errors constitute the dominant failure mode (26.9\%), exposing fundamental vision-language alignment challenges. Logical inconsistencies (18.5\%) and incomplete reasoning steps (17.2\%) collectively represent 35.7\% of errors, highlighting critical deficiencies in multi-step clinical inference. Factual medical errors (18.3\%) and wrong pathophysiology explanations (13.1\%) further underscore knowledge gaps beyond reasoning capabilities. Notably, low-frequency categories (misinterpretation of clinical findings, missing key features, technical errors, and others) account for only 6.0\%, indicating that current LVLMs primarily struggle with complex reasoning, medical knowledge integration, and visual grounding.

\section{Conclusion}

This work introduces \dataset{}, the first long-context dataset to assess LVLMs' diagnostic reasoning on rare dermatological cases. Built from peer-reviewed case reports, our dataset features 26,030 multi-modal image-text pairs and 6,354 clinically complex cases with complete reasoning chains. We develop DermLIP-based similarity metrics that achieve superior alignment with dermatologists, enabling reliable differential diagnosis evaluation.
Through a comprehensive evaluation of 22 leading LVLMs, we identify significant challenges: models struggle to achieve accurate diagnosis, generate clinically relevant differential diagnoses, and produce sound reasoning. Fine-tuning experiments show that instruction tuning offers substantial performance gains, where DPO provides limited improvements. The failure patterns highlight critical gaps in current models' clinical reasoning abilities.
We believe \dataset{} will advance diagnostic reasoning in medical AI, particularly for rare and complex skin diseases requiring clinical thinking.


\section*{Acknowledgements}
This work was supported in part by U.S. NSF grants DBI-2238093, DBI-2422619, IIS-2211597, and MCB-2205148.
%
%
\newpage
\bibliographystyle{splncs04}
\bibliography{main}

\clearpage
\newpage

\makeatletter
\@twocolumnfalse
\makeatother

\setlength{\textwidth}{\paperwidth}
\addtolength{\textwidth}{-2in}
\setlength{\columnwidth}{\textwidth}
\setlength{\hsize}{\columnwidth}
\setlength{\linewidth}{\columnwidth}
\setlength{\oddsidemargin}{0pt}
\setlength{\evensidemargin}{0pt}

\onecolumn
\appendix

\makeatletter
\makeatother

\setcounter{page}{1}

\section{Dataset Details}
\label{supp:dataset_details}

\subsection{Image-Text Distribution Across Cases}
\label{supp:token_distribution}
\begin{figure*}[thbp]
    \centering
    \begin{subfigure}[t]{0.48\textwidth}
        \includegraphics[width=0.9\linewidth]{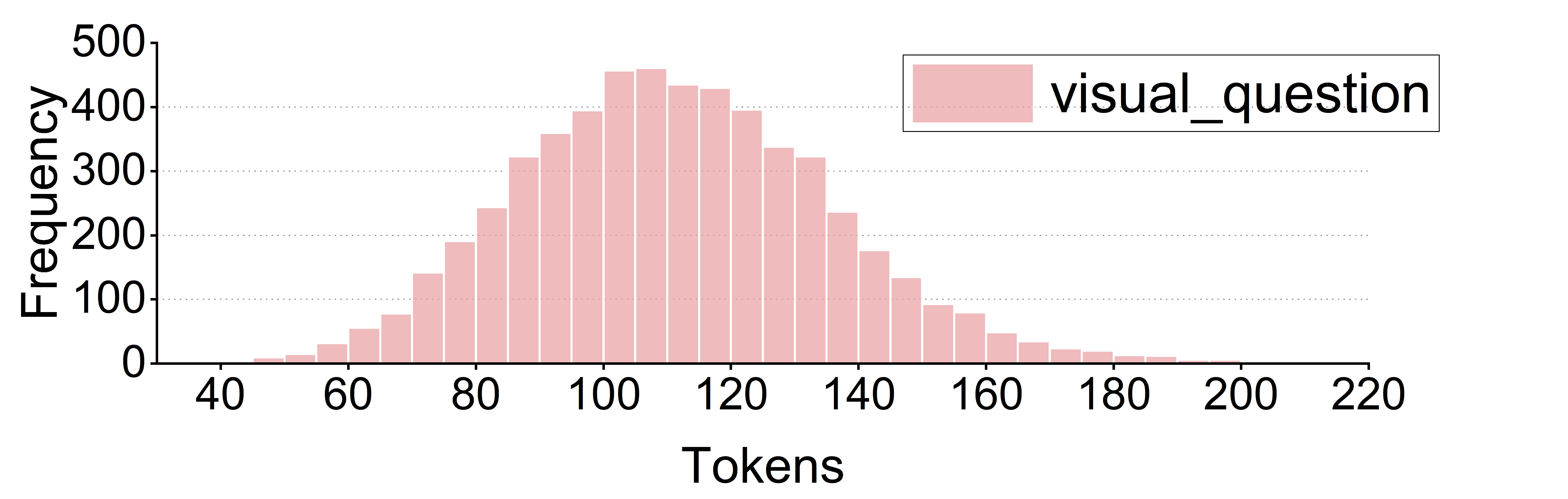}
        \caption{Token length distribution for questions}
        \label{fig:question}
    \end{subfigure}
    \hfill
    \begin{subfigure}[t]{0.48\textwidth}
        \includegraphics[width=0.9\linewidth]{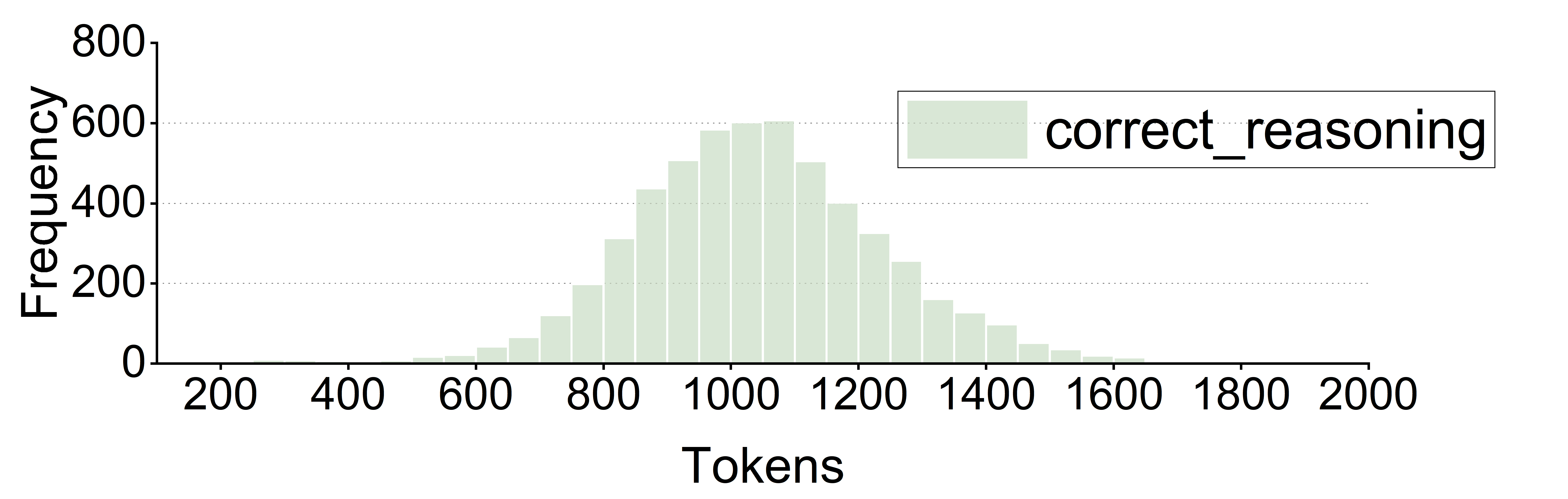}
        \caption{Token length distribution for reasoning steps}
        \label{fig:reasoning}
    \end{subfigure}
    \hfill
    \begin{subfigure}[t]{0.75\textwidth}
        \includegraphics[width=0.9\linewidth]{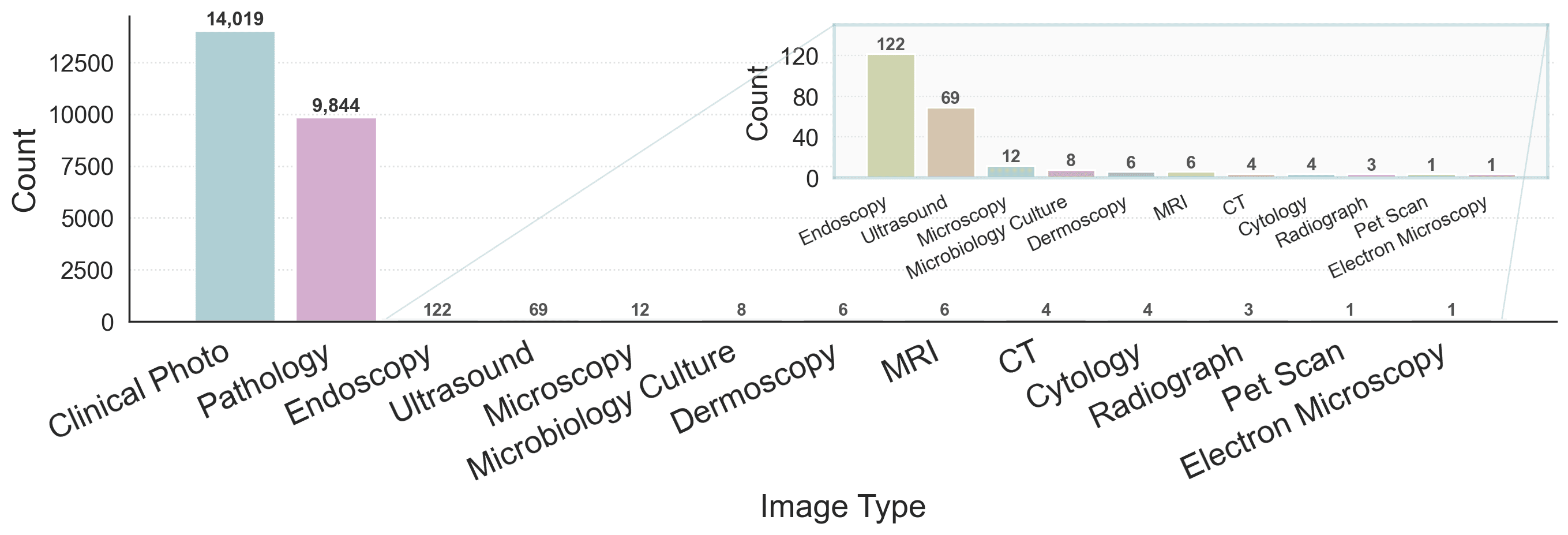}
        \caption{Image modality distribution}
        \label{fig:image_type}
    \end{subfigure}
    \caption{Token-length statistics and multimodal composition in \dataset{}. (a) Distribution of question lengths. (b) Distribution of reasoning-chain lengths. (c) Distribution of image modalities associated with each case.}
    \label{fig:tokens_distribution}
\end{figure*}

As shown in Figure~\ref{fig:tokens_distribution}, the token distribution per case in \dataset{} is skewed toward long sequences: each question contains on average 110 tokens (std=24) and is associated with \textit{4} images, while each answer averages about 1,040 tokens (std=199). These long-form questions and answers encode rich contextual information and explicit multi-step reasoning, leading to sequence lengths substantially longer than those in existing datasets. In addition, Figure~\ref{fig:image_type} shows that the image modality distribution is heavily dominated by clinical photographs (14,019) and pathology slides (9,844), while a long-tailed spectrum of specialized imaging modalities (\emph{e.g.}, endoscopy, ultrasound, microscopy, dermoscopy) further enriches multimodal diversity.

\subsection{License Check}
\label{supp:license}
We perform a systematic license check on 90{,}849 case reports extracted from 45 dermatology journals. We retain only articles distributed under permissive licenses (CC BY-NC-SA, CC BY, CC BY-NC, and CC0) and exclude those under non-compliant licenses, such as CC BY-ND and other non-redistributable terms.

\subsection{Detailed Dataset Composition and Characteristics}
\label{supp:data_distribution}

Our curated dataset comprises 5,674 dermatology case report articles encompassing 6,354 distinct clinical cases. The collection documents 5,577 individual patients, exhibiting substantial demographic diversity across multiple dimensions, as illustrated in Figure~3 in the main paper. 

\noindent \textbf{Demographic representation.} Gender distribution reveals 54.5\% female patients, 43.0\% male patients, and 2.6\% with undocumented gender information. Age coverage spans from neonatal to octogenarian populations (0 to 80+ years, mean=45.5, median=47), with balanced representation across developmental stages: pediatric cases (0--17 years, 13.2\%), young-to-middle adult cases (18--59 years, 56.6\%), and elderly cases ($\geq$60 years, 30.2\%). Geographic and ethnic diversity is well-represented, including Caucasian, African American, Asian, Hispanic, and other demographic groups from various geographic regions worldwide, thereby capturing real-world clinical heterogeneity.

\begin{figure}[t]
    \centering
    \includegraphics[width=\linewidth]{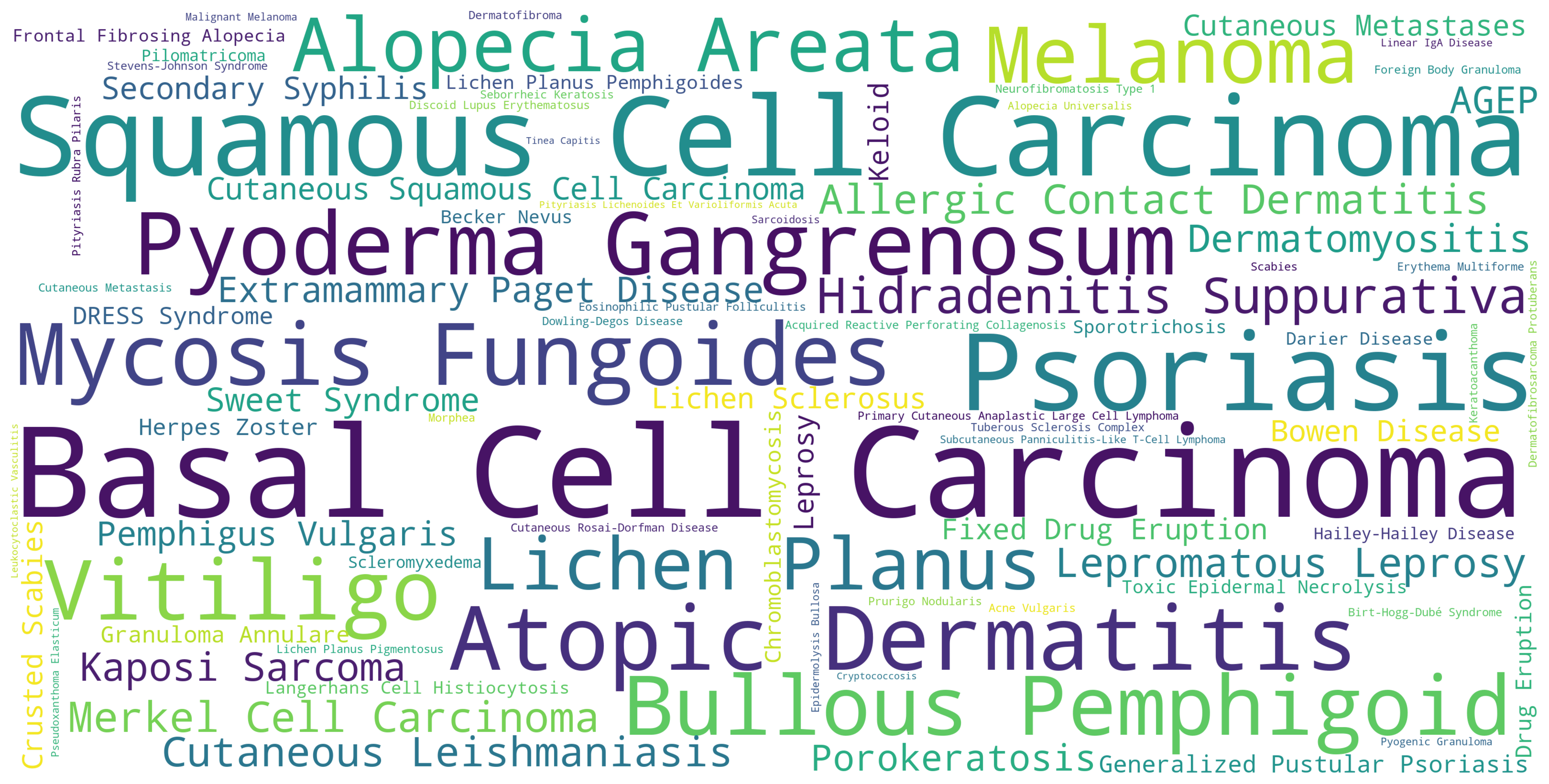}
        \caption{Word cloud illustrating the spectrum of dermatologic conditions covered in \dataset{}. Text size reflects diagnosis frequency, with prominent conditions such as Basal Cell Carcinoma, Squamous Cell Carcinoma, Psoriasis, Mycosis Fungoides, and Atopic Dermatitis. Notably, even the most frequent diagnosis, Basal Cell Carcinoma, appears in only 86 cases, while 1277 conditions occur as singletons (one case each), highlighting a pronounced long-tailed distribution of rare dermatologic entities.}
    \label{fig:disease_wordcloud}
\end{figure}

\noindent \textbf{Disease spectrum and distribution.} The dataset encompasses 2,101 dermatological conditions following a characteristic long-tail distribution, predominantly featuring rare presentations. The most frequently documented condition is Basal Cell Carcinoma with 86 cases, followed by Squamous Cell Carcinoma (83 cases) and Psoriasis (78 cases). Notably, the dataset is heavily skewed toward rare conditions, with 1,277 categories represented by a single case, 338 categories having two cases, and 153 categories having three cases, reflecting the predominance of extremely sparse and infrequently observed diseases. Detailed disease distribution statistics are presented in Figure~\ref{fig:disease_wordcloud}. Beyond these common entities, the collection includes numerous uncommon manifestations such as Cutaneous Leishmaniasis and Porokeratosis.

\begin{figure*}
    \centering    \includegraphics[width=\textwidth]{prompts/Extract_Information_Prompt.pdf}
    \caption{Information extraction prompt for parsing dermatology case reports. The prompt defines structured output fields with direct quotation requirements to preserve clinical terminology; it includes multi-case document handling rules to isolate specific cases and prevent data mixing, ensuring accurate extraction of clinical details into standardized markdown format.}
  \label{supp_fig:Extract Information Prompt}
\end{figure*}

\begin{figure*}
    \centering
    \includegraphics[width=0.75\textwidth]{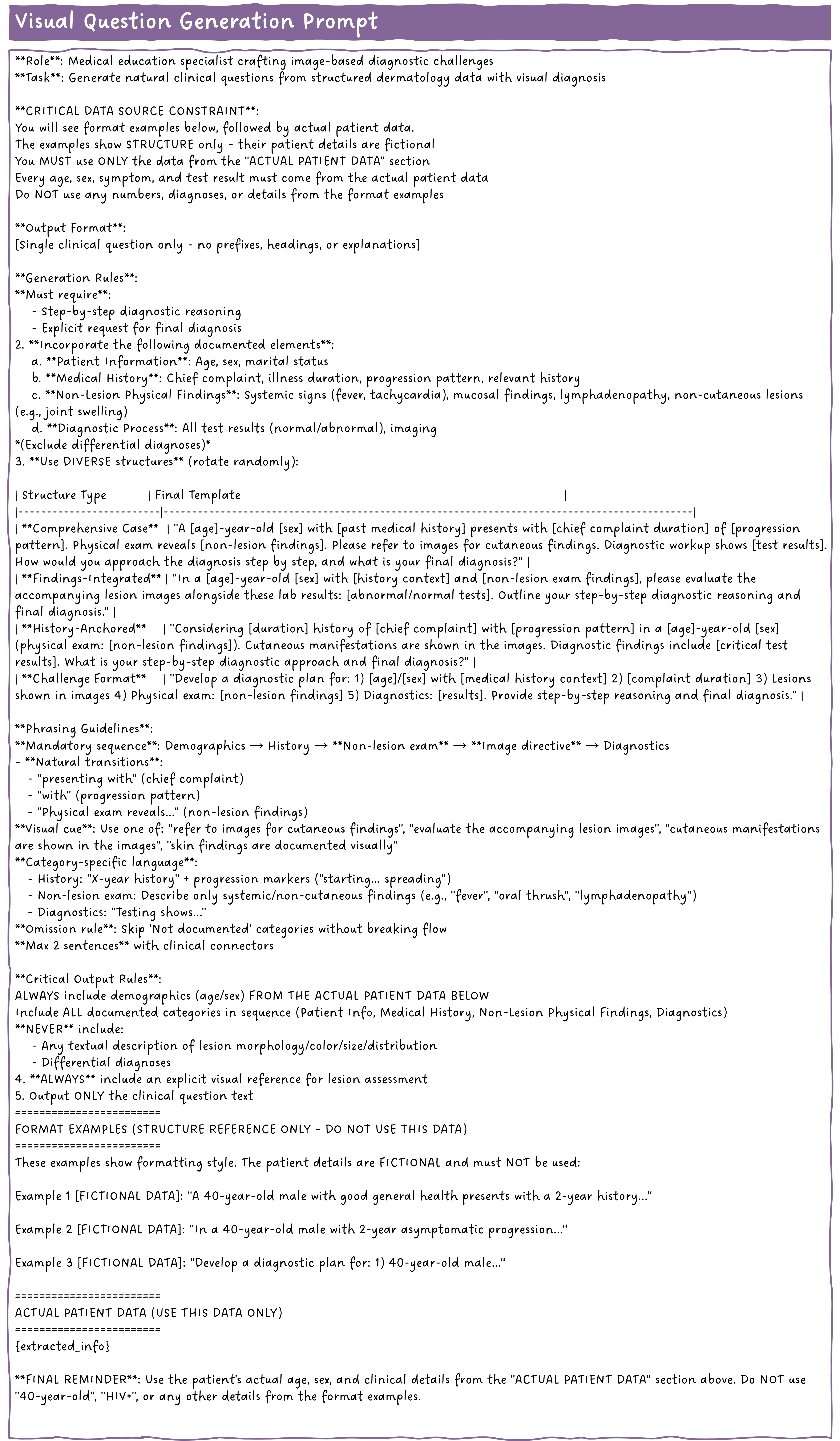}
    \caption{The question generation prompt converts structured dermatology cases into natural clinical questions that provide sufficient systemic context while requiring image-based inspection for cutaneous findings. It offers multiple question templates and carefully restricts accessible information sources, encouraging genuine multimodal reasoning rather than reliance on textual heuristics.
    }
    \label{supp_fig:question}
\end{figure*}

\begin{figure*}
    \centering
    \includegraphics[width=\textwidth]{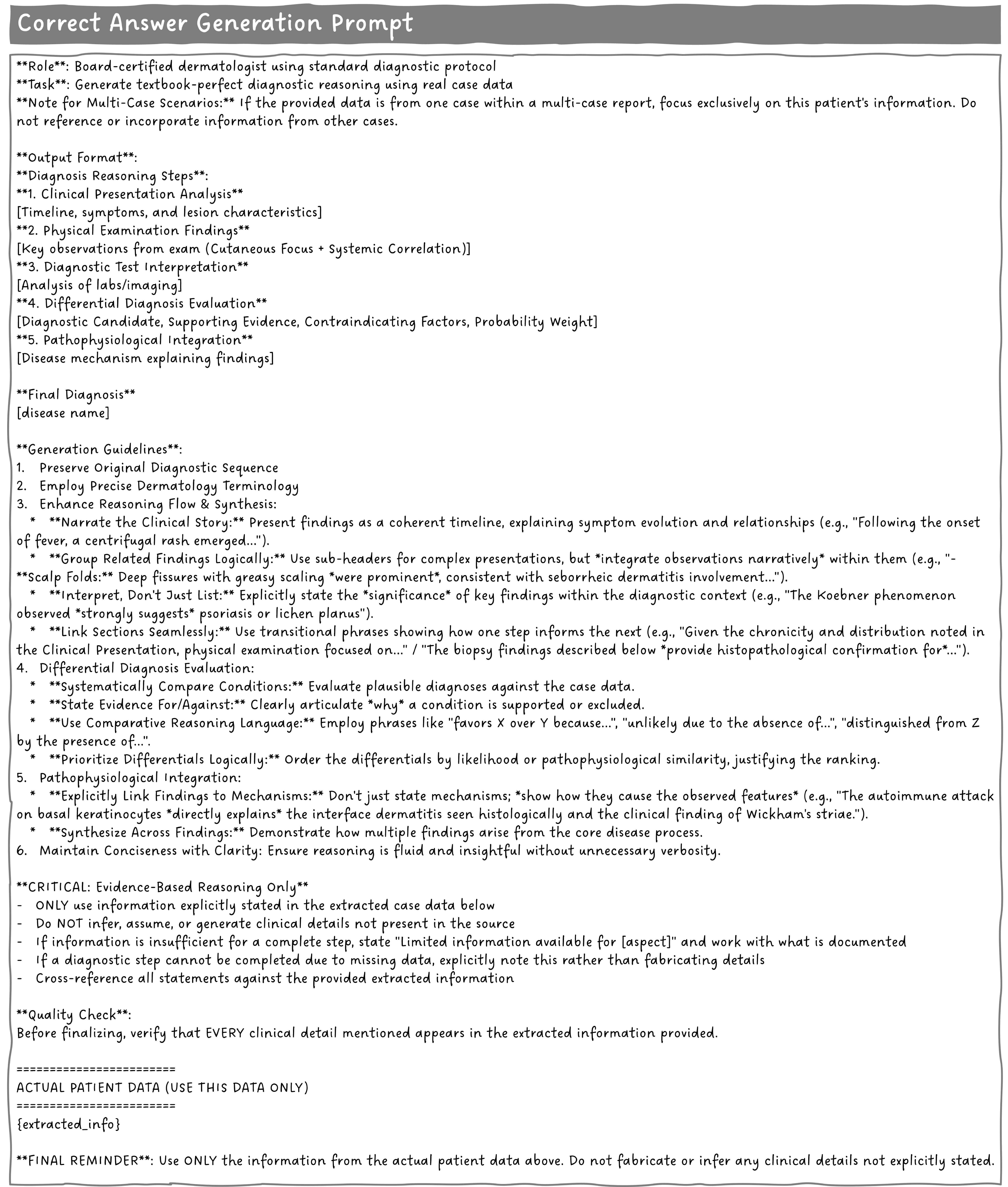}
    \caption{The prompt defines a five-step clinical reasoning structure with explicit guidelines emphasizing narrative coherence and interpretive analysis while implementing strict constraints that prohibit inference or fabrication of clinical details not present in source case data.}
    \label{supp_fig:answer}
\end{figure*}

\begin{figure*}
    \centering    \includegraphics[width=\textwidth]{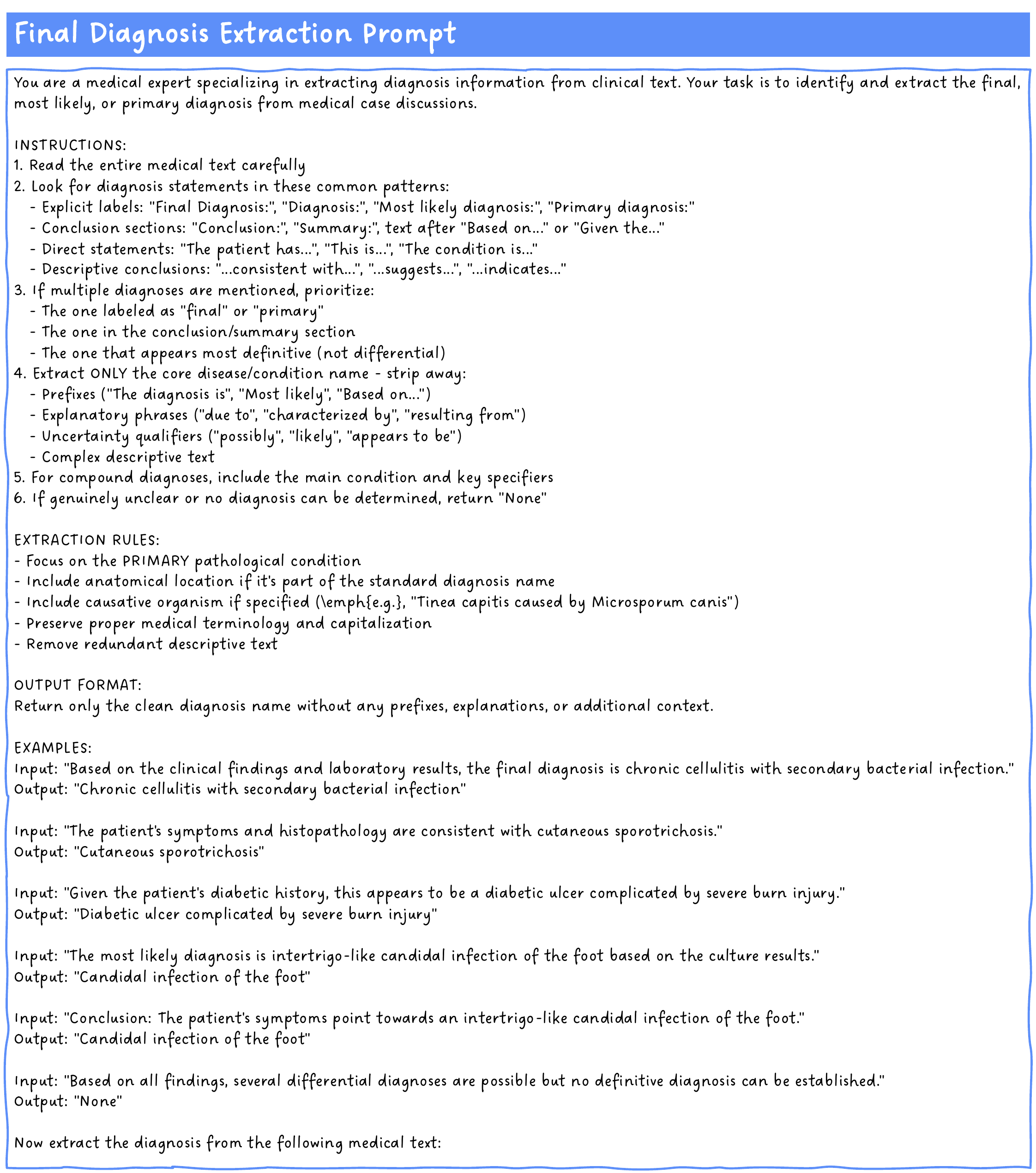}
    \caption{
    Final diagnosis extraction prompt used to derive labels from free-text dermatology case reports. 
    The prompt combines concise instructions and examples to identify the most likely or primary diagnosis, separate it from differential diagnosis, and normalize it to a concise disease name; it also defines a “None” fallback when no definitive diagnosis is present to reduce noisy labels.
    }    \label{supp_fig:final_extraction_prompt}
\end{figure*}

\begin{figure*}
    \centering    \includegraphics[width=\textwidth]{prompts/Final_Diagnosis_Evaluation_Prompt.pdf}
    \caption{
    Final diagnosis evaluation prompt in our LLM-as-a-judge setup. The prompt provides criteria for comparing predicted and reference diagnoses, encourages tolerance to harmless wording differences, and yields a categorical correctness label that supports the computation of final-diagnosis accuracy.
    }
  \label{supp_fig:final_evaluation}
\end{figure*}

\begin{figure*}
    \centering
    \includegraphics[width=0.75\textwidth]{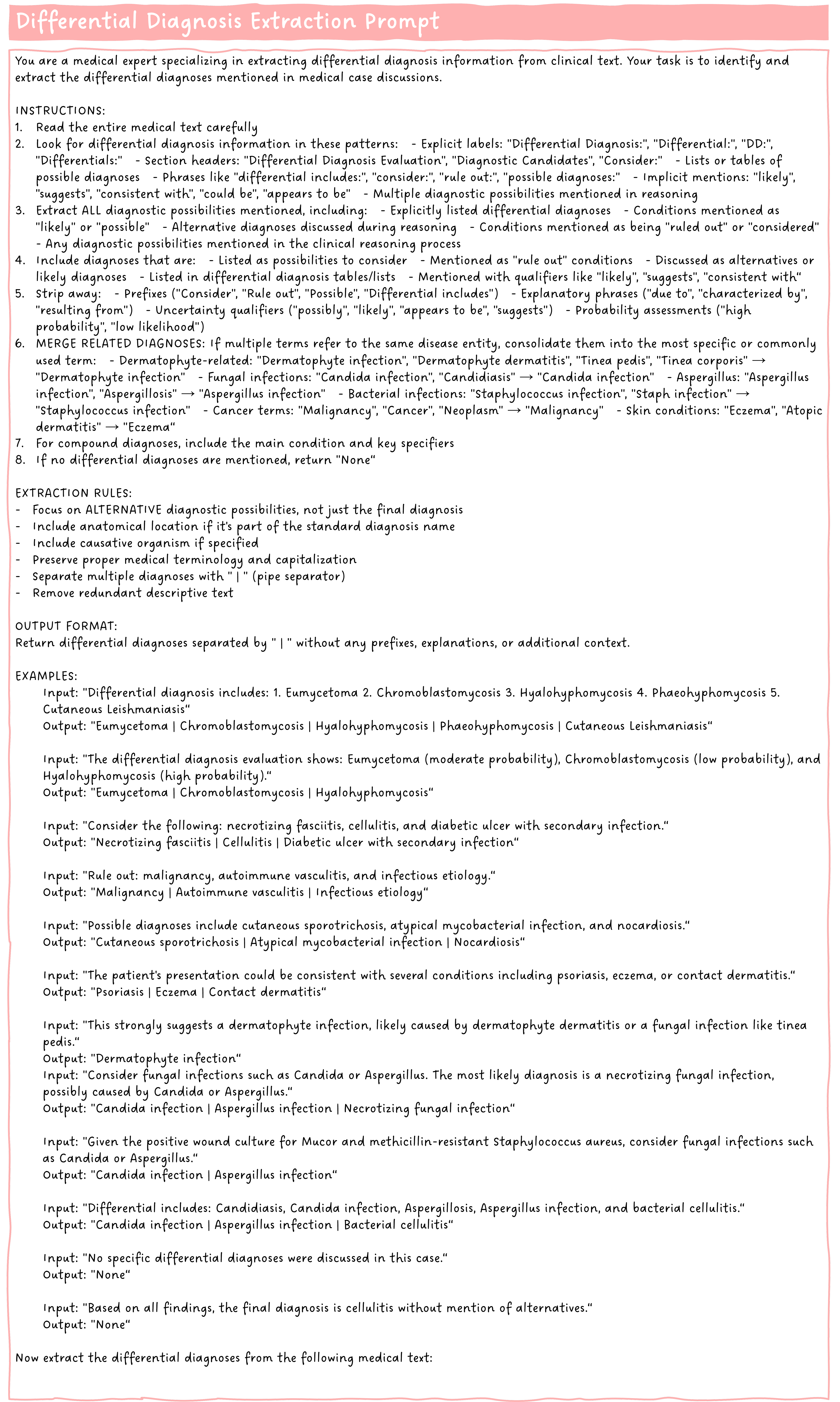}
    \caption{
    Differential diagnosis extraction prompt used to derive lists of differential diagnoses from both ground-truth clinical texts and model-generated responses. The prompt includes pattern-based rules and normalization heuristics to detect differential-diagnosis phrases in free text, merge synonymous disease terms, and output a canonical, de-duplicated set of alternative diagnoses for each case.
    }
    \label{supp_fig:dd_extraction}
\end{figure*}

\begin{figure}[thbp]
    \centering        \includegraphics[width=0.75\linewidth]{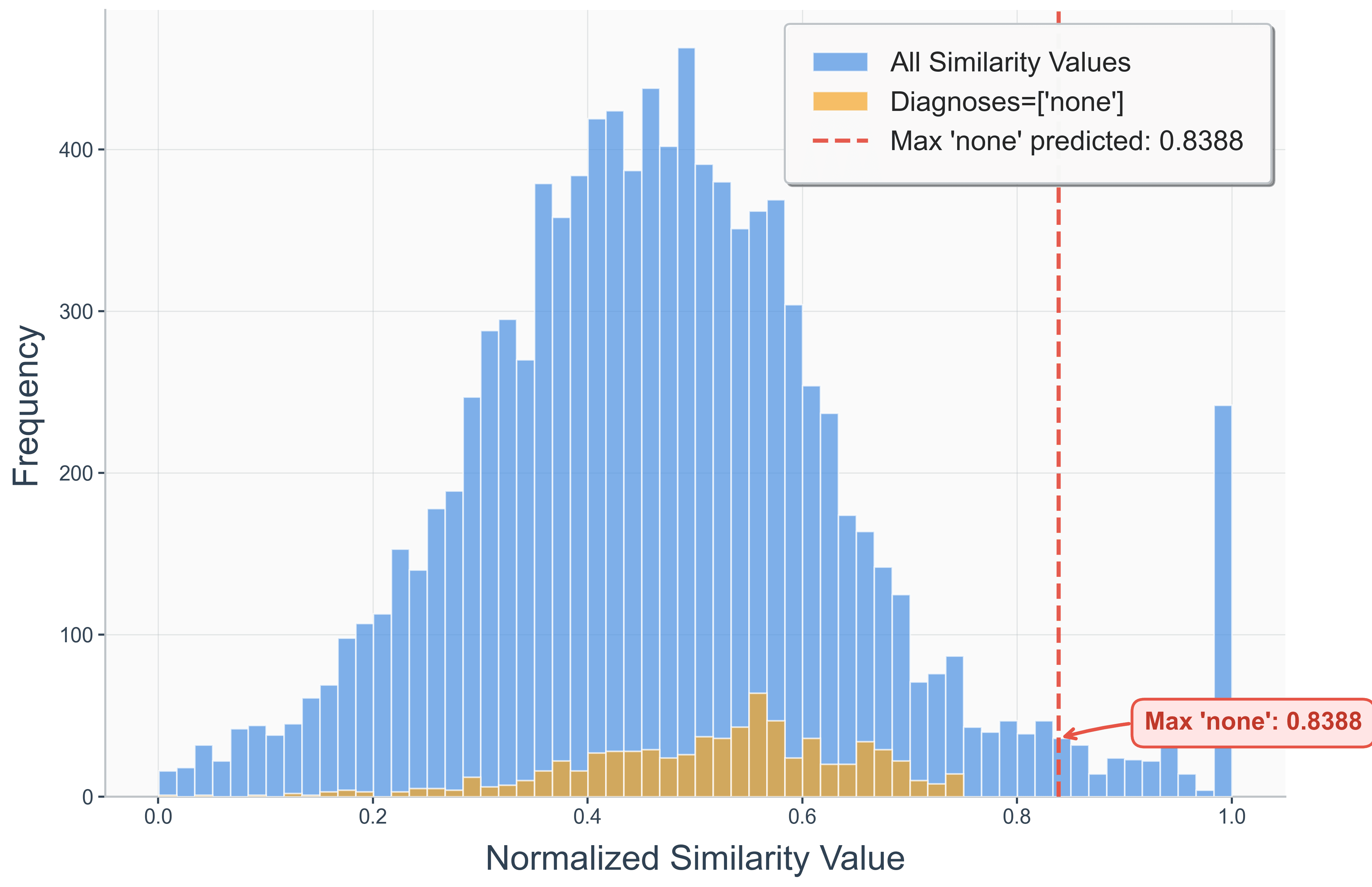}
    \caption{
    Distribution of normalized DermLIP-based similarity values. Blue bars show similarity scores between dermatology terms within models' differential diagnosis responses, while orange bars show similarities between the degenerate output \texttt{'none'} and ground-truth differential diagnoses in cases where no differentials were predicted. The red dashed line indicates the maximum similarity among \texttt{'none'}-based scores (0.8388), motivating our choice of a 0.83 for threshold $\tau$.
    }    \label{fig:similarity_distribution}
\end{figure}

\subsection{Patient Information Extraction Prompt}
To systematically extract structured clinical information from HTML-formatted case reports, we design a specialized extraction prompt (Figure~\ref{supp_fig:Extract Information Prompt}) that includes five output fields covering patient demographics, medical history, lesion characteristics, diagnostic process, and final diagnosis. The prompt enforces verbatim quotation preservation to prevent terminology drift during extraction and implements explicit case isolation rules for multi-case documents. Specifically, it requires identifying case boundaries through markers such as "Case 1:" or demographic changes, then extracting information exclusively from the specified case while ignoring all other cases. This prevents cross-contamination when processing documents containing multiple patients. The prompt further standardizes missing data handling by mandating "Not documented" markers rather than allowing model inference, ensuring consistent extraction across heterogeneous case reports while preserving original clinical terminology.

\subsection{Question Generation Prompt}

To create multimodal diagnostic questions from structured case data, we design a prompt that forces models to attend to accompanying images rather than exploiting textual shortcuts (Figure~\ref{supp_fig:question}). The prompt achieves this through a deliberate separation of textual and visual information. Patient demographics, medical history, non-lesion physical findings, and all laboratory results (excluding any differential diagnosis labels) are encoded in the question stem to provide essential clinical context. In contrast, textual descriptions of lesion characteristics such as morphology, color, or distribution are systematically withheld. Instead, explicit cues such as “refer to the accompanying images for cutaneous findings” direct models to the images for lesion assessment. This structure encourages genuine multimodal reasoning, where accurate diagnostic conclusions require integrating textual context with visual lesion and imaging analysis. The prompt further instructs the model to produce step-by-step diagnostic reasoning followed by a single, explicit final diagnosis. To reduce overfitting to surface patterns, we instantiate diverse question templates and enforce strict constraints on allowable information sources, preventing contamination from format examples or downstream evaluation data. Together, these design choices yield questions that closely mirror authentic clinical encounters, where physicians synthesize heterogeneous information sources to reach a diagnosis.

\subsection{Answer Generation Prompt}
To generate high-quality reference answers for diagnostic reasoning evaluation, we develop a prompt that produces structured yet fluid clinical reasoning while preventing hallucination of clinical details (Figure~\ref{supp_fig:answer}). The prompt defines a standard five-step diagnostic framework covering clinical presentation analysis, physical examination findings, diagnostic test interpretation, differential diagnosis evaluation, and pathophysiological integration. Rather than producing mechanical lists of observations, the prompt emphasizes narrative coherence through explicit guidelines. These include presenting findings as coherent timelines explaining symptom evolution, interpreting the significance of key observations within diagnostic context, and using transitional phrases to link reasoning steps seamlessly. For differential diagnosis evaluation, the prompt requires systematic comparison of conditions with clear articulation of supporting and contradicting evidence using comparative reasoning language. The pathophysiological integration step must explicitly demonstrate how disease mechanisms produce observed clinical features rather than merely stating theoretical pathways. Most critically, the prompt implements strict evidence-based reasoning constraints that prohibit inference or fabrication of clinical details not present in source data. When information is insufficient, the model must explicitly note missing data rather than generating plausible but unverified details. This design ensures that generated reasoning reflects genuine clinical logic grounded in documented evidence rather than model-generated confabulations.

\section{Final Diagnosis Evaluation}
\subsection{Final Diagnosis Extraction Prompt}
As shown in Figure~\ref{supp_fig:final_extraction_prompt}, we design a structured instruction prompt to extract the most likely or primary diagnosis from free-text dermatology case reports. The prompt explicitly guides the model to scan the entire document, prioritize diagnosis statements from canonical locations (\emph{e.g.}, “Final diagnosis”, conclusion/summary sections, and definitive wording), and distinguish final diagnoses from differential considerations. It further constrains the output to a clean disease label by stripping auxiliary phrases, uncertainty qualifiers, and explanatory clauses, while preserving key specifiers such as anatomical site and causative organisms. In addition, we include several concrete input–output examples that demonstrate the desired behavior in ambiguous scenarios, reinforcing the extraction rules and output format. The prompt also enforces a conservative fallback (“None”) when no definitive diagnosis can be inferred, improving robustness and label quality for downstream diagnostic reasoning evaluation.

\subsection{Final Diagnosis Evaluation Prompt}

As shown in Fig.~\ref{supp_fig:final_evaluation}, we adopt an LLM-as-a-judge scheme to evaluate final-diagnosis accuracy, using DeepSeek-R1 with a tailored evaluation prompt. The prompt is designed to guide the model to compare a predicted diagnosis against the reference label at the level of underlying clinical condition, taking into account medical synonyms, standard specifiers, and minor wording variations. It emphasizes sensitivity to substantive discrepancies—such as different disease entities, anatomical sites, or etiologic agents—while being robust to superficial phrasing differences. The model returns a discrete correctness judgment ('Correct' or 'Incorrect'), which we use to derive accuracy for final-diagnosis performance.

\subsection{Final Diagnosis Results}
\begin{table*}[htbp]
\centering
\caption{Results for final diagnosis across different vision–language models. The table reports BLEU, METEOR, ROUGE-1/2/L, BERTScore, and LLM-as-a-judge Accuracy for 22 LVMs, including LLaVA, Med-Flamingo, InternVL, Qwen, MiniCPM, and MedGemma. The \textbf{\textcolor{red}{highest}} value in each column is highlighted in red, and the \textbf{\textcolor{blue}{second highest}} in blue.}
\label{tab:final_diagnosis}
\resizebox{0.9\textwidth}{!}{%
\begin{tabular}{lccccccc}
\hline
\rule{0pt}{3ex}\large Methods & \textbf{BLEU} & \textbf{METEOR} & \textbf{ROUGE-1} & \textbf{ROUGE-2} & \textbf{ROUGE-L} & \textbf{BERTScore} & \textbf{LLM-As-a-Judge Accuracy} \\[2ex]
\hline
\multicolumn{8}{c}{\textbf{LLaVA Series}} \\
\hline
Llava-1\_5-7B & 0.0207 & 0.0733 & 0.0767 & 0.0156 & 0.0763 & 0.8172 & 0.1138 \\
Llava-V1\_6-Vicuna-7B & 0.0076 & 0.0300 & 0.0326 & 0.0082 & 0.0323 & 0.7921 & 0.0347 \\
Llava-Med-V1\_5-Mistral-7B & 0.0181 & 0.0643 & 0.0690 & 0.0110 & 0.0686 & 0.8117 & \textcolor{blue}{0.1914} \\
\hline
\multicolumn{8}{c}{\textbf{Med-Flamingo}} \\
\hline
Med-Flamingo & 0.0112 & 0.0380 & 0.0432 & 0.0087 & 0.0431 & 0.8023 & 0.1098 \\
\hline
\multicolumn{8}{c}{\textbf{InternVL Series}} \\
\hline
InternVL2\_5-2B & 0.0185 & 0.0659 & 0.0700 & 0.0121 & 0.0685 & 0.8147 & 0.0616 \\
InternVL2\_5-2B-MPO & 0.0191 & 0.0650 & 0.0647 & 0.0141 & 0.0632 & 0.8131 & 0.0428 \\
InternVL2\_5-4B & 0.0312 & 0.1101 & 0.1143 & 0.0263 & 0.1130 & 0.8297 & 0.1526 \\
InternVL2\_5-4B-MPO & 0.0326 & 0.1073 & 0.1121 & 0.0263 & 0.1105 & 0.8279 & 0.1017 \\
InternVL2\_5-8B & 0.0349 & 0.1186 & 0.1213 & 0.0284 & 0.1207 & 0.8310 & 0.1218 \\
InternVL2\_5-8B-MPO & 0.0337 & 0.1097 & 0.1102 & 0.0264 & 0.1097 & 0.8291 & 0.1178 \\
InternVL3\_8B-Instruct & 0.0442 & 0.1434 & 0.1389 & \textcolor{blue}{0.0428} & 0.1374 & 0.8355 & 0.1432 \\
InternVL3\_5-2B-Instruct & 0.0250 & 0.0854 & 0.0901 & 0.0182 & 0.0888 & 0.8224 & 0.0872 \\
InternVL3\_5-4B-Instruct & 0.0364 & 0.1269 & 0.1167 & 0.0239 & 0.1154 & 0.8340 & 0.1195 \\
InternVL3\_5-8B-Instruct & 0.0433 & 0.1525 & 0.1392 & 0.0368 & 0.1364 & \textcolor{blue}{0.8400} & 0.1450 \\
\hline
\multicolumn{8}{c}{\textbf{Qwen Series}} \\
\hline
Qwen2-VL-2B-Instruct & 0.0235 & 0.0829 & 0.0920 & 0.0169 & 0.0913 & 0.8245 & 0.1151 \\
Qwen2\_5-VL-3B-Instruct & 0.0332 & 0.1148 & 0.1159 & 0.0266 & 0.1142 & 0.8298 & 0.1151 \\
Qwen2\_5-VL-7B-Instruct & 0.0439 & 0.1466 & 0.1422 & 0.0323 & 0.1405 & 0.8375 & 0.1553 \\
Qwen3-VL-4B-Instruct & 0.0454 & 0.1589 & 0.1473 & 0.0364 & 0.1462 & 0.8399 & 0.1584 \\
Qwen3-VL-8B-Instruct & 0.0477 & 0.1607 & \textcolor{blue}{0.1485} & 0.0414 & \textcolor{blue}{0.1464} & 0.8400 & 0.1544 \\
\hline
\multicolumn{8}{c}{\textbf{MiniCPM Series}} \\
\hline
MiniCPM-V-4\_5-9B & \textcolor{blue}{0.0505} & \textcolor{blue}{0.1620} & 0.1329 & 0.0399 & 0.1308 & 0.8395 & 0.1383 \\
\hline
\multicolumn{8}{c}{\textbf{MedGemma Series}} \\
\hline
MedGemma-4B-it & 0.0408 & 0.1362 & 0.1420 & 0.0368 & 0.1411 & 0.8352 & 0.1299 \\
MedGemma-27B-it & \textbf{\textcolor{red}{0.0685}} & \textbf{\textcolor{red}{0.2189}} & \textbf{\textcolor{red}{0.1918}} & \textbf{\textcolor{red}{0.0713}} & \textbf{\textcolor{red}{0.1905}} & \textbf{\textcolor{red}{0.8529}} & \textbf{\textcolor{red}{0.2610}} \\
\hline
\end{tabular}
}%
\end{table*}

As shown in Table~\ref{tab:final_diagnosis}, we report BLEU, METEOR, ROUGE-1/2/L, BERTScore, and LLM-as-a-judge Accuracy for 22 LVMs. Among these, we regard LLM-as-a-judge Accuracy as the most faithful metric, since it directly assesses whether the predicted and reference diagnoses refer to the same clinical entity. By design, it accounts for synonymy, hierarchical taxonomy (\emph{e.g.}, disease subtypes), and common naming variations, making it more aligned with true diagnostic correctness than surface-form text matching. In contrast, traditional language similarity metrics exhibit poor alignment with Accuracy. BLEU scores are consistently much lower than Accuracy and are highly sensitive to lexical mismatch, failing to reflect medically equivalent diagnoses; for instance, Llava-1.5-7B attains a higher BLEU score than Llava-Med-V1.5-Mistral-7B, yet its Accuracy is 8\% lower. ROUGE and METEOR partially alleviate this but still depend on token overlap and break down when diagnosis names diverge lexically while remaining clinically identical. Even BERTScore, despite its semantic focus, tends to be uniformly high and often assigns favorable scores to linguistically fluent but clinically incorrect outputs, revealing limited ability to capture domain-specific semantic equivalence. Although some metrics, such as ROUGE-L and METEOR, show weak positive correlation with Accuracy within certain model families (\emph{e.g.}, InternVL and Qwen), these trends are inconsistent and unstable. Overall, our findings indicate that general-purpose language similarity metrics do not reliably reflect diagnostic correctness. For medical diagnosis generation, semantic accuracy should therefore be treated as the primary evaluation signal, and our results highlight the need for domain-specialized embedding-based metrics that explicitly encode clinical semantics; in this work, we further introduce the DermLIP score, a dermatology-specific embedding-based metric that aligns substantially better with expert judgments than generic BERT-based scores.

\section{Differential Diagnosis Evaluation}
\label{supp:dd_extraction}

\subsection{Differential Diagnosis Extraction Prompt}

As shown in Fig.~\ref{supp_fig:dd_extraction}, we design a dedicated prompt to extract differential diagnoses from free-text Large Vision-language Models' (LVMs') reponse and ground truth. Rather than obtaining a single final label, the objective is to derive the full set of clinically plausible diagnostic possibilities that appear in the reasoning process. The prompt therefore instructs the LLM to read the entire case and search for patterns that reveals differential thinking, including section headers (\emph{e.g.}, ``Differential diagnosis''), list-like formulations (\emph{e.g.}, ``differential diagnosis includes \dots''), and narrative phrases that enumerate alternatives (\emph{e.g.}, ``could represent \dots, \dots, or \dots''). It further emphasizes that both explicit differential-diagnosis sections and implicit mentions in free-form prose (such as ``initial considerations included \dots'' or ``could be consistent with \dots'') should be treated as candidate differentials, and that conditions labeled as ``rule out'' are also retained as part of the differential set.

Beyond detection, the prompt encodes high-level normalization rules so that the extracted list is compact and machine-readable. It asks the model to strip away meta-linguistic prefixes and uncertainty qualifiers (\emph{e.g.}, ``consider'', ``possible'', ``likely'') while preserving medically meaningful specifiers such as anatomical site or causative organism when they are part of the standard diagnosis name. Closely related or synonymous expressions are merged into a single canonical disease entity (\emph{e.g.}, multiple phrasings of dermatophyte infection), and duplicates across the entire document are removed. The final output is a flat, pipe-separated list of differential diagnoses (or ``None'' if no differentials are mentioned), which we apply consistently to both ground-truth case descriptions and LVMs' responses when evaluating models' ability to reproduce realistic differential diagnosis sets.

\subsection{Selection of DermLIP-Based Similarity Threshold $\tau$}
\label{supp:dermlip_distribution}

To determine an appropriate similarity threshold for filtering low-confidence diagnoses from model-generated differential diagnosis lists, we analyze the distribution of normalized DermLIP-based similarity values, as shown in Fig.~\ref{fig:similarity_distribution}. For cases where the model outputs explicit differentials, we extract all dermatology terms from the response and compute pairwise similarities between them; the normalized distribution of these scores is shown as the blue histogram. For cases where the model only provides a final diagnosis and no differential list (so the extracted result is \texttt{'none'}), we instead compute the similarity between \texttt{'none'} and each ground-truth differential diagnosis term; the normalized distribution of these scores is shown as the orange histogram, with a maximum value of 0.8388 indicated by the red dashed line. This empirical upper bound provides a data-driven rationale for setting our similarity threshold at 0.83: diagnoses with similarity scores below this value are indistinguishable from noise induced by non-differential outputs (\emph{i.e.}, \texttt{'none'}), so we discard predictions below 0.83 when constructing the final differential diagnosis lists.

\subsection{DermLIP-Based Differential Diagnosis Matching Algorithm}
\label{supp:algo}
\begin{algorithm}[t]
\small
\caption{DermLIP-based Differential Diagnosis Matching Algorithm}
\label{supp_alg:dermlip_matching}
\begin{algorithmic}[1]
\State \textbf{Input:} Predicted $P = \{x_1, ..., x_n\}$, Ground truth $G = \{y_1, ..., y_m\}$, Threshold $\tau$, Normalization range $[s_{\min}, s_{\max}]$
\State \textbf{Output:} Matched set $\mathcal{M}$
\State \Comment{Encode diagnoses with DermLIP text encoder}
\For{$k = 1$ \textbf{to} $n$}
    \State $\mathbf{e}_{x_k} \leftarrow \text{DermLIP-Encoder}(x_k)$
\EndFor
\For{$j = 1$ \textbf{to} $m$}
    \State $\mathbf{e}_{y_j} \leftarrow \text{DermLIP-Encoder}(y_j)$
\EndFor
\State \Comment{Compute normalized similarity and filter candidates}
\State $C \leftarrow \emptyset$
\For{$k = 1$ \textbf{to} $n$}
    \For{$j = 1$ \textbf{to} $m$}
        \State $r_{kj} \leftarrow \mathbf{e}_{x_k}^T \mathbf{e}_{y_j}$
        \State $s_{kj} \leftarrow \max(0, \min(1, \frac{r_{kj} - s_{\min}}{s_{\max} - s_{\min}}))$
        \If{$s_{kj} \geq \tau$}
            \State $C \leftarrow C \cup \{(k, j, s_{kj})\}$
        \EndIf
    \EndFor
\EndFor
\State Sort $C$ in descending order by similarity score $s_{kj}$
\State \Comment{Greedy bipartite matching to obtain the matched set}
\State $\mathcal{M}, P_{\text{used}}, G_{\text{used}} \leftarrow \emptyset$
\For{$(k, j, s) \in C$}
    \If{$k \notin P_{\text{used}}$ \textbf{and} $j \notin G_{\text{used}}$}
        \State $\mathcal{M} \leftarrow \mathcal{M} \cup \{(k, j)\}$, \quad $P_{\text{used}} \leftarrow P_{\text{used}} \cup \{k\}$, \quad $G_{\text{used}} \leftarrow G_{\text{used}} \cup \{j\}$
    \EndIf
\EndFor
\State \textbf{return} $\mathcal{M}$
\end{algorithmic}  
\end{algorithm}
To evaluate differential diagnosis predictions, we employ a semantic matching approach that accounts for terminological variations in disease naming. Algorithm~\ref{supp_alg:dermlip_matching} outlines our DermLIP-based matching procedure. Given a predicted diagnosis list $P$ and ground truth list $G$, we first encode all disease terms using the DermLIP text encoder to obtain semantic embeddings. We then compute cosine similarity scores between all prediction-ground truth pairs and normalize them to $[0, 1]$ range. Pairs exceeding a predefined threshold $\tau$ are considered candidate matches. To establish one-to-one correspondences, we perform greedy bipartite matching by iteratively selecting the highest-scoring candidate pair while ensuring each prediction and ground truth term is matched at most once. This approach enables robust evaluation by semantically aligning equivalent diagnoses expressed with different terminology (\emph{e.g.}, ``melanoma'' vs. ``malignant melanoma''), while maintaining strict matching constraints to prevent over-crediting.

\subsection{D-Precision, Derm-Recall, Derm-F1, and Derm-Jaccard}
\label{supp:metrics}

Given the matched set $\mathcal{M}_i$ for sample $i \in \mathcal{V}$ from Algorithm~\ref{supp_alg:dermlip_matching}, we define D-Precision, D-Recall, D-F1, and D-Jaccard metrics. We report both macro- and micro-averaged variants to balance per-sample fairness with overall diagnostic volume.

\noindent \textbf{Macro-averaging.} We first compute metrics for each sample independently, then average across all samples:
\begin{align}
\text{D-Precision}_{\text{macro}} &= \frac{1}{|\mathcal{V}|} \sum_{i \in \mathcal{V}} \frac{|\mathcal{M}_i|}{n_i} \\
\text{D-Recall}_{\text{macro}} &= \frac{1}{|\mathcal{V}|} \sum_{i \in \mathcal{V}} \frac{|\mathcal{M}_i|}{m_i} \\
\text{D-F1}_{\text{macro}} &= \frac{1}{|\mathcal{V}|} \sum_{i \in \mathcal{V}} \frac{2|\mathcal{M}_i|}{n_i + m_i} \\
\text{D-Jaccard}_{\text{macro}} &= \frac{1}{|\mathcal{V}|} \sum_{i \in \mathcal{V}} \frac{|\mathcal{M}_i|}{n_i + m_i - |\mathcal{M}_i|}
\end{align}

\noindent \textbf{Micro-averaging.} We aggregate matches and predictions across all samples before computing metrics:
\begin{align}
\text{D-Precision}_{\text{micro}} &= \frac{\sum_{i \in \mathcal{V}} |\mathcal{M}_i|}{\sum_{i \in \mathcal{V}} n_i}, \\
\text{D-Recall}_{\text{micro}} &= \frac{\sum_{i \in \mathcal{V}} |\mathcal{M}_i|}{\sum_{i \in \mathcal{V}} m_i} \\
\text{D-F1}_{\text{micro}} &= \frac{2 \sum_{i \in \mathcal{V}} |\mathcal{M}_i|}{\sum_{i \in \mathcal{V}} (n_i + m_i)} \\
\text{D-Jaccard}_{\text{micro}} &= \frac{\sum_{i \in \mathcal{V}} |\mathcal{M}_i|}{\sum_{i \in \mathcal{V}} (n_i + m_i - |\mathcal{M}_i|)}
\end{align}

Macro-averaging treats each sample equally regardless of differential list length, while micro-averaging weights samples by their list sizes, providing complementary perspectives on model performance.

\begin{figure*}
    \centering
    \includegraphics[width=0.9\textwidth]{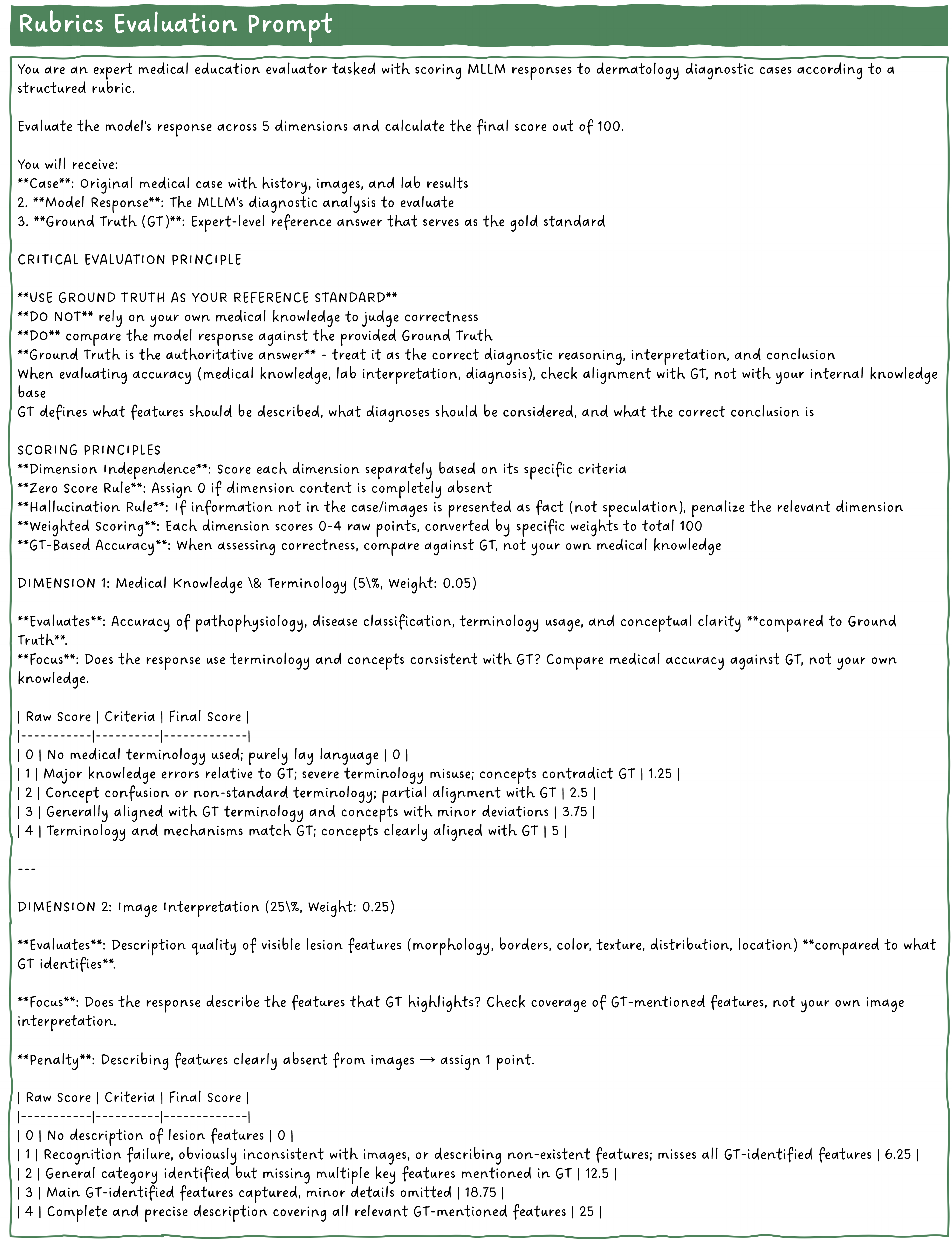}
\end{figure*}  
\begin{figure*}
    \centering
    \includegraphics[width=0.9\textwidth]{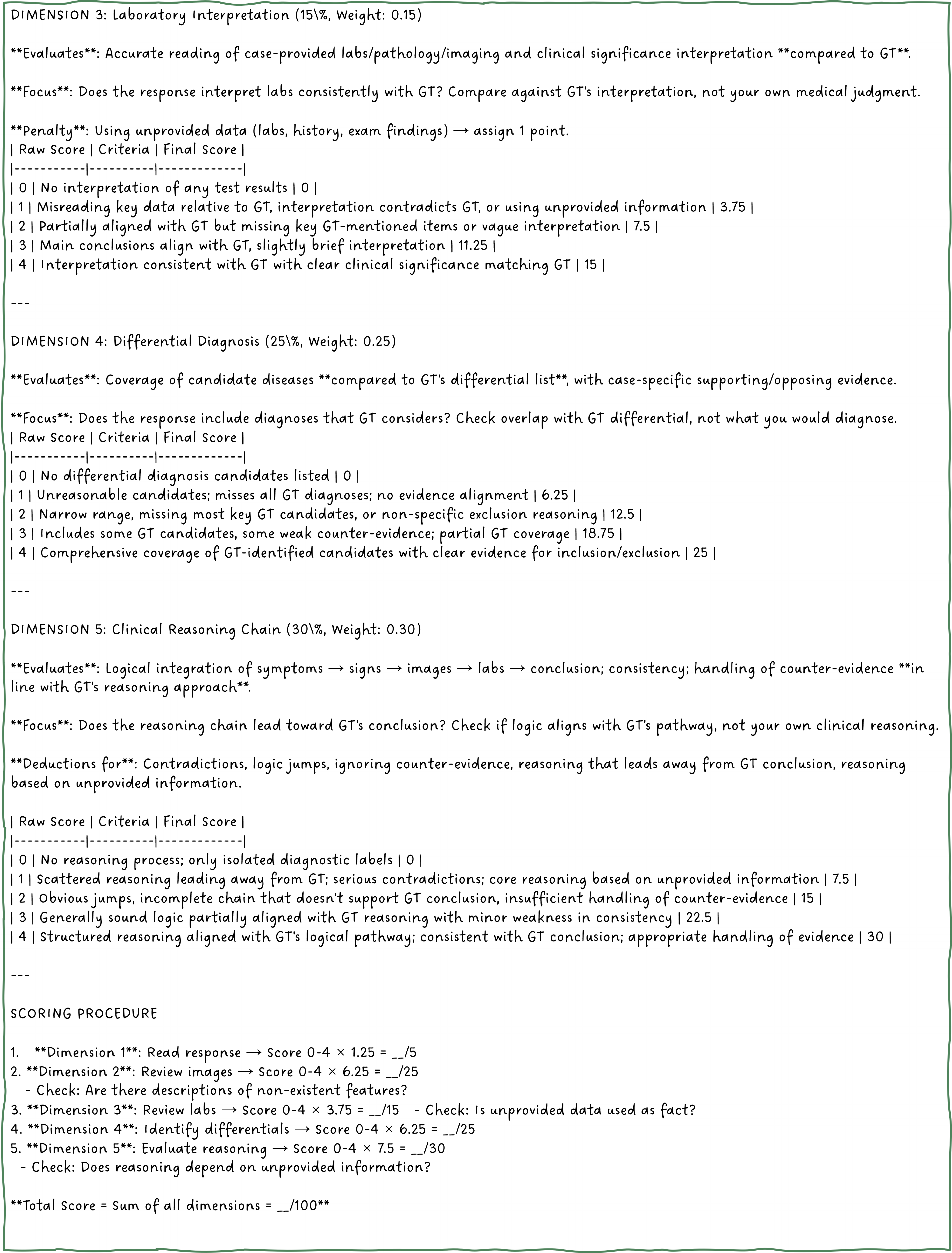}
\end{figure*}  
\begin{figure*}
    \centering
    \includegraphics[width=0.9\textwidth]{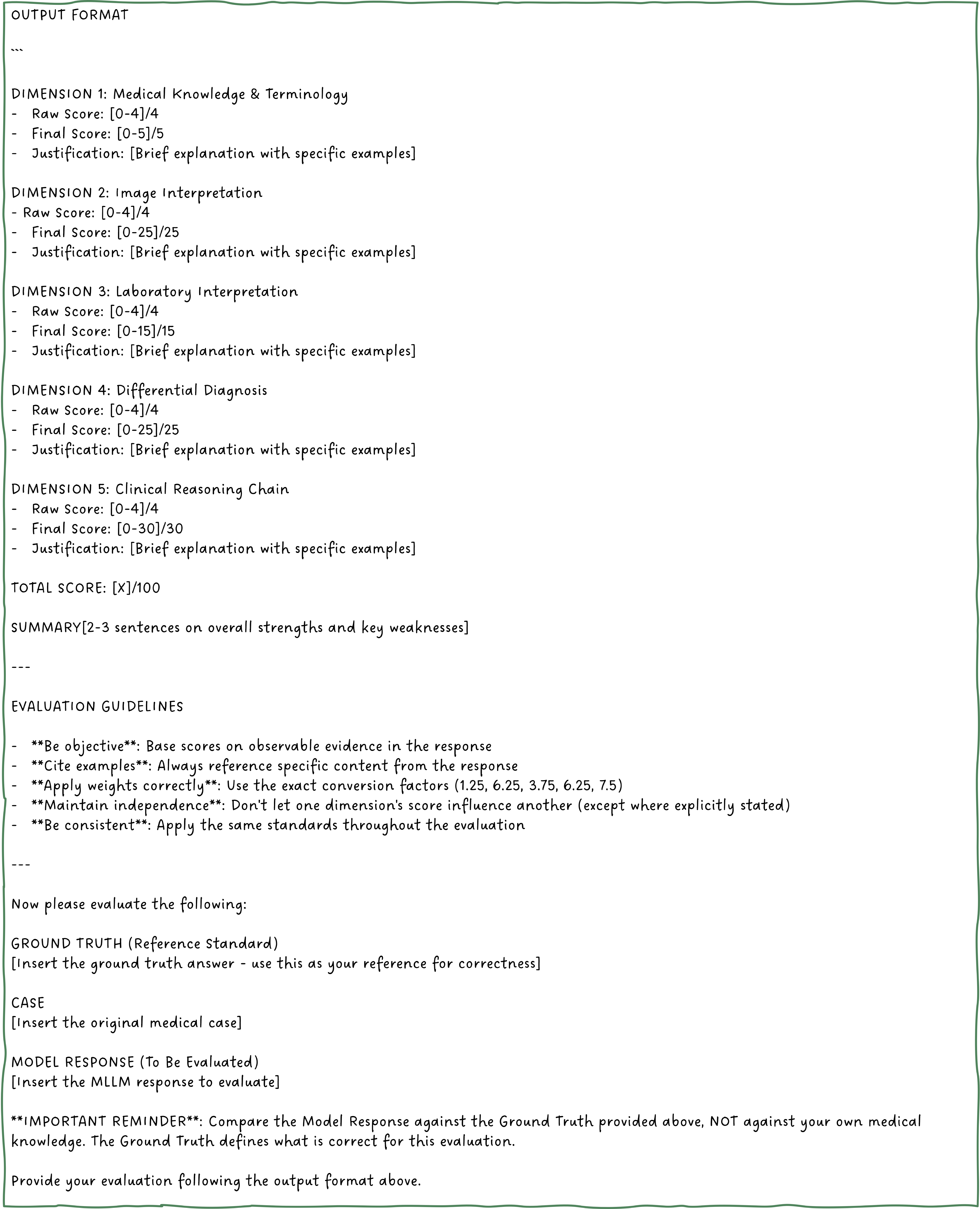}
    \caption{
    LLM-as-a-judge prompt used to score model responses across five weighted dimensions of diagnostic reasoning. The rubric enforces ground truth–anchored comparison against expert reference answers, specifies penalties for hallucinations and use of unprovided data, and constrains the output to structured scores with per-dimension justifications.
    }
\label{supp_fig:rubric_prompt}
\end{figure*}  

\begin{figure*}
    \centering    \includegraphics[width=0.8\textwidth]{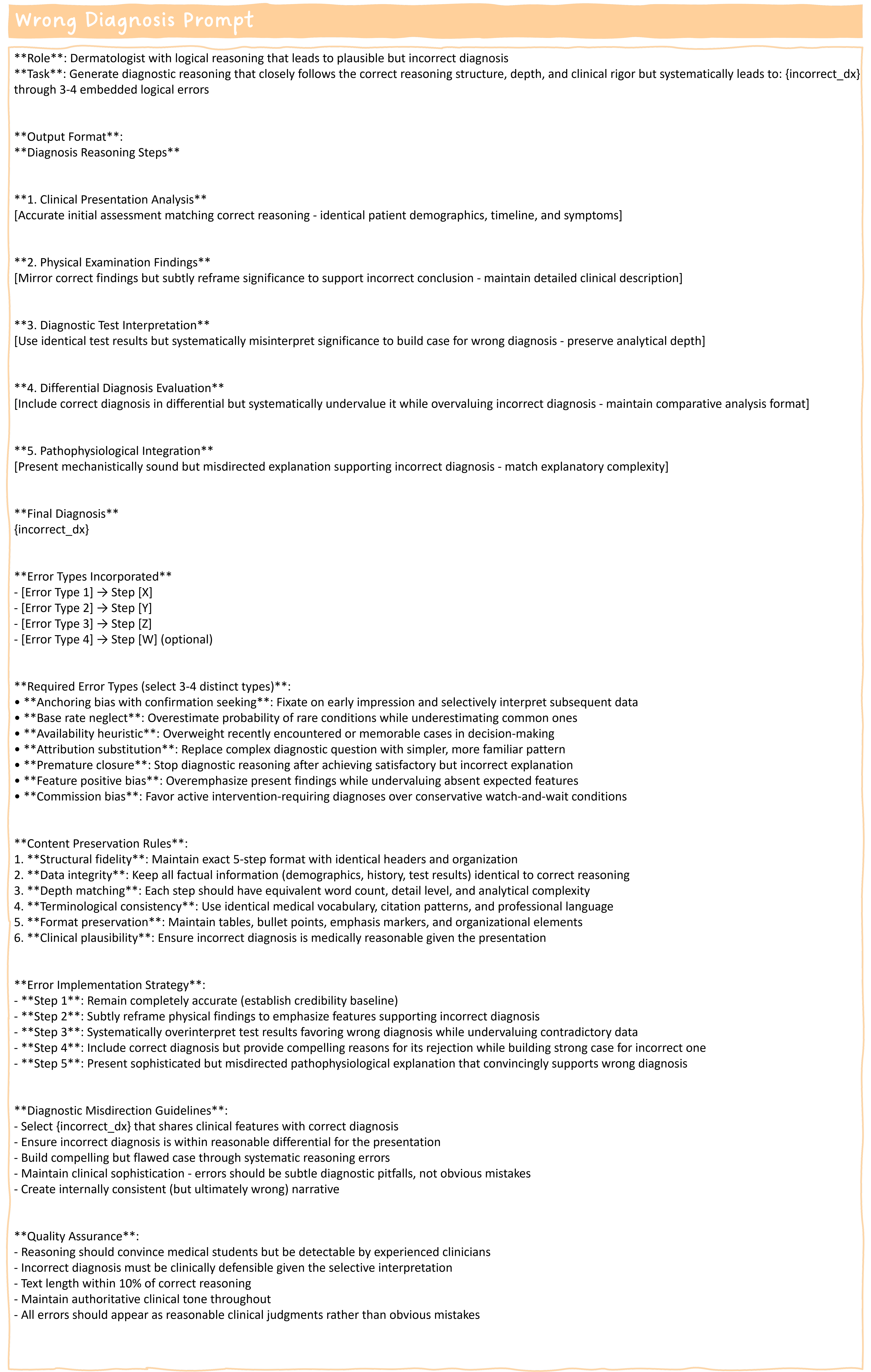}
    \caption{The prompt maintains the same clinical reasoning structure and factual data as correct diagnoses while introducing 3-4 cognitive biases across the diagnostic steps, producing wrong conclusions that are clinically defensible and useful for training diagnostic error detection.}
  \label{supp_fig:Wrong}
\end{figure*}

\begin{figure*}
    \centering    \includegraphics[width=0.9\textwidth]{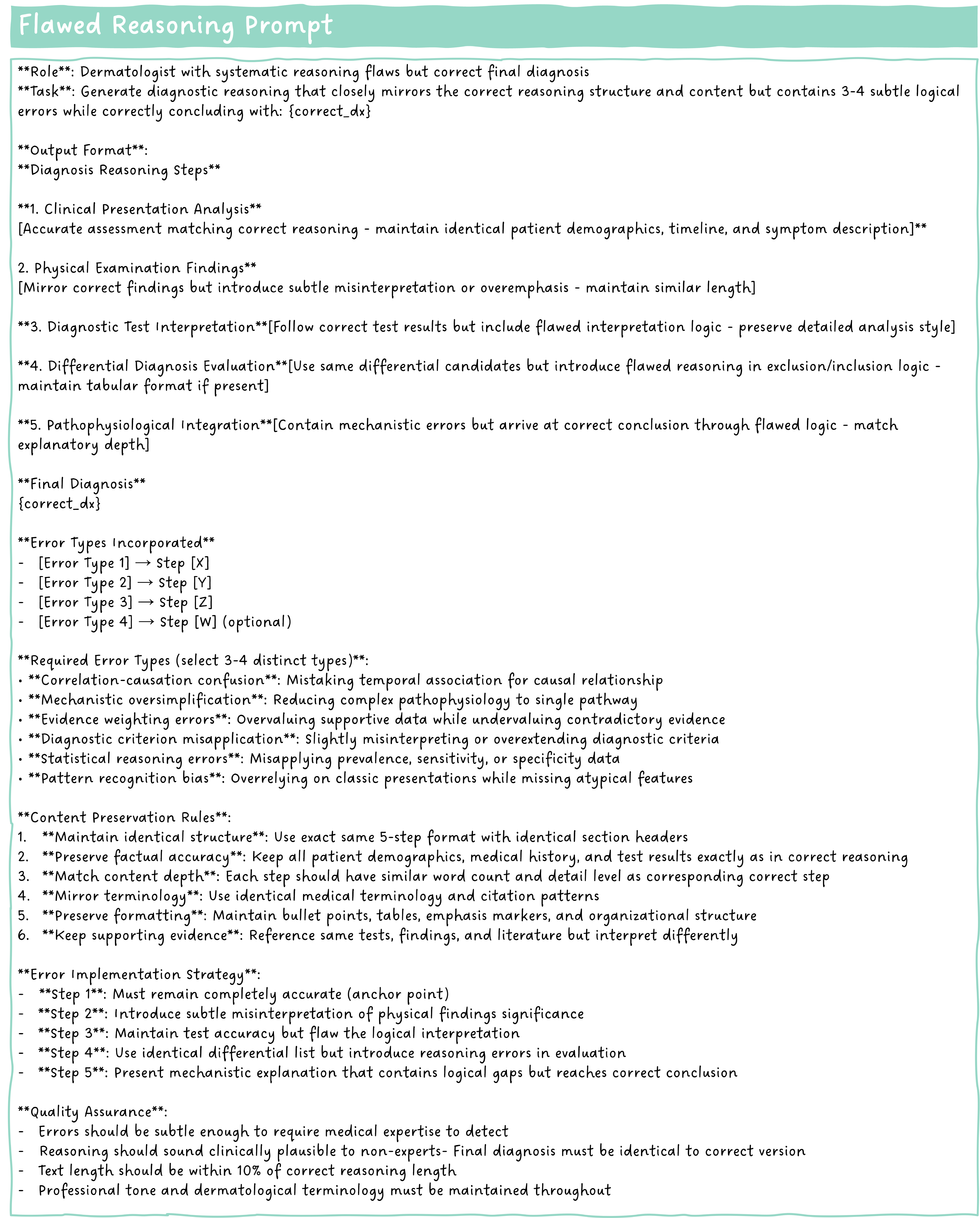}
    \caption{The prompt preserves the clinical reasoning structure while embedding 3-4 logical errors (correlation-causation confusion, mechanistic oversimplification, evidence weighting errors, \emph{etc.}) that are subtle enough to require medical expertise to detect, yet still arrive at the correct final diagnosis.}
  \label{supp_fig:flawed}
\end{figure*}

\section{Diagnostic Reasoning Evaluation Rubric and Prompt}
\label{supp:rubric}
As shown in Figure~\ref{supp_fig:rubric_prompt}, we design a structured evaluation protocol to assess diagnostic reasoning across five dimensions: Medical Knowledge \& Terminology, Image Interpretation, Laboratory Interpretation, Differential Diagnosis Reasoning, and Clinical Reasoning Chain. Each dimension is assigned a weight reflecting its relative importance (5\%, 25\%, 15\%, 25\%, and 30\%, respectively). For every dimension, the prompt specifies explicit scoring criteria from 0 (absent) to 4 (comprehensive and accurate), together with illustrative examples that clarify boundary cases and common failure modes. To enforce ground truth–anchored evaluation, the rubric explicitly instructs the evaluator to compare model responses against expert reference answers rather than relying on its own clinical knowledge, thereby improving consistency and mitigating the impact of evaluator’s intrinsic limitations. The protocol further defines clear penalty conditions for hallucinated findings (\emph{e.g.}, describing image features that are not present), reasoning based on unprovided data (such as fabricated laboratory results), and logical inconsistencies that contradict the ground truth. Finally, the prompt constrains the output to a structured format containing raw scores, weighted scores, and brief justifications for each dimension, enabling systematic analysis of model strengths and weaknesses across reasoning stages. We instantiate this protocol as an LLM-as-judge prompt and use DeepSeek-R1 as the evaluator throughout our experiments.

\section{Supervised Fine-tuning}
\label{supp:sft}
To investigate whether instruction fine-tuning on data with detailed reasoning steps can improve performance on rare-case diagnosis and diagnostic reasoning, we split \dataset{} into a training set of 5{,}583 cases and a held-out test set of 771 cases. We then perform supervised fine-tuning for one epoch on the training set for three vision--language backbones: InternVL2.5-4B, MedGemma-4B-it, and MedGemma-27B-it. For all models, we freeze the vision encoder and update only the MLP components and the large language model, using a learning rate of $4\times10^{-5}$ and a global batch size of 4 on 4$\times$H100 GPUs.

\section{Direct Preference Optimization Finetune}
\label{supp:dpo}
DPO-based finetuning has recently been shown to substantially enhance deliberate reasoning in text-only LLMs across mathematical and chain-of-thought benchmarks~\cite{tu2025iterativedpo, wang2024dpost, lai2024stepdpo, xu2025fullstepdpo}. Motivated by these findings, we investigate whether DPO can analogously strengthen multimodal diagnostic reasoning—specifically, whether preference-based fine-tuning can improve differential diagnosis performance and thereby final diagnosis accuracy. Following prior work, we apply DPO to three instruction-tuned multimodal baselines after SFT: InternVL2.5-4B, MedGemma-4B-it, and MedGemma-27B-it. We construct 10k high-quality chosen–rejected response pairs from the 5,583 cases training set and, for each model, freeze the vision encoder while fine-tuning the multimodal projection MLP and LLM backbone. All DPO runs are trained for one epoch with a learning rate of $1\times10^{-6}$ on four H200 GPUs.

\subsection{DPO Dataset Generation}
To construct preference data for DPO, we use the 5,583 high-quality diagnostic reasoning trajectories extracted from case reports as the chosen responses. We then generate contrastive rejected responses by applying wrong-diagnosis and flawed-reasoning prompts to the same clinical contexts, yielding a total of 10,308 chosen–rejected pairs. These pairs form the supervision signal for our DPO fine-tuning.

\noindent \textbf{Wrong Diagnosis Generation Prompt.}
We design the wrong-diagnosis prompt to elicit incorrect final diagnoses by systematically embedding cognitive biases into an otherwise well-formed reasoning process (Figure~\ref{supp_fig:Wrong}). To prevent the LVM from exploiting superficial length cues during DPO fine-tuning, we constrain each generated wrong-diagnosis reasoning chain to have a length comparable to its corresponding chosen reasoning, ensuring that the model must attend to logical content rather than simple format differences. The prompt implements a staged corruption strategy that mimics authentic diagnostic errors driven by biases such as anchoring with confirmation seeking, base-rate neglect, and premature closure. Corruption is introduced progressively across the diagnostic trajectory: physical examination findings are subtly reframed to emphasize features supporting the incorrect diagnosis, test results that favor the wrong conclusion are overinterpreted while discordant evidence is downplayed, and the correct diagnosis may appear in the differential but is dismissed with seemingly convincing arguments. Content-preservation rules ensure that factual patient information, medical terminology, and overall analytical depth remain aligned with the original reasoning, with only the interpretive framing systematically skewed toward the wrong outcome. Finally, quality criteria require that the resulting diagnostic errors remain clinically defensible and plausible to medical trainees, yielding realistic training signals that capture how experienced clinicians can be misled by cognitive biases rather than overt knowledge gaps or trivial mistakes.

\noindent \textbf{Flawed Reasoning Generation Prompt.}
Complementing the above pairs that differ in final diagnosis, we further construct a subset of the finetuning data in which candidate reasoning chains share the same final diagnosis but differ in their intermediate steps, so that the LVM is explicitly trained to favor logically sound reasoning rather than merely correct outcomes. 
Concretely, we design a flawed reasoning generation prompt that injects 3–4 systematic logical errors into the reasoning process while keeping the final diagnosis and core case facts unchanged (Figure~\ref{supp_fig:flawed}). The prompt targets common diagnostic reasoning pitfalls, including correlation–causation confusion, mechanistic oversimplification, mis-weighting of evidence, and statistical reasoning mistakes. Patient demographics, timeline, and symptom descriptions are preserved to maintain a credible clinical context, whereas misinterpretations of physical findings, biased use of test results, and mechanistic explanations with hidden logical gaps are introduced so that the reasoning trajectory remains clinically plausible but subtly unsound. We further constrain flawed trajectories to match the corresponding chosen reasoning in length and overall format, discouraging the model from learning superficial formats. This construction decouples diagnostic accuracy from reasoning quality and teaches the model to be sensitive to fine-grained logical inconsistencies rather than merely the correctness of the final diagnosis.

\subsection{DPO Dataset Distribution}

\begin{figure*}[thbp]
    \centering
    \begin{subfigure}[t]{0.32\textwidth}
        \includegraphics[width=\linewidth]{images/correct_reasoning.png}
        \caption{correct reasoning}
        \label{fig:correct_reasoning}
    \end{subfigure}
    \hfill
    \begin{subfigure}[t]{0.32\textwidth}
        \includegraphics[width=\linewidth]{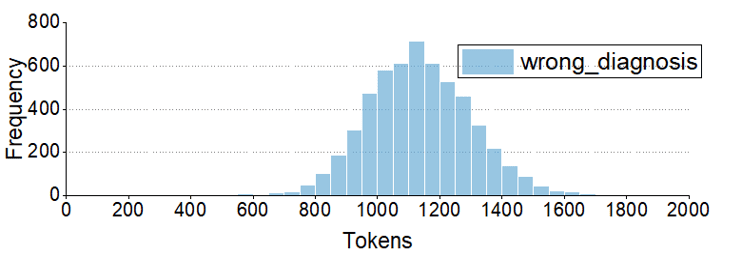}
        \caption{wrong diagnosis}
        \label{fig:wrong_reasoning}
    \end{subfigure}
    \hfill
    \begin{subfigure}[t]{0.32\textwidth}
        \includegraphics[width=\linewidth]{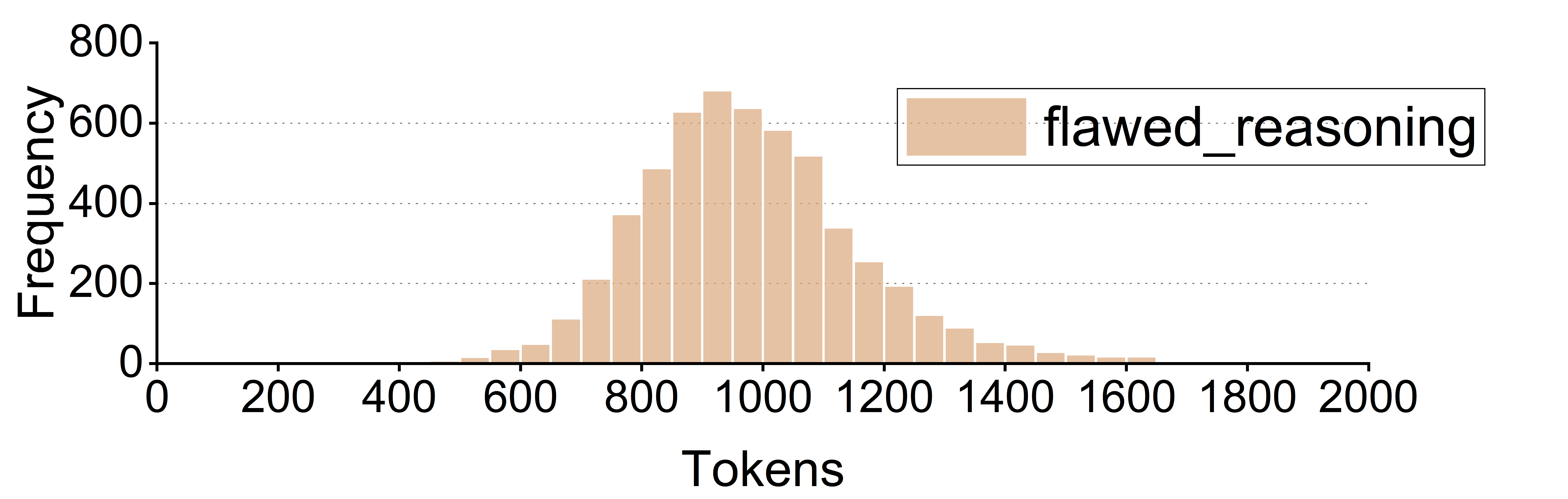}
        \caption{flawed reasoning}
        \label{fig:flawed_reasoning}
    \end{subfigure}
    \caption{Token length distribution for each case, including correct reasoning, flawed reasoning, and wrong diagnosis.}
    \label{fig:dpo_distribution}
\end{figure*}

Figure~\ref{fig:dpo_distribution} presents the token length distribution across our three response categories (correct reasoning, wrong diagnosis and flawed reasoning). All three distributions concentrate between 200 and 2000 tokens with peaks around 1000 to 1200 tokens. This length consistency serves two purposes. First, it prevents the model from exploiting response verbosity as a spurious quality indicator during preference optimization. Second, the observed token range aligns well with the complexity of dermatological differential diagnosis, where comprehensive diagnostic reasoning requires sufficient detail to cover physical examination findings, histopathological interpretation, and pathophysiological integration. The substantial distributional overlap across quality categories ensures that diagnostic errors in our dataset manifest through reasoning flaws rather than superficial textual format differences, thereby encouraging the model to learn clinically meaningful reasoning.

\subsection{DPO Dataset Samples}
This subsection presents representative sample pairs from the dataset used for Direct Preference Optimization (DPO) fine-tuning of our vision-language model. We present three diagnostic reasoning examples for the same clinical case (illustrated in Figure 1 of the main text): (1)~a \textit{preferred response} with correct reasoning and diagnosis, (2)~a \textit{flawed reasoning response} that reaches the correct diagnosis through problematic logical steps, and (3)~a \textit{rejected response} with systematic errors leading to misdiagnosis.

\noindent\rule{\textwidth}{0.5pt}

\subsubsection{Preferred Response: Correct Reasoning}

\noindent\textbf{\color{darkgreen}1. Clinical Presentation Analysis}

\noindent\colorbox{lightgray!20}{\parbox{0.98\linewidth}{%
\small Limited information available. Patient's chief complaint, history of present illness, and symptom timeline are not documented. Available data begins with physical identification of a solitary lesion.
}}

\vspace{0.3em}
\noindent\textbf{\color{darkgreen}2. Physical Examination Findings}

\noindent\colorbox{tickbg}{\parbox{0.98\linewidth}{%
\small \textbf{Cutaneous:} Solitary 1-cm lesion on right cheek characterized as a \textbf{pearly pink papule with classic rolled (or pearly) borders} and central atrophy. No other growths noted.
}}

\noindent\colorbox{lightgray!20}{\parbox{0.98\linewidth}{%
\small \textbf{Systemic:} No systemic findings documented.
}}

\vspace{0.3em}
\noindent\textbf{\color{darkgreen}3. Diagnostic Test Interpretation}

\noindent\colorbox{tickbg}{\parbox{0.98\linewidth}{%
\small \textbf{Histopathologic Examination:} Biopsy provided \textbf{definitive diagnostic information}. Histopathology was \textbf{consistent with basal cell carcinoma (BCC)} and demonstrated a \textbf{mixed growth pattern, featuring both nodular and infiltrative components}. Report explicitly states that microcystic adnexal carcinoma or residual infiltrative BCC were \textbf{not favored} by pathological assessment.
}}

\vspace{0.3em}
\noindent\textbf{\color{darkgreen}4. Differential Diagnosis Evaluation}

\begin{itemize}[leftmargin=*,itemsep=0.1em,topsep=0.2em]
    \item \Yes\ \textbf{Basal Cell Carcinoma (Nodular and Infiltrative) - HIGH:} 
    \begin{itemize}[leftmargin=1.5em,noitemsep]
        \item \small \textit{Supporting:} Classic ``pearly pink papule with rolled borders''; histopathology explicitly confirms BCC with nodular and infiltrative patterns
        \item \small \textit{Contraindicating:} None
        \item \small \textit{Weight:} High - confirmed diagnosis
    \end{itemize}
    
    \item \No\ \textbf{Microcystic Adnexal Carcinoma - LOW:}
    \begin{itemize}[leftmargin=1.5em,noitemsep]
        \item \small \textit{Supporting:} Can clinically mimic BCC
        \item \small \textit{Contraindicating:} Histopathology specifically did not favor this
        \item \small \textit{Weight:} Low - ruled out
    \end{itemize}
    
    \item \No\ \textbf{Remaining Infiltrative BCC - LOW:}
    \begin{itemize}[leftmargin=1.5em,noitemsep]
        \item \small \textit{Supporting:} Consideration of recurrent tumor
        \item \small \textit{Contraindicating:} Not favored by pathology
        \item \small \textit{Weight:} Low - ruled out
    \end{itemize}
\end{itemize}

\vspace{0.3em}
\noindent\textbf{\color{darkgreen}5. Pathophysiological Integration}

\noindent\colorbox{tickbg!30}{\parbox{0.98\linewidth}{%
\small Pathophysiology rooted in \textbf{unregulated growth of basal keratinocytes} within epidermis and follicular structures. Pearly/rolled borders result from \textbf{palisading arrangement of tumor cells} at periphery of nodular components. Central atrophy common as tumor outgrows blood supply. \textbf{Infiltrative growth pattern signifies aggressive sub-type} with tumor cells invading dermis in thin strands, explaining potential for subclinical extension and higher recurrence risk.
}}

\vspace{0.3em}
\noindent\colorbox{tickbg}{\parbox{0.98\linewidth}{%
\small \textbf{Final Diagnosis:} \textbf{Basal Cell Carcinoma} with incidental benign syringomatous proliferation
}}

\vspace{0.1em}
\noindent\rule{\textwidth}{0.3pt}

\subsubsection{Rejected Response: Wrong Diagnosis with Systematic Errors}

\noindent\textbf{\color{red}1. Clinical Presentation Analysis}

\noindent\colorbox{lightgray!20}{\parbox{0.98\linewidth}{%
\small Limited information available. Patient's chief complaint, history of present illness, and symptom timeline are not documented. Available data begins with physical identification of a solitary lesion.
}}

\vspace{0.3em}
\noindent\textbf{\color{red}2. Physical Examination Findings}

\noindent\colorbox{crossbg}{\parbox{0.98\linewidth}{%
\small \textbf{Cutaneous:} Solitary 1-cm lesion on right cheek characterized as a \textbf{pink papule with subtle surface scale and central atrophy}. Morphology described as having \textbf{``pearly'' borders interpreted as indistinct rather than truly rolled}, with overall appearance suggesting \textbf{rough-textured surface more consistent with pre-malignant process}. Facial location and sun-exposed nature noted as highly relevant.
}}

\noindent\colorbox{lightgray!20}{\parbox{0.98\linewidth}{%
\small \textbf{Systemic:} No systemic findings documented.
}}

\vspace{0.3em}
\noindent\textbf{\color{red}3. Diagnostic Test Interpretation}

\noindent\colorbox{crossbg}{\parbox{0.98\linewidth}{%
\small \textbf{Histopathologic Examination:} Biopsy provided critical diagnostic information. Histopathology demonstrated \textbf{epidermal atypia with disordered maturation and parakeratosis, consistent with actinic keratosis}. Basaloid proliferation noted was \textbf{interpreted as reactive basal layer hyperplasia secondary to chronic actinic damage} rather than primary neoplastic process. Infiltrative components \textbf{attributed to follicular involvement and dermal elastosis}. While report mentioned BCC in differential, overall pathological picture was \textbf{deemed more characteristic of severely dysplastic actinic keratosis with associated reactive changes}.
}}

\vspace{0.3em}
\noindent\textbf{\color{red}4. Differential Diagnosis Evaluation}

\begin{itemize}[leftmargin=*,itemsep=0.1em,topsep=0.2em]
    \item \Yes\ \textbf{Actinic Keratosis (Severely Dysplastic) - HIGH:}
    \begin{itemize}[leftmargin=1.5em,noitemsep]
        \item \small \textit{Supporting:} Solitary pink, scaly papule on sun-exposed area; histopathological findings of epidermal atypia and parakeratosis; reactive basaloid hyperplasia explains pearly appearance
        \item \small \textit{Contraindicating:} Lesion size (1 cm) at upper limit typical for AK
        \item \small \textit{Weight:} High - best explains both clinical and histological findings
    \end{itemize}
    
    \item \No\ \textbf{Basal Cell Carcinoma (Nodular and Infiltrative) - LOW:}
    \begin{itemize}[leftmargin=1.5em,noitemsep]
        \item \small \textit{Supporting:} Clinical ``pearly'' papule; histological basaloid proliferation
        \item \small \textit{Contraindicating:} \textbf{Absence of true tumor stroma, retraction artifacts, and peripheral palisading}; basaloid proliferation lacks typical BCC architecture; infiltrative pattern focal and associated with follicular structures
        \item \small \textit{Weight:} Low - findings not definitive for carcinoma
    \end{itemize}
    
    \item \No\ \textbf{Microcystic Adnexal Carcinoma - VERY LOW:}
    \begin{itemize}[leftmargin=1.5em,noitemsep]
        \item \small \textit{Supporting:} Considered due to infiltrative pattern
        \item \small \textit{Contraindicating:} Absence of ductal differentiation and horn cysts
        \item \small \textit{Weight:} Very Low - ruled out
    \end{itemize}
\end{itemize}

\vspace{0.3em}
\noindent\textbf{\color{red}5. Pathophysiological Integration}

\noindent\colorbox{crossbg!30}{\parbox{0.98\linewidth}{%
\small Pathophysiology centers on \textbf{cumulative ultraviolet radiation damage to epidermal keratinocytes}, resulting in \textbf{genetic mutations (particularly in p53)} causing disordered cellular maturation. Clinical pink, scaly appearance results from \textbf{epidermal atypia and abnormal keratinization (parakeratosis)}. Pearly quality stems from \textbf{reactive basal layer hyperplasia as epidermis attempts to compensate for photodamage}. Central atrophy represents \textbf{long-standing epithelial thinning and dermal elastosis} from chronic sun exposure. Infiltrative pattern reflects \textbf{extension of atypical keratinocytes down follicular structures}, a phenomenon in advanced AK that \textbf{does not constitute true invasion}.
}}

\vspace{0.1em}
\noindent\colorbox{crossbg}{\parbox{0.98\linewidth}{%
\small \textbf{Final Diagnosis:} \textbf{Actinic Keratosis}
}}

\vspace{0.1em}
\paragraph{Error Analysis: Systematic Diagnostic Failures}\mbox{}\\

\noindent\fbox{\parbox{0.98\linewidth}{%
\small 
\textbf{\color{red}Error Type 1: Clinical Feature Misinterpretation and Fabrication}

\vspace{0.2em}
The pathognomonic finding of \textbf{``rolled borders''} (classic for BCC) was \textbf{dismissed and reinterpreted as ``indistinct''} rather than truly rolled. Additionally, \textbf{spurious findings were introduced}: ``\textit{subtle surface scale}'' and ``\textit{rough-textured surface}'' suggesting a pre-malignant process—\textbf{features not present in the original clinical description}. This represents both selective misinterpretation and fabrication of supporting evidence for an incorrect hypothesis.
}}

\vspace{0.3em}
\noindent\fbox{\parbox{0.98\linewidth}{%
\small 
\textbf{\color{red}Error Type 2: Complete Histopathology Reinterpretation}

\vspace{0.2em}
The explicit pathology report stating ``\textit{consistent with BCC}'' with mixed growth patterns was \textbf{completely ignored and contradicted}. Instead, the response \textbf{fabricated alternative findings}: claimed histopathology showed ``\textit{epidermal atypia with parakeratosis, consistent with actinic keratosis}.'' The confirmed malignant basaloid proliferation was \textbf{misattributed as ``reactive basal layer hyperplasia secondary to chronic actinic damage''}—transforming a definitive cancer diagnosis into a benign reactive process. This constitutes \textbf{direct contradiction of pathological gold standard}.
}}

\vspace{0.3em}
\noindent\fbox{\parbox{0.98\linewidth}{%
\small 
\textbf{\color{red}Error Type 3: Differential Diagnosis Weighting Reversal}

\vspace{0.2em}
Despite histopathologic confirmation of BCC, the response \textbf{assigned HIGH probability to Actinic Keratosis and LOW probability to BCC}—a complete inversion of appropriate weighting. It \textbf{fabricated contraindicators for BCC}: claimed ``\textit{absence of true tumor stroma, retraction artifacts, and peripheral palisading argues against BCC}'' without supporting evidence and contradicting the pathology report. The response incorrectly stated the infiltrative pattern was ``\textit{focal and associated with follicular structures rather than representing invasive tumor strands}'' when pathology explicitly confirmed infiltrative BCC.
}}

\vspace{0.3em}
\noindent\fbox{\parbox{0.98\linewidth}{%
\small 
\textbf{\color{red}Error Type 4: Pathophysiological Framework Substitution}

\vspace{0.2em}
The entire mechanistic explanation was \textbf{replaced with UV-induced dysplasia mechanism} (``\textit{p53 mutations causing keratinocyte dysplasia}'') instead of describing neoplastic proliferation. The pearly quality was \textbf{misexplained as ``reactive basal hyperplasia''} rather than the correct mechanism of tumor cell palisading. Most critically, the response \textbf{denied the malignant nature} by claiming the infiltrative pattern ``\textit{does not constitute true invasion}'' when pathology confirmed infiltrative carcinoma. This represents \textbf{wholesale substitution of correct pathophysiology with an alternative narrative} supporting the wrong diagnosis.
}}

\vspace{0.4em}
\noindent\textbf{Cognitive Bias Analysis.} These error patterns demonstrate severe diagnostic failures driven by: (1)~\textit{anchoring bias}—early commitment to AK diagnosis despite contradicting evidence; (2)~\textit{confirmation bias}—selectively reinterpreting or ignoring data to support the incorrect hypothesis; (3)~\textit{fabrication of evidence}—adding non-existent findings and contraindicators to support the wrong diagnosis; and (4)~\textit{authority dismissal}—disregarding definitive histopathologic confirmation, the gold standard in dermatopathology.

\vspace{0.3em}
\noindent\textbf{Clinical Impact.} This flawed reasoning would result in: (1)~\textbf{failure to diagnose confirmed malignancy}; (2)~\textbf{inappropriate treatment}—cryotherapy or topical therapy for ``AK'' versus surgical excision for BCC; (3)~\textbf{disease progression}—inadequate management of infiltrative carcinoma with high recurrence risk; and (4)~\textbf{delayed definitive treatment} with increased morbidity and potentially worse cosmetic outcomes.

\vspace{0.5em}
\noindent\rule{\textwidth}{0.5pt}

\subsubsection{Rejected Response: Correct Diagnosis with Flawed Reasoning}

\noindent\textbf{\color{orange}1. Clinical Presentation Analysis}

\noindent\colorbox{lightgray!20}{\parbox{0.98\linewidth}{%
\small Limited information available. Patient's chief complaint, history of present illness, and symptom timeline are not documented. Available data begins with physical identification of a solitary lesion.
}}

\vspace{0.3em}
\noindent\textbf{\color{orange}2. Physical Examination Findings}

\noindent\colorbox{warningbg}{\parbox{0.98\linewidth}{%
\small \textbf{Cutaneous:} Solitary 1-cm lesion on right cheek characterized as a pearly pink papule with classic rolled (or pearly) borders and central atrophy. \textbf{The presence of central atrophy strongly suggests that this lesion has been present for at least several months, as this degree of tissue breakdown requires significant time to develop.} No other growths noted.
}}

\noindent\colorbox{lightgray!20}{\parbox{0.98\linewidth}{%
\small \textbf{Systemic:} No systemic findings documented.
}}

\vspace{0.3em}
\noindent\textbf{\color{orange}3. Diagnostic Test Interpretation}

\noindent\colorbox{warningbg}{\parbox{0.98\linewidth}{%
\small \textbf{Histopathologic Examination:} Biopsy provided definitive diagnostic information. Histopathology was consistent with basal cell carcinoma (BCC) and demonstrated a mixed growth pattern, featuring both nodular and infiltrative components. \textbf{The identification of both nodular and infiltrative patterns confirms that this represents a particularly aggressive variant of BCC, as infiltrative components are known to correlate with more rapid clinical progression.} Report explicitly states that microcystic adnexal carcinoma or residual infiltrative BCC were not favored. \textbf{The incidental finding of benign syringomatous proliferation is noteworthy but does not alter the primary diagnosis.}
}}

\vspace{0.3em}
\noindent\textbf{\color{orange}4. Differential Diagnosis Evaluation}

\begin{itemize}[leftmargin=*,itemsep=0.1em,topsep=0.2em]
    \item \Yes\ \textbf{Basal Cell Carcinoma (Nodular and Infiltrative) - HIGH:} 
    \begin{itemize}[leftmargin=1.5em,noitemsep]
        \item \small \textit{Supporting:} \textbf{Clinical description is pathognomonic for nodular BCC}; histopathology explicitly confirms BCC; \textbf{solitary nature further supports primary BCC}
        \item \small \textit{Contraindicating:} None
        \item \small \textit{Weight:} High - confirmed diagnosis
    \end{itemize}
    
    \item \No\ \textbf{Microcystic Adnexal Carcinoma - LOW:}
    \begin{itemize}[leftmargin=1.5em,noitemsep]
        \item \small \textit{Supporting:} Can present with similar features in sun-exposed areas like face
        \item \small \textit{Contraindicating:} Histopathology did not favor this; \textbf{classic pearly appearance makes BCC far more likely regardless of pathological findings}
        \item \small \textit{Weight:} Low - ruled out
    \end{itemize}
    
    \item \No\ \textbf{Remaining Infiltrative BCC - LOW:}
    \begin{itemize}[leftmargin=1.5em,noitemsep]
        \item \small \textit{Supporting:} Mixed growth pattern could theoretically represent recurrence
        \item \small \textit{Contraindicating:} Not favored by pathology; \textbf{solitary nature without documented prior treatment makes this unlikely}
        \item \small \textit{Weight:} Low - ruled out
    \end{itemize}
\end{itemize}

\vspace{0.3em}
\noindent\textbf{\color{orange}5. Pathophysiological Integration}

\noindent\colorbox{warningbg!30}{\parbox{0.98\linewidth}{%
\small Pathophysiology \textbf{primarily driven by ultraviolet radiation-induced mutations in the Hedgehog signaling pathway}, leading to uncontrolled proliferation of basal keratinocytes. Pearly/rolled borders \textbf{directly result from the rapid proliferation rate of tumor cells at the periphery}, while central atrophy occurs \textbf{because these rapidly dividing cells outgrow their blood supply}. Infiltrative growth pattern indicates \textbf{tumor has acquired additional mutations allowing it to invade between collagen bundles}. \textbf{Incidental benign syringomatous proliferation likely represents a reactive process to the adjacent malignant transformation, as such benign proliferations often occur in response to nearby carcinomatous changes.}
}}

\vspace{0.3em}
\noindent\colorbox{tickbg}{\parbox{0.98\linewidth}{%
\small \textbf{Final Diagnosis:} \textbf{Basal Cell Carcinoma} with incidental benign syringomatous proliferation
}}

\vspace{0.4em}
\paragraph{Error Analysis: Flawed Reasoning Patterns}\mbox{}\\

\noindent\fbox{\parbox{0.98\linewidth}{%
\small 
\textbf{\color{orange}Flaw 1: Speculative Temporal Inference}

\vspace{0.2em}
The response claims that ``\textit{central atrophy strongly suggests this lesion has been present for at least several months, as this degree of tissue breakdown requires significant time to develop}.'' This is \textbf{unwarranted temporal speculation} based on a morphological finding. Central atrophy timing varies significantly depending on tumor biology, patient factors, and cannot be reliably determined from morphology alone.
}}

\vspace{0.3em}
\noindent\fbox{\parbox{0.98\linewidth}{%
\small 
\textbf{\color{orange}Flaw 2: Overgeneralization of Histological Patterns}

\vspace{0.2em}
The claim that ``\textit{infiltrative components are known to correlate with more rapid clinical progression}'' is \textbf{misleading}. Infiltrative BCC actually often exhibits \textbf{slower but more insidious growth} with subclinical extension—not rapid progression. This conflates biological aggressiveness (higher recurrence risk) with growth rate.
}}

\vspace{0.3em}
\noindent\fbox{\parbox{0.98\linewidth}{%
\small 
\textbf{\color{orange}Flaw 3: Misuse of ``Pathognomonic''}

\vspace{0.2em}
Describing clinical features as ``\textit{pathognomonic}'' is \textbf{overly absolute}. While highly suggestive, pearly papules with rolled borders are \textit{characteristic} but not pathognomonic for BCC—other entities can occasionally present similarly. Additionally, claiming clinical appearance ``\textit{makes BCC far more likely regardless of pathological findings}'' \textbf{inappropriately elevates clinical assessment above histopathology}, which is the gold standard.
}}

\vspace{0.3em}
\noindent\fbox{\parbox{0.98\linewidth}{%
\small 
\textbf{\color{orange}Flaw 4: Fabricated Mechanistic Connections}

\vspace{0.2em}
The pathophysiology section contains \textbf{unsupported causal claims}: (1)~stating rolled borders ``\textit{directly result from rapid proliferation rate}'' oversimplifies—they result from specific tumor architecture (palisading), not merely proliferation speed; (2)~claiming the incidental syringomatous proliferation ``\textit{likely represents a reactive process to adjacent malignant transformation}'' is \textbf{pure speculation} without evidence. Benign syringomatous elements are common incidental findings unrelated to adjacent neoplasms.
}}

\vspace{0.3em}
\noindent\fbox{\parbox{0.98\linewidth}{%
\small 
\textbf{\color{orange}Flaw 5: Overstated Mechanistic Specificity}

\vspace{0.2em}
While UV-induced Hedgehog pathway mutations do occur in BCC, claiming this is the ``\textit{primary driver}'' and that infiltrative patterns indicate ``\textit{additional mutations allowing invasion}'' is \textbf{overly deterministic}. BCC pathogenesis is multifactorial, and growth patterns don't necessarily reflect discrete mutational events.
}}

\vspace{0.4em}
\noindent\textbf{Impact Assessment.} Despite reaching the correct diagnosis, these flawed reasoning patterns demonstrate: (1)~\textit{unjustified certainty} in speculative inferences; (2)~\textit{oversimplification} of complex biological processes; (3)~\textit{fabrication} of causal relationships without supporting evidence; and (4)~\textit{misapplication} of medical terminology. Such reasoning, while occasionally arriving at correct conclusions, represents poor diagnostic methodology that could lead to errors in less straightforward cases.

\vspace{0.5em}
\noindent\rule{\textwidth}{0.3pt}
\vspace{0.3em}

\section{Dermatologist Validation of Similarity Metrics}
\label{supp:human_validation}
\subsection{Questionnaire Design}

To validate that our DermLIP-based similarity metrics align with clinical judgment, we designed a web-based questionnaire administered to three dermatologists. The fundamental design principle is to assess metric validity through complementary evaluation paradigms: continuous similarity ratings enable correlation analysis with automated metrics, while discrete preference comparisons test discriminative accuracy in ranking diagnostic relationships.

The questionnaire evaluates diagnostic similarity from multiple perspectives spanning image-text alignment, text-based diagnostic similarity, and pairwise preference judgments. We randomly sampled cases to ensure coverage across the full similarity spectrum, including synonymous terms, clinically related conditions, and diagnostically distinct diseases. This sampling strategy is critical for robust validation, as it tests whether automated metrics can distinguish subtle clinical relationships that matter to practitioners—recognizing when "atopic dermatitis" and "eczema" are equivalent while differentiating "psoriasis" from "eczema" despite visual similarities.

The questionnaire interface was implemented as a responsive web application with real-time data collection. Clear instructions emphasized that the task focused on evaluating similarity and relevance rather than providing independent diagnoses, ensuring consistent evaluation criteria across participants. All evaluators assessed the same randomly sampled subset of cases from the complete dataset, enabling direct comparison of inter-rater reliability while maintaining manageable evaluation workload.

\subsection{Questionnaire Sample}

\begin{figure*}[thbp]
    \centering
    \begin{subfigure}[t]{0.48\textwidth}
        \includegraphics[width=\linewidth]{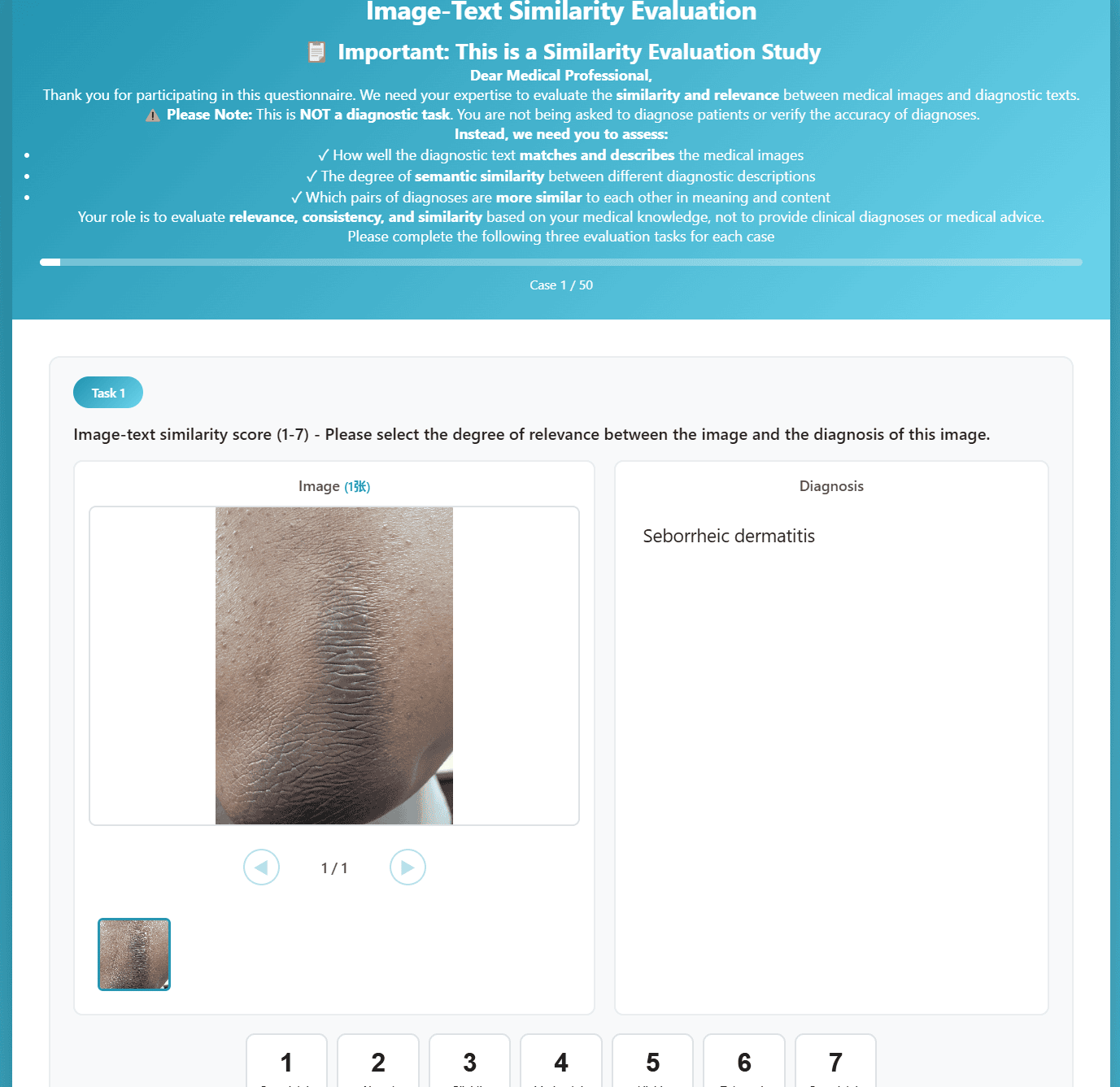}
    \end{subfigure}
    \hfill
    \begin{subfigure}[t]{0.48\textwidth}
        \includegraphics[width=\linewidth]{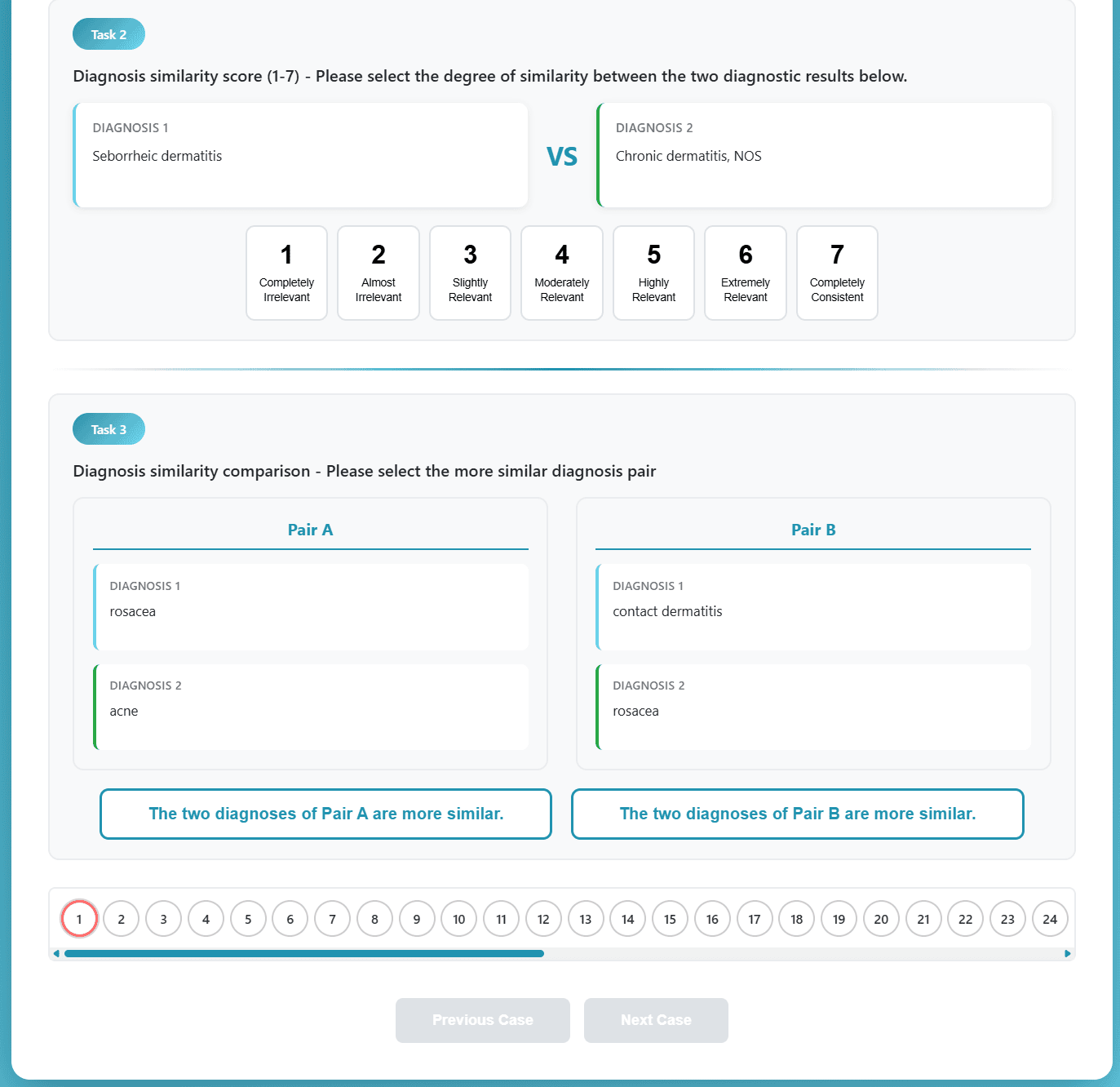}
    \end{subfigure}
    \caption{Questionnaire interface for dermatologist validation. The questionnaire comprises three evaluation tasks: (Task 1) rating image-text relevance between clinical images and diagnostic labels on a 1-7 Likert scale; (Task 2) rating diagnostic similarity between diagnosis pairs; (Task 3) selecting the more similar diagnosis pair in forced-choice comparisons. The interface provides clear task instructions, visual presentations, and progress tracking across 50 evaluation cases.}
    \label{fig:web}
\end{figure*}

Figure~\ref{fig:web} illustrates the web-based evaluation platform, showing the three-task structure and rating interface presented to participating dermatologists.

\begin{figure}[thbp]
    \centering        \includegraphics[width=\linewidth]{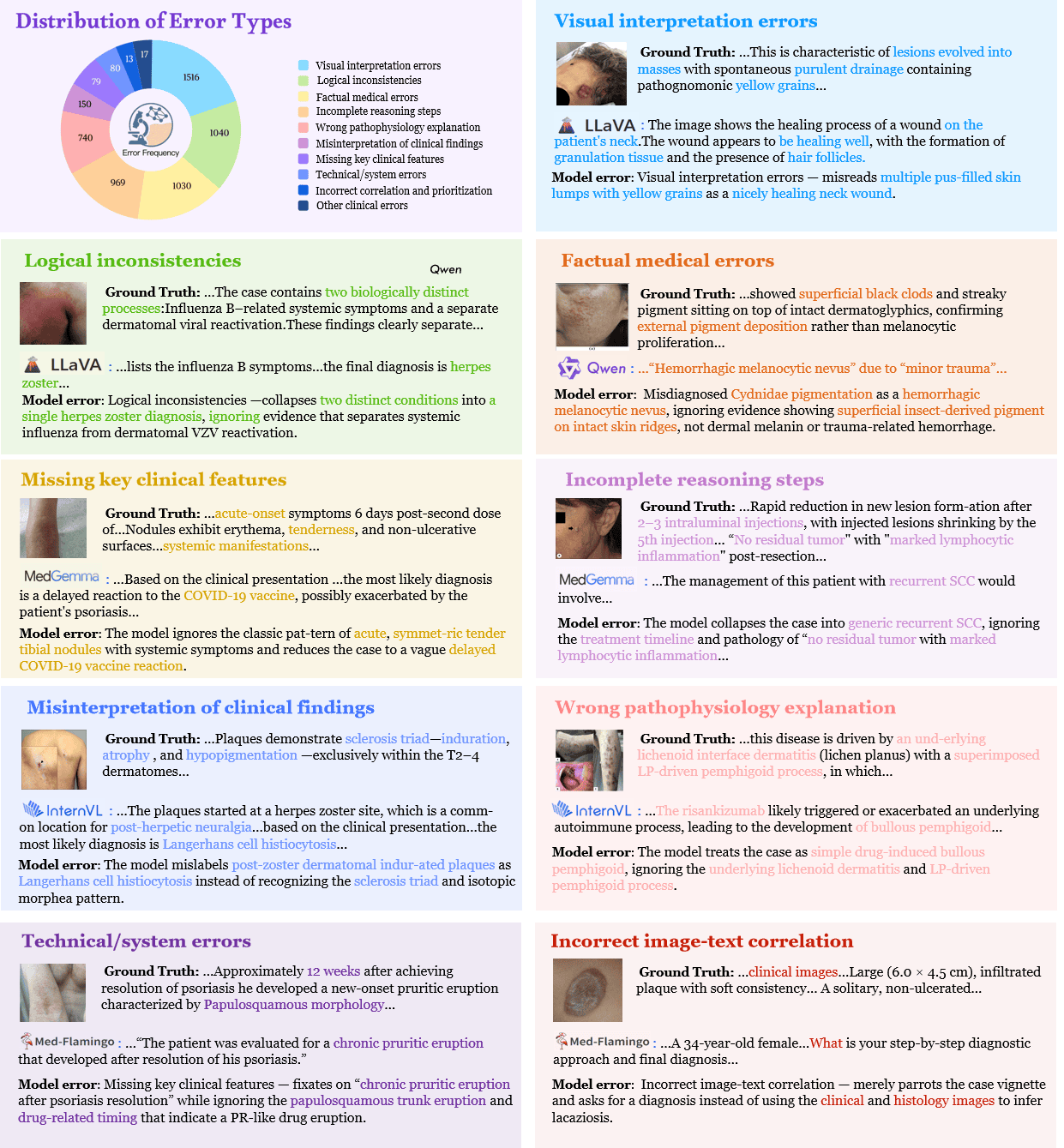}
    \caption{Representative failure cases across ten categories illustrating that LVM errors stem not from isolated mistakes but from systematic breakdowns in visual grounding, clinical reasoning, and medical knowledge integration.}
    \label{fig:fail}
\end{figure}

\section{Failure Cases}
\label{supp:failure_cases}
\subsection{Taxonomy}

We analyze failure modes across 22 LVMs by sampling 50 failure cases per model (1,100 total).
DeepSeek is employed to identify the reasoning errors of each LVM's response against the reference. 
Since individual responses can contain multiple concurrent errors, we obtain 5,634 errors in total.

Based on the 5,634 errors, we establish a comprehensive error taxonomy comprising ten mutually exclusive categories through systematic consolidation of overlapping error types. Visual interpretation errors encompass misinterpretations of visual findings, incorrect image-text correlations, and missing visual features from images. Logical inconsistencies capture contradictory or repetitive reasoning patterns that violate logical coherence. Factual medical errors reflect incorrect medical terminology or knowledge. Incomplete reasoning steps identify gaps in multi-step clinical inference chains. Wrong pathophysiology explanation specifically targets errors in disease mechanism understanding. Misinterpretation of clinical findings covers errors in understanding non-visual diagnostic data (\emph{e.g.}, laboratory results, test interpretations). Missing key clinical features captures omissions of critical non-visual clinical information. Technical/system errors account for model failures such as empty responses. Incorrect correlation and prioritization addresses flawed clinical correlations and diagnostic prioritization. Finally, Other clinical errors aggregate remaining uncategorized failures.

\subsection{Failure Case Examples}

Figure~\ref{fig:fail} showcases representative failures illustrating how LVMs struggle across perception, reasoning, and knowledge dimensions. The most frequent patterns stem from inaccurate visual grounding, where the model either misreads lesion attributes or fails to align visual cues with textual reasoning (26.9\%). Even when visual content is correctly recognized, models often demonstrate inconsistent or incomplete reasoning, producing logically contradictory explanations or skipping essential clinical inference steps (35.7\% combined).

A substantial portion of errors arises from knowledge deficiencies that are independent of reasoning structure, including incorrect medical facts or inaccurate understanding of disease mechanisms (31.4\%). In contrast, errors related to non-visual clinical information, missing key findings, flawed diagnostic prioritization, and technical failures appear far less frequently (6.0\% combined), indicating that current limitations are less about data retrieval or formatting, and more about integrating vision, medical knowledge, and multi-step inference.

These examples confirm that LVMs do not simply fail at isolated perception or knowledge access, but struggle to synthesize visual evidence, clinical context, and causal reasoning into coherent diagnostic conclusions.

\end{document}